\documentclass{article}
\ifdefined\pdfsuppressptexinfo \pdfsuppressptexinfo=-1 \fi
\usepackage{microtype}
\usepackage{graphicx}
\usepackage{booktabs}
\usepackage[accepted,nohyperref]{mlsys2025}
\usepackage{etoolbox}
\makeatletter\renewcommand{\Notice@String}{\noindent Code: {\urlstyle{same}\url{https://github.com/vicky157/grpo-validation}}.}
\patchcmd{\printAffiliationsAndNotice}{\ \\ \Notice@String}{\Notice@String}{}{\typeout{arXiv: the notice line was not patched}}\makeatother
\usepackage[T1]{fontenc}
\usepackage[utf8]{inputenc}

\usepackage{amsmath,amsfonts,bm}

\def\eqref#1{equation~\ref{#1}}

\def\1{\bm{1}}

\DeclareMathAlphabet{\mathsfit}{\encodingdefault}{\sfdefault}{m}{sl}
\SetMathAlphabet{\mathsfit}{bold}{\encodingdefault}{\sfdefault}{bx}{n}

\newcommand{\E}{\mathbb{E}}

\newcommand{\R}{\mathbb{R}}

\usepackage{xurl}
\usepackage{amsmath,amssymb,amsthm}
\usepackage{mathtools}
\usepackage{array}
\usepackage{tabularx}
\usepackage{tikz}
\usetikzlibrary{arrows.meta,positioning,calc,shapes.geometric,backgrounds}
\usepackage{xcolor}
\usepackage{enumitem}
\usepackage{cleveref}
\usepackage{textcomp}
\definecolor{pvChar}{HTML}{252B34}
\definecolor{pvBlue}{HTML}{1C4F91}
\definecolor{pvRed}{HTML}{B02D36}
\definecolor{pvGreen}{HTML}{18704A}
\definecolor{pvGray}{HTML}{66717E}
\newlength{\ciW}
\def\cimin{-0.05}\def\cimax{0.05}
\newcommand{\setciscale}[2]{\gdef\cimin{#1}\gdef\cimax{#2}}
\newcommand{\cifrac}[2]{\pgfmathsetmacro{#1}{((#2)-(\cimin))/((\cimax)-(\cimin))}}
\newcommand{\ciback}{\cifrac{\xa}{-0.02}\cifrac{\xb}{0.02}\cifrac{\xz}{0}%
  \fill[pvGray!14] (\xa\ciW,-0.2) rectangle (\xb\ciW,0.2);
  \draw[pvGray!70,line width=0.3pt] (\xz\ciW,-0.2) -- (\xz\ciW,0.2);
  \path (0,0) (\ciW,0);}
\newcommand{\ci}[3]{\tikz[baseline=-0.55ex]{\ciback
  \cifrac{\xm}{#1}\cifrac{\xl}{#2}\cifrac{\xh}{#3}%
  \pgfmathtruncatemacro{\cc}{((#2)>0)?1:(((#3)<0)?2:0)}%
  \ifcase\cc\def\cicol{pvChar}\or\def\cicol{pvBlue}\or\def\cicol{pvRed}\fi
  \draw[\cicol,line width=0.9pt] (\xl\ciW,0) -- (\xh\ciW,0);
  \fill[\cicol] (\xm\ciW,0) circle (1.25pt);}}
\newcommand{\cidots}[3]{\tikz[baseline=-0.55ex]{\ciback
  \foreach \v in {#1,#2,#3} {\cifrac{\xm}{\v}\fill[pvChar] (\xm\ciW,0) circle (1.0pt);}}}
\newcommand{\ciaxis}[1]{\tikz[baseline=-0.4ex]{\path (0,0) (\ciW,0);
  \foreach \t/\l in {#1} {\cifrac{\xt}{\t}\node[font=\sffamily\tiny,text=pvGray,inner sep=0pt] at (\xt\ciW,0) {\l};}}}
\newcommand{\claim}[1]{\textbf{\textit{#1}}}%
\newcommand{\answer}[1]{\par\vspace{1pt}\noindent{\setlength{\fboxsep}{2.5pt}\colorbox{pvGray!12}{\parbox{\dimexpr\columnwidth-2\fboxsep\relax}{\small\textbf{Answer.} #1}}}\par\vspace{1pt}}
\newcommand{\expl}{\textsuperscript{\textdagger}}
\newcommand{\nruns}{1{,}668}
\newcommand{\gpuh}{1{,}468}
\newtheorem{theorem}{Theorem}
\newtheorem{proposition}{Proposition}

\newcommand{\grad}{\operatorname{grad}}
\newcommand{\vecop}{\operatorname{vec}}
\newcommand{\colspace}{\operatorname{colspace}}
\newcommand{\rank}{\operatorname{rank}}
\newcommand{\Span}{\operatorname{span}}

\newcommand{\gout}{g_{\mathrm{out}}}

\newcommand{\Pperp}{P_{\!\perp}}
\newcommand{\PV}{P_{V}}
\newcommand{\wtraj}{w_{\mathrm{traj}}}
\newcommand{\within}{w_{\mathrm{within}}}
\newcommand{\Rbud}{N_{\mathrm{roll}}}
\Crefname{proposition}{Proposition}{Propositions}
\Crefname{theorem}{Theorem}{Theorems}
\Crefname{corollary}{Corollary}{Corollaries}
\Crefname{definition}{Definition}{Definitions}
\Crefname{remark}{Remark}{Remarks}\Crefname{lemma}{Lemma}{Lemmas}\crefname{lemma}{Lemma}{Lemmas}
\crefname{proposition}{Proposition}{Propositions}
\crefname{theorem}{Theorem}{Theorems}
\crefname{remark}{Remark}{Remarks}

\newcommand{\goutVresid}{0.012\pm0.008}

\newcommand{\alignAB}{0.95}

\newcommand{\alignedVsShuf}{+0.97\,[+0.93,+1.01]}
\newcommand{\nmodels}{8}
\newcommand{\famWithin}{0.88\,[0.84,0.92]}
\newcommand{\famPartialTwo}{0.90\,[0.85,0.94]}

\newcommand{\lrDriftMax}{0.77}

\mlsystitlerunning{Are Dense Rewards Worth Their Tokens in GRPO?}
\begin{document}
\twocolumn[{
\mlsystitle{Are Dense Rewards Worth Their Tokens in GRPO?}
\begin{mlsysauthorlist}
\mlsysauthor{Vikash Singh}{cwru}
\mlsysauthor{Weicong Chen}{cwru}
\mlsysauthor{Debargha Ganguly}{cwru}
\mlsysauthor{Yanyan Zhang}{cwru}
\mlsysauthor{Mohsen Hariri}{cwru}
\mlsysauthor{Sreehari Sankar}{cwru}
\mlsysauthor{Wang Yang}{cwru}
\mlsysauthor{Chaoda Song}{cwru}
\mlsysauthor{Alexander Nemecek}{cwru}
\mlsysauthor{Osama Zafar}{cwru}
\mlsysauthor{Yuting Shao}{cwru}
\mlsysauthor{Kai Ye}{cwru}
\mlsysauthor{Biyao Zhang}{cwru}
\mlsysauthor{Xinpeng Li}{cwru}
\mlsysauthor{Shouren Wang}{cwru}
\mlsysauthor{Erman Ayday}{cwru}
\mlsysauthor{Jing Ma}{cwru}
\mlsysauthor{Sanmukh Rao Kuppannagari}{cwru}
\mlsysauthor{Vipin Chaudhary}{cwru}
\end{mlsysauthorlist}
\mlsysaffiliation{cwru}{Department of Computer and Data Sciences, Case Western Reserve University, Cleveland, OH, USA}
\mlsyscorrespondingauthor{Vipin Chaudhary}{vipin@case.edu}
\mlsyskeywords{Reinforcement learning, language models, reward diagnostics}
\vskip 0.3in
\begin{abstract}
RLVR trains language models to reason by rewarding correct final answers. A widely used algorithm, group-relative policy optimization (GRPO), gives every token of a solution the same credit, so dense process rewards have been proposed to credit intermediate steps, for example with a Monte-Carlo reward that scores each step by sampling continuations from it. Comparing such a reward with GRPO per training update, or vetting it with a gradient test on a single batch, leaves out what it costs: in our setting, the continuations alone take $91\%$ of all generated tokens. We introduce a matched-compute protocol that gives every training intervention the same budget, measured in both generated tokens and training GPU-hours, and tests it in preregistered studies with controls for side effects. Across $\nruns$ runs on six models, we find that matching compute reverses the ranking. On Qwen2.5-1.5B, the Monte-Carlo reward leads standard GRPO per update, yet ends $14.3$ percentage points lower at equal tokens and $7.8$ points lower at equal training time, and none of four dense-reward variants outperforms GRPO even when given as much training time. A per-token gradient test orders these rewards as training does but overestimates each by one to two points, as much as the gains it is meant to detect. In contrast, training on the prompts whose predicted success is nearest $1/2$ improves accuracy at matched compute, by $1.7$ points on Qwen3-0.6B and $0.8$ on Qwen2.5-1.5B. These gains replicate on fresh seeds and unseen problems, and a simple rule, that the candidate prompts are predicted to fail more often than not, predicted where they appear in five of six preregistered tests. The protocol also measures what such gains are worth: they come early in training, and standard GRPO with twice the initial learning rate comes within half a point of them. We argue that training interventions should be validated at matched compute, which overturns some apparent gains and confirms others.

\end{abstract}
}]

\printAffiliationsAndNotice{}
\section{Introduction}
\label{sec:intro}

\begin{figure*}[t]
\centering
\begin{tikzpicture}[x=1cm,y=1cm,font=\sffamily\scriptsize,
  lane/.style={font=\sffamily\footnotesize\bfseries,anchor=north west,text=pvChar,inner sep=0pt},
  sub/.style={font=\sffamily\scriptsize,anchor=north west,text=pvGray,inner sep=0pt,align=left},
  note/.style={font=\sffamily\scriptsize,anchor=west,text=pvGray,inner sep=0pt,align=left},
  tiny/.style={font=\sffamily\tiny,anchor=west,text=pvGray,inner sep=0pt},
  hdr/.style={font=\sffamily\scriptsize\bfseries,text=pvGray,anchor=south west,inner sep=0pt,align=left},
  res/.style={font=\sffamily\footnotesize\bfseries,anchor=west,inner sep=0pt},
  inbar/.style={font=\sffamily\scriptsize,text=white,inner sep=0pt},
  prompt/.style={draw=pvChar,line width=0.4pt,rounded corners=1pt,fill=white,minimum width=0.40cm,minimum height=0.28cm,inner sep=0pt},
  flow/.style={-{Stealth[length=1.5mm,width=1.3mm]},line width=0.5pt,draw=pvGray}]
\newcommand{\compl}[7]{%
  \foreach \c [count=\i from 0] in {#2,#3,#4,#5,#6,#7} {
    \fill[\c] ({#1+0.34*\i},-0.155) rectangle ({#1+0.34*\i+0.32},0.055);}
  \draw[pvChar,line width=0.35pt] (#1,-0.155) rectangle ({#1+2.02},0.055);}
\newcommand{\verdict}[2]{\node[draw=pvChar,line width=0.35pt,fill=#2,minimum size=0.21cm,inner sep=0pt] at (#1,-0.05) {};}
\newcommand{\budgetbar}[1]{%
  \fill[pvBlue!50] (9.75,-0.17) rectangle ({9.75+3.3*#1},0.07);
  \fill[pvRed!60] ({9.75+3.3*#1},-0.17) rectangle (13.05,0.07);}
\newcommand{\resbar}[5]{\fill[#5] ({16.30+0.18*(#1)},#3) rectangle ({16.30+0.18*(#2)},#4);}
\tikzset{val/.style={font=\sffamily\scriptsize\bfseries,inner sep=0pt},
         lval/.style={font=\sffamily\scriptsize,inner sep=0pt}}
\node[hdr] at (0,0.42) {Intervention};
\node[hdr] at (3.75,0.42) {How one training iteration assigns credit};
\node[hdr] at (9.75,0.42) {Generated tokens up to \textit{B}};
\node[hdr] at (13.30,0.42) {Accuracy vs.\ outcome-only, at\\equal tokens; light: GPU-hours};
\draw[pvGray!70,line width=0.4pt] (16.30,0.30) -- (16.30,-4.08);
\draw[pvGray!45,line width=0.4pt] (0,0.36) -- (17.0,0.36);
\begin{scope}[yshift=-0.12cm]
\node[lane] at (0,0.20) {Outcome-only GRPO};
\node[sub] at (0,-0.10) {one advantage per completion};
\node[prompt] (p) at (3.98,-0.05) {};
\draw[flow] (p.east) -- (4.52,-0.05);
\compl{4.60}{pvBlue!25}{pvBlue!25}{pvBlue!25}{pvBlue!25}{pvBlue!25}{pvBlue!25}
\verdict{6.80}{pvGreen!75}
\node[note] at (7.05,0.07) {same credit on};
\node[note] at (7.05,-0.19) {every token};
\budgetbar{1}
\node[inbar] at (11.40,-0.05) {rollouts};
\node[tiny] at (9.75,-0.34) {584 updates};
\fill[pvChar] (16.30,-0.05) circle (0.045);
\node[val,text=pvChar,anchor=east] at (16.22,-0.05) {reference};
\end{scope}
\draw[pvGray!25,line width=0.3pt] (0,-0.70) -- (17.0,-0.70);
\begin{scope}[yshift=-1.29cm]
\node[lane] at (0,0.20) {Monte-Carlo reward};
\node[sub] at (0,-0.10) {4 continuations at each of 5 cuts};
\node[prompt] (p) at (3.98,-0.05) {};
\draw[flow] (p.east) -- (4.52,-0.05);
\compl{4.60}{pvGreen!45}{pvGreen!18}{pvRed!20}{pvGreen!35}{pvRed!40}{pvRed!14}
\verdict{6.80}{pvRed!75}
\foreach \k in {1,...,5} {
  \foreach \a in {18,34,50,66} {\draw[pvRed!70,line width=0.3pt] ({4.60+0.34*\k-0.01},0.055) -- ++(\a:0.34);}
  \fill[pvRed] ({4.60+0.34*\k-0.01},0.055) circle (0.025);}
\node[note] at (7.05,0.07) {credit per segment from};
\node[note] at (7.05,-0.19) {continuation success};
\budgetbar{0.09}
\node[inbar] at (11.55,-0.05) {91\% on continuations};
\node[tiny] at (9.75,-0.34) {50 updates};
\resbar{-14.3}{0}{0.03}{0.29}{pvRed}
\node[val,text=white,anchor=west] at ({16.30+0.18*(-14.3)+0.07},0.16) {\textminus14.3};
\resbar{-7.8}{0}{-0.31}{-0.05}{pvRed!35}
\node[lval,text=pvRed,anchor=east] at ({16.30+0.18*(-7.8)-0.06},-0.18) {\textminus7.8};
\node[tiny] at (13.30,-0.46) {ahead at equal updates};
\end{scope}
\draw[pvGray!25,line width=0.3pt] (0,-1.87) -- (17.0,-1.87);
\begin{scope}[yshift=-2.46cm]
\node[lane] at (0,0.20) {Learned-critic reward};
\node[sub] at (0,-0.10) {ridge critic on hidden states};
\node[prompt] (p) at (3.98,-0.05) {};
\draw[flow] (p.east) -- (4.52,-0.05);
\compl{4.60}{pvGreen!40}{pvGreen!24}{pvRed!14}{pvGreen!30}{pvRed!34}{pvRed!20}
\verdict{6.80}{pvRed!75}
\node[draw=pvGreen,line width=0.4pt,rounded corners=1pt,fill=pvGreen!8,font=\sffamily\tiny,inner sep=1.2pt,text=pvGreen] (cr) at (5.61,0.36) {critic};
\foreach \k in {0,...,5} {\draw[pvGreen!80,line width=0.3pt,densely dotted] (cr.south) -- ({4.76+0.34*\k},0.06);}
\node[note] at (7.05,0.07) {no continuations;};
\node[note] at (7.05,-0.19) {critic refit online};
\budgetbar{1}
\node[inbar] at (11.40,-0.05) {rollouts};
\node[tiny] at (9.75,-0.34) {584 updates};
\resbar{-0.4}{0.7}{0.03}{0.29}{pvGreen}
\node[val,text=pvGreen,anchor=east] at ({16.30+0.18*(-0.4)-0.06},0.16) {\textminus0.4 to +0.7};
\resbar{-1.2}{-0.7}{-0.31}{-0.05}{pvGreen!40}
\node[lval,text=pvGreen,anchor=east] at ({16.30+0.18*(-1.2)-0.06},-0.18) {\textminus1.2 to \textminus0.7};
\end{scope}
\draw[pvGray!25,line width=0.3pt] (0,-3.04) -- (17.0,-3.04);
\begin{scope}[yshift=-3.63cm]
\node[lane] at (0,0.20) {Prompt selection};
\node[sub] at (0,-0.10) {keep the 6 of 24 candidates\\[-1pt]closest to predicted success $\tfrac12$\\[-1pt]and 2 more at random};
\foreach \i/\s/\k in {0/12/0,1/55/1,2/30/0,3/80/0,4/48/1,5/20/0,6/62/0,7/5/0,
                      8/45/1,9/90/0,10/15/0,11/52/1,12/35/2,13/70/0,14/8/0,15/58/1,
                      16/25/0,17/50/1,18/85/0,19/40/0,20/65/0,21/10/2,22/75/0,23/28/0} {
  \pgfmathtruncatemacro{\cx}{mod(\i,8)}\pgfmathtruncatemacro{\cy}{div(\i,8)}
  \ifnum\k=0 \def\dr{pvGray!45}\def\lw{0.2pt}\else\ifnum\k=1 \def\dr{pvBlue}\def\lw{0.7pt}\else\def\dr{pvChar}\def\lw{0.7pt}\fi\fi
  \filldraw[fill=pvBlue!\s,draw=\dr,line width=\lw] ({3.75+0.165*\cx},{0.13-0.155*\cy}) rectangle ++(0.125,-0.115);}
\node[font=\sffamily\tiny,text=pvGray,anchor=north west,inner sep=0pt] at (3.75,-0.36) {critic scores candidates};
\draw[flow] (5.12,-0.05) -- (5.45,-0.05);
\foreach \j in {0,1,2} {\node[prompt,minimum width=0.28cm,minimum height=0.17cm] at ({5.70+0.12*\j},{0.04-0.09*\j}) {};}
\draw[flow] (6.18,-0.05) -- (6.48,-0.05);
\compl{6.56}{pvBlue!25}{pvBlue!25}{pvBlue!25}{pvBlue!25}{pvBlue!25}{pvBlue!25}
\verdict{8.76}{pvGreen!75}
\budgetbar{1}
\node[inbar] at (11.40,-0.05) {rollouts};
\node[tiny] at (9.75,-0.34) {+ one prefill pass to score candidates};
\resbar{0}{0.8}{0.03}{0.29}{pvBlue}
\node[val,text=pvBlue,anchor=east] at (16.22,0.16) {+0.8;\ +1.7 on Qwen3-0.6B};
\resbar{0}{0.5}{-0.31}{-0.05}{pvBlue!40}
\node[lval,text=pvBlue,anchor=east] at (16.22,-0.18) {+0.5;\ +1.0 on Qwen3-0.6B};
\end{scope}
\draw[pvGray!60,line width=0.3pt] (13.60,-4.08) -- (16.84,-4.08);
\foreach \v/\l in {-15/\textminus15,-10/\textminus10,-5/\textminus5,0/0} {
  \draw[pvGray!60,line width=0.3pt] ({16.30+0.18*\v},-4.08) -- ++(0,0.05);
  \node[tiny,anchor=north] at ({16.30+0.18*\v},-4.12) {\l};}
\node[tiny,anchor=north west] at (16.45,-4.12) {points};
\end{tikzpicture}
\vspace{-4pt}
\caption{\textbf{Counting every token changes which intervention wins.} Four ways to train Qwen2.5-1.5B-Instruct with GRPO on MATH, compared at a matched budget $B$ of tokens generated during training ($12.25$M). Bars show accuracy minus outcome-only GRPO, to scale ($5$ seeds for the Monte-Carlo reward, $24$ for the critics, and $60$ for prompt selection, $54$ on Qwen3-0.6B); light bars give the same comparisons at equal training GPU-hours on H200s (exploratory).}
\label{fig:teaser}
\end{figure*}
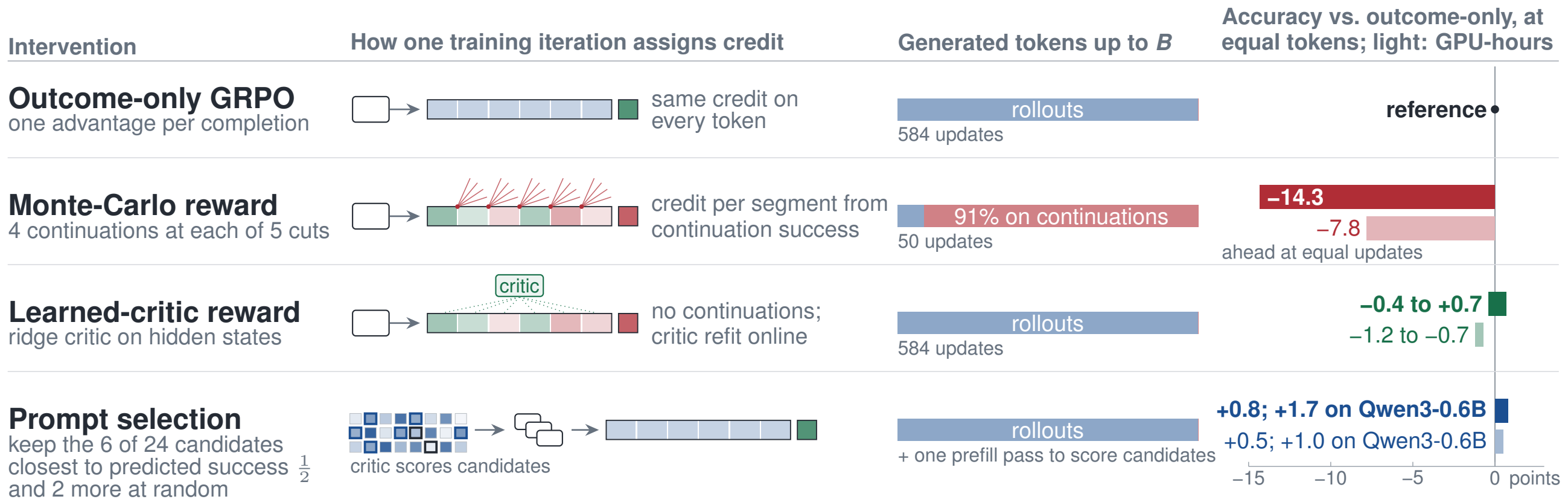

Reinforcement learning with verifiable rewards (RLVR) has become a standard way to improve the reasoning of large language models \citep{shao2024deepseekmath,guo2025deepseekr1}. The model samples solutions to problems whose final answers a \emph{verifier} can check, and training makes the correct solutions more likely. A widely used algorithm, group-relative policy optimization (GRPO), weights every token of a sampled solution, or completion, by the same \emph{advantage}, the completion's reward relative to the others sampled for the same prompt. A solution that sets up a problem correctly but slips in its last step is penalized for its good steps as well as its bad one. \emph{Dense rewards}, also known as process rewards, give different steps different credit, derived from human step labels \citep{lightman2023verify}, from a prover's progress \citep{setlur2024pav}, or from rewards learned on outcome labels \citep{cui2025prime}. A step can also be scored by the probability of reaching the correct answer from it, estimated by sampling continuations from the partial solution, giving a \emph{Monte-Carlo reward} \citep{wang2023mathshepherd,kazemnejad2024vineppo}, or predicted by a learned value model (a \emph{critic}).\looseness=-1 %

\textbf{The hidden cost of dense rewards.} How should we decide whether a dense reward helps? Two checks are cheap enough to run on any candidate reward, and a reward that passes both looks worth using. A \emph{gradient test} takes a fixed batch of samples and asks whether the reward's update points in the same direction as an independent estimate of the gradient of expected accuracy (\Cref{sec:tests}). A \emph{per-update comparison} trains one model with the reward and one with \emph{outcome-only GRPO}, the algorithm above, which credits only the final answer, and asks which is more accurate after the same number of updates. However, neither check considers what a dense reward costs. First, it consumes generated tokens. In our setting, the Monte-Carlo reward's continuations account for $91\%$ of all generated tokens, so with the same tokens, outcome-only GRPO makes $584$ updates and the Monte-Carlo reward $50$ (\Cref{fig:teaser}). Second, it consumes GPU time, though not in proportion to its tokens. The continuations are only generated and scored, never trained on, and an inference engine generates tokens in large batches more cheaply than a training step processes them. With the same tokens, outcome-only GRPO therefore trains three times as long. \textbf{A reward that wins per update can lose per token, and a comparison that counts only tokens can mislead about GPU-hours.}

Even without counting cost, the gradient test can mislead. In our probes, the Monte-Carlo reward's update aligns closely with a reference gradient (mean cosine $0.84$). Yet even with the same number of updates as outcome-only GRPO, adding it with a small weight is equivalent to outcome-only GRPO within $\pm3$ points of greedy accuracy, and adding it with a large weight loses $12.4$ points (\Cref{sec:equalround}). We therefore ask whether a dense reward that passes the gradient test still improves GRPO once its compute is counted, and if not, where the generated tokens should go instead.

\textbf{Our approach.} We compare training interventions at \emph{matched compute}: every method gets the same budget, measured on the same runs in generated tokens, including continuations and discarded samples, and in training GPU-hours, broken down by training stage. We apply this protocol to two interventions that spend compute in opposite ways. \emph{Dense rewards} spend generated tokens to improve each update, whereas \emph{prompt selection} spends one forward pass over candidate prompts to choose those whose predicted success is near $1/2$, since a group whose completions are all correct or all wrong produces no update. We reimplemented the GRPO trainer and ran $\nruns$ runs ($\gpuh$ GPU-hours) on six Qwen and Llama models. Because the gains we expected were small, comparable to the variation between random seeds, we \emph{preregistered} each study, i.e., fixed its hypotheses, primary comparisons, and equivalence margins in a written plan before running it. Configurations compared within a study share prompts and evaluation samples under each seed, so each seed yields a paired difference. Controls separate an intervention's intended effect from its side effects: random choice among the same candidate prompts (for a faster pass through the prompt stream), a baseline with a retuned learning rate (for an under-tuned baseline), and a strict verifier (for drift in what the reward accepts).

\textbf{Findings.} Our metric is the accuracy of sampled solutions to held-out MATH problems (pass@1), and we report differences in percentage points.
\begin{enumerate}[leftmargin=*,itemsep=0pt,topsep=0pt,label=(\arabic*)]
\item \emph{Matching compute reverses the ranking.} The Monte-Carlo reward leads outcome-only GRPO at equal updates but ends $14.3$ points behind at equal tokens and $7.8$ behind at equal training time, in every seed. Even given as much training time as outcome-only GRPO, none of the four dense-reward variants we tested shows a gain (\Cref{fig:time}a).
\item \emph{Gradient tests rank but cannot certify.} Per generated token, the efficiency test orders the Monte-Carlo variants as training does, yet it overestimates the accuracy reached with every reward we checked by one to two points, as large as the gains we measured.
\item \emph{Choosing which prompts to train on yields a gain that the protocol confirms and explains.} Training on the candidate prompts whose predicted success is nearest $1/2$ improves accuracy by $1.7$ points on Qwen3-0.6B and $0.8$ on Qwen2.5-1.5B at equal tokens, and by $1.0$ and $0.5$ at equal training time (exploratory). Both gains replicate on fresh seeds and on $1{,}543$ unseen MATH test problems, and random choice among the same candidates does not reproduce them. The gain appears where the candidates are predicted to fail more often than not, a rule that held in five of six preregistered tests, and outcome-only GRPO with twice the initial learning rate comes within half a point of it.
\item \emph{A lenient verifier changes what training optimizes.} With a verifier that also accepts answers outside \texttt{\textbackslash boxed\{\}}, runs gradually stopped boxing their answers. A strict verifier removed this drift without lowering accuracy.
\end{enumerate}

Our contributions are a matched-compute protocol for evaluating RL training interventions, whose trainer, preregistrations, and run records we release (\Cref{sec:setup}), and two case studies in which the protocol rejects one intervention and confirms another (\Cref{sec:dense,sec:selection}). We also characterize exactly when a dense reward's update on a fixed batch equals some reweighting of whole completions (\Cref{sec:theory}), and we draw lessons for validating training interventions (\Cref{sec:discussion}).

\section{Background and Protocol}
\label{sec:setup}

\subsection{GRPO and Credit}
\label{sec:grpo}

For a prompt $x$, GRPO samples a group of $G$ completions (rollouts) from the policy $\pi_\theta$, and a verifier gives each completion a reward $R_g\in\{0,1\}$, which is $1$ if its final answer is correct. Each completion's advantage is its reward relative to the rest of the group:
\begin{equation}
\small
\label{eq:adv}
A_g=\frac{R_g-\bar R}{\operatorname{std}(R)+\epsilon},\qquad
\bar R=\frac1G\sum_{g=1}^GR_g,\quad\epsilon>0.
\end{equation}
A policy-gradient update multiplies the log-probability gradient of each sampled token by a scalar weight, the token's \emph{credit} $c_{g,h}$:
\begin{equation}
\small
\label{eq:update}
u=\sum_{g=1}^G\sum_hc_{g,h}\,s_{g,h},
\end{equation}
where $s_{g,h}=\nabla_\theta\log\pi_\theta(a_{g,h}\mid x,a_{g,<h})$ is the log-probability gradient, or \emph{score}, of the $h$-th token $a_{g,h}$ of completion $g$. Outcome-only GRPO gives every token of a completion the same credit, $c_{g,h}=A_g$. Its update, $\gout=\sum_gA_gq_g$, is therefore a weighted sum of the completions' summed scores $q_g=\sum_hs_{g,h}$. We write $V=\Span\{q_1,\dots,q_G\}$ for the set of all such weighted sums, that is, of all updates that weight each completion as a whole. Any update in $V$ could also come from reweighting whole completions, so what a dense reward adds beyond such a reweighting lies outside $V$ (\Cref{sec:tests}). A group whose completions all succeed or all fail has $A_g=0$ for every $g$ and produces no update.

A \emph{dense reward} makes the credit vary within a completion. It adds a token-level term, $c_{g,h}=A_g+w_{g,h}$, that raises the credit of steps estimated to increase the chance of success and lowers the credit of steps estimated to decrease it. Its update can therefore leave $V$. Ratio clipping, KL penalties, and the optimizer can transform $u$.

\subsection{Interventions and What They Cost}
\label{sec:interventions}

We test two kinds of intervention (\Cref{fig:teaser}, \Cref{tab:coverage}): dense rewards, which change the credit $c_{g,h}$, and prompt selection, which changes which prompts are sampled. Their costs fall on generated tokens, on training time, or on both. \Cref{sec:dense} also tests a learned critic and two variants that reuse samples, and \Cref{sec:selection} tests simpler selection rules; we describe each where we test it.

\textbf{Monte-Carlo reward.} Following \citet{wang2023mathshepherd,kazemnejad2024vineppo}, we cut each completion at five fixed token positions, which splits it into six segments, and sample $k$ continuations from the prefix that ends at each cut. The fraction of continuations that end with a correct answer estimates the prefix's \emph{value}, i.e., the probability that it leads to a correct answer. Each segment's credit is the change in estimated value from its start to its end. The continuations are only scored and then discarded, so this reward costs generated tokens: at $k=4$, it uses $91\%$ of all generated tokens.

\textbf{Prompt selection.} Because uniform groups produce no update (\Cref{sec:grpo}), several methods aim training at prompts whose success rate is near $1/2$ \citep{gao2025pcl,qu2025mopps,sun2025dots}. In each training iteration, our \emph{critic filter} draws $24$ candidate prompts and predicts each one's success with a ridge critic on the policy's hidden state at the end of the prompt. It trains on the $6$ candidates whose predictions are nearest $1/2$, plus $2$ others chosen at random. Scoring the candidates takes one forward (prefill) pass and generates no tokens. As a control, \emph{random choice} trains on $8$ of the same $24$ candidates chosen at random.

\subsection{Matched-Compute Protocol}
\label{sec:protocol}

\textbf{Models and recipe.} Resolving small effects takes many paired seeds, so we train six small models: Qwen2.5-1.5B-Instruct \citep{qwen2.5,qwen2} with completions of up to $380$ tokens, and Qwen3 0.6B, 1.7B, and 4B in non-thinking mode \citep{yang2025qwen3} and Llama-3.2-1B- and 3B-Instruct \citep{grattafiori2024llama3} with completions of up to $768$ tokens. The training pool consists of MATH problems with numeric answers \citep{hendrycks2021math}, so that the verifier's answer check is well defined. It uses levels 4--5 in the dense-reward studies and levels 3--5 in the prompt-selection studies, except those that test level 5 or levels 1--2 alone. Each training iteration uses $8$ prompts with $8$ completions each and takes one on-policy Adam step at learning rate $10^{-6}$, without ratio clipping, sample reuse (except in the reuse variants), or a KL penalty.

\textbf{Two budgets.} Generated tokens and training time can rank methods differently (\Cref{sec:intro}), so we set both. The \emph{token budget} $B$ counts every token generated during training, including continuations and the groups that dynamic sampling discards; a run ends once it has generated $B$ tokens. A preregistered rule set $B$ for each model to the number of tokens that the $k=4$ Monte-Carlo reward generates in $48$ iterations ($B=12.25$M for Qwen2.5-1.5B; \Cref{app:m-setup} lists the others). Even with the most expensive reward, training thus makes about $50$ updates. \emph{Training time}, in GPU-hours, counts every training stage of a run: prompt scoring, generation, credit computation, the update, and the weight sync, but not evaluation or logging. To compare at equal training time, we compare the runs of one study, on one GPU type, at the time the faster configuration takes to reach $B$. We also report every comparison at an equal number of updates, the comparison we argue against.

\textbf{Metric.} We measure \emph{accuracy}, the quantity that training and the gradient tests target: the fraction of $8$ completions, sampled at temperature $1$, that the verifier marks correct (pass@1), averaged over $200$ held-out numeric-answer MATH test problems. We take it from the first evaluation at or after the budget $B$, without selecting checkpoints. We report differences in accuracy in percentage \emph{points}; figure axes show them as fractions ($0.01$ is one point).

\textbf{Preregistration and statistics.} Before its runs, each study fixed in a plan file its hypotheses, its primary contrasts, its Holm families, and its equivalence margin ($\pm2$ points, or $\pm1$ in four studies on Qwen2.5-1.5B). We report means with $95\%$ paired-$t$ intervals over seeds in brackets. We label each claim \emph{primary} (preregistered, with $p$-values adjusted for multiple comparisons by Holm's method), \emph{secondary} (preregistered, unadjusted), or \emph{exploratory} (not preregistered, marked \expl). To show that two configurations are equivalent within a margin, we use two one-sided tests (TOST). A crossed bootstrap over seeds and test problems checks every primary contrast. \Cref{tab:matched,tab:matched-b} list every contrast, and \Cref{app:matched} reports every secondary and exploratory analysis.

\textbf{Verifier.} The verifier defines the reward, so a lenient one can change what training optimizes. The default verifier credits the boxed answer, or the last number in the completion when no answer is boxed; the strict verifier credits only boxed answers. Runs trained with the default verifier can drift away from boxing their answers, which can bias a comparison in either direction. The strict verifier gives no reward for unboxed answers and so rules out this drift; studies marked ``strict'' train with it.

\subsection{Gradient Tests}
\label{sec:tests}

Gradient tests are a cheap way to vet a dense reward before training it (\Cref{sec:intro}), so we measure two and check in \Cref{sec:q2} whether they predict training. Both compare a method's update, on fixed batches at one policy, with an estimate $\widehat h$ of the gradient of expected accuracy from separately sampled continuations. \textbf{Alignment} is the cosine between the components of a dense update and of $\widehat h$ that lie outside $V$, the only part of a dense update that can improve, to first order, on reweighting whole completions. \textbf{Efficiency}, $(\E\langle u,\widehat h\rangle)^2/\E\|u\|^2$ for an update direction $u$, also accounts for how noisy an update is, and for a smooth objective it is proportional to the best expected gain of one SGD step along $u$. We compute it over $96$ fixed batches at the initial policy and report it relative to outcome-only GRPO, per update and per generated token; the per-token value divides by the tokens each update consumes.

\Cref{app:background} gives further details of the recipe, the statistics, the verifier, and the gradient tests.

\section{Case Study: Dense Rewards}
\label{sec:dense}

Of the interventions we test, dense rewards cost the most compute, and they are the ones that per-update comparisons and gradient tests favor. We examine them in three ways: per update, where they lead (\Cref{sec:equalround}); per generated token and per GPU-hour, where they lose (\Cref{sec:q1}); and through the gradient efficiency test, which anticipates the reversal but overestimates every reward it rates (\Cref{sec:q2}).

\subsection{Does a Dense Reward Help per Update?}
\label{sec:equalround}

\begin{figure*}[t]
\centering
\includegraphics[width=0.95\textwidth]{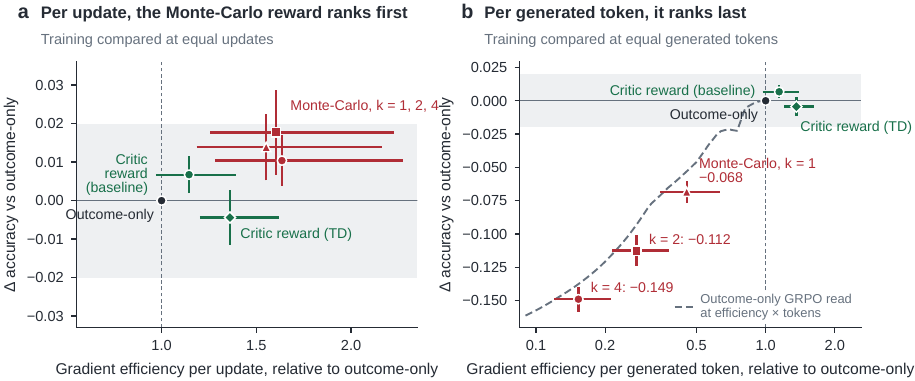}
\vspace{-14pt}
\caption{\textbf{The gradient test ranks the Monte-Carlo reward first per update and last per generated token, the same order training follows at matched tokens; the critics' higher per-token efficiency brings at most $0.7$ points.} Qwen2.5-1.5B on MATH L4--5. Horizontal: gradient efficiency relative to outcome-only GRPO (\Cref{sec:tests}), per update (a) or per generated token (b). Vertical: accuracy minus outcome-only GRPO on the same seeds at equal updates (a) or equal tokens (b). Red: the Monte-Carlo reward with $k=1$ (triangle), $2$ (square), and $4$ (circle) continuations per cut ($12$ fresh seeds); green: the two critic forms ($24$ fresh seeds). In (a), outcome-only GRPO is interpolated to each reward's update count (exploratory). Dashed line in (b): the heuristic extrapolation for a reward of that efficiency. Bars are $95\%$ CIs; the band marks $\pm0.02$.}
\label{fig:matched}
\end{figure*}

\claim{The gradient tests favor it.} We start with a per-update comparison, an equal-iteration study. It compares configurations after the same number of iterations and uses its own recipe: Qwen2.5-1.5B-Instruct, fully fine-tuned at learning rate $10^{-5}$ for $32$ iterations and scored by greedy validation accuracy (\Cref{app:equalround}). In that study, the Monte-Carlo reward is centered within each completion, normalized, and added with weight $\alpha$: $c_{g,h}=A_g+\alpha w_{g,h}$. On four full-parameter probes with disjoint rollouts, the component of its update outside $V$ has mean cosine $0.84$ with that of the reference gradient (\Cref{app:equalround}). To check that alignment is not automatic for any reward whose update leaves $V$, we include two control rewards, whose updates lie outside $V$ to a similar extent (the part outside $V$ carries $0.86$--$0.95$ of the update's norm for all three rewards) but do not align with the reference. One is an implicit reward $\log\pi_\theta-\log\pi_{\mathrm{ref}}$ in the style of PRIME; the other shuffles the Monte-Carlo reward within each completion. Their cosines are $-0.01\pm0.08$ and $-0.01\pm0.06$. In the matched-compute recipe, the $k=4$ Monte-Carlo reward is $1.56\times$ $[1.28,2.03]$ as efficient as outcome-only GRPO per update, and with the same number of updates it leads outcome-only GRPO by $1.5$ $[1.0,2.1]$ points\expl{} (\Cref{fig:matched}a).

\claim{Training does not follow, even at an equal number of iterations.} In the equal-iteration study, outcome-only GRPO improves greedy accuracy by $5.6$ $[4.1,7.1]$ points. In a prespecified expansion to $14$ paired seeds, the Monte-Carlo reward with a low weight ($\alpha=0.25$, $0.5$, or $1$) is equivalent to outcome-only GRPO within $\pm3$ points (Holm-corrected bootstrap and paired-$t$ TOST), and with the high weight $\alpha=5$ it loses $12.4$ $[8.7,16.1]$ points ($n=8$). Other recipes point the same way. On Qwen2.5-7B, no dense-reward configuration in a $43$-run sweep exceeds the outcome baseline's mean gain, SmolLM3-3B and a NuminaMath study show the same pattern, and only single-seed Phi-4 pilots are positive (\Cref{app:generality}).

\claim{Alignment stays high even as training fails.} In checkpoints taken every four iterations, the high-weight reward keeps an alignment of $0.969$ to $1.000$ with the reference on disjoint samples, while greedy accuracy falls from $25\%$ to $12\%$ and response length stays within $1\%$ of the low-weight runs. Entropy instead rises from $0.21$ to $0.58$ (low weight: $0.18$ to $0.20$), and the KL from the initial policy is about $30$ times that of the low-weight runs. Because alignment does not decay, the loss is not explained by the reward pointing the wrong way; the rising entropy and KL are consistent with over-optimizing it (\Cref{app:mechanism}).

\answer{The gradient tests say yes, and the Monte-Carlo reward leads at equal updates, but in the equal-iteration study it does not beat outcome-only GRPO.}

\subsection{Does a Dense Reward Pay for Its Compute?}
\label{sec:q1}

\claim{Per generated token, the Monte-Carlo reward is far behind.} At $k=4$, continuations take $91\%$ of the generated tokens, so by the time the token budget $B$ is used up, training with it has made $50$ updates and outcome-only GRPO $584$. Its accuracy at $B$ is $14.3$ $[12.7,16.0]$ points lower ($n=5$, Holm $p=5\times10^{-5}$), and $14.9$ $[14.0,15.8]$ points lower on twelve fresh seeds (\Cref{fig:matched}b). The deficit recurs on Qwen3 0.6B, 1.7B, and 4B ($8.0$, $9.1$, and $10.0$ points; $n=3$ each), on GSM8K ($9.8$), and with $768$-token completions ($6.6$, $n=8$).

\claim{Per GPU-hour, it is behind too.} Not all tokens cost the same. With the same generated tokens, training takes $0.14$ hours on an H200 with the $k=4$ reward and $0.43$ with outcome-only GRPO. Outcome-only GRPO spends $61\%$ of its time on updates, whereas the Monte-Carlo reward spends $73\%$ of its time generating continuations, which the inference engine generates in large batches (\Cref{app:time}). Even when it stops at the reward's training time, outcome-only GRPO leads by $7.8$ $[6.8,8.8]$ points, in all twelve seeds\expl.

\claim{Cheaper rewards lose less.} The randomness of continuation sampling contributes only $4\%$ of the second moment of the $k=4$ update, so using fewer continuations should cut the cost while adding little noise. A preregistered study tests $k=1$, $2$, and $4$ on twelve fresh seeds. With one continuation per cut, which still spends $70\%$ of the budget on continuations, the reward trails outcome-only GRPO by $6.8$ $[6.0,7.6]$ points and leads the $k=4$ reward by $8.1$ $[7.1,9.0]$ (both Holm $p<10^{-8}$, in all twelve seeds); $k=2$ falls in between ($-11.2$ $[-12.4,-10.1]$). Compared at their own training times, the $k=2$ and $k=1$ rewards trail by $5.1$ $[4.0,6.2]$ and $2.2$ $[1.7,2.8]$ points, again in every seed\expl. Fewer continuations narrow the gap but do not close it.

\claim{Reusing samples helps the reward but not the baseline.} Another way to lower the cost is to get more out of each sample. A VinePPO-style variant \citep{kazemnejad2024vineppo} uses value differences as advantages and, as in PPO, takes four clipped optimizer steps on each batch, and continuation reuse, adapted from SPO-tree and TreeRL \citep{guo2025spo,hou2025treerl}, also trains on the continuations instead of discarding them (\Cref{tab:adapt}). In three-seed pilots, this reuse raises the $k=4$ reward's accuracy at matched tokens by $7.7$ points (Holm $p=.007$) but lowers outcome-only GRPO's by $5.9$ (\Cref{app:pilots}).

\begin{figure*}[t]
\centering
\includegraphics[width=0.8\textwidth]{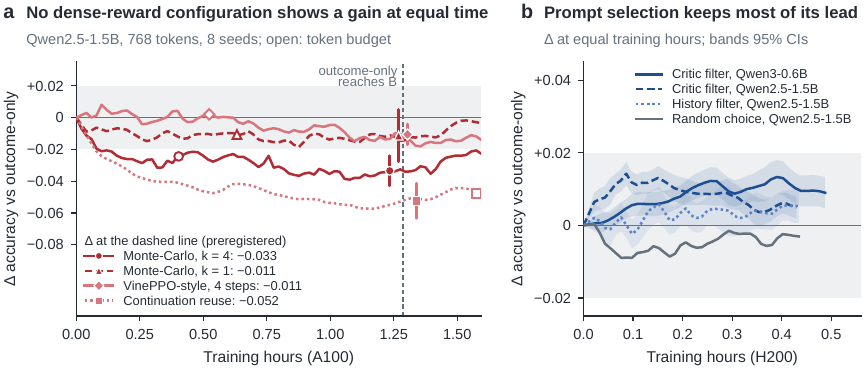}
\vspace{-6pt}
\caption{\textbf{At equal training time, no tested dense-reward variant showed a gain, and prompt selection keeps most of its lead.} \textbf{(a)} A preregistered study ran every configuration beyond the training time that outcome-only GRPO needs to reach the token budget $B$ (dashed line). Curves show each configuration minus outcome-only GRPO at equal training hours (MATH L4--5, mean over seeds); open markers show where each configuration reaches $B$. Bars give the preregistered contrasts at the dashed line with $95\%$ CIs, and the legend gives their means. \textbf{(b)} The prompt filters, and random choice among the filter's candidates, minus outcome-only GRPO at equal training hours, up to the time the faster configuration takes to reach the budget (exploratory, H200 GPUs). Bands are $95\%$ CIs; the gray bands mark $\pm0.02$.}
\label{fig:time}
\end{figure*}

\claim{Even with as much training time as outcome-only GRPO, no variant shows a significant gain.} The comparisons above stop when the faster configuration reaches $B$. A preregistered study with $768$-token completions therefore trains four variants beyond $T_c$, the training time that outcome-only GRPO needs to reach $B$ ($1.29$ A100 GPU-hours; $8$ seeds, strict verifier), and compares each with outcome-only GRPO at $T_c$ (Holm over the four; \Cref{fig:time}a, \Cref{app:p39}). Two variants trail in every seed: the $k=4$ Monte-Carlo reward by $3.3$ $[2.4,4.3]$ points (Holm $p=2\times10^{-4}$), and continuation reuse by $5.2$ $[4.1,6.3]$ ($p<10^{-4}$), because its updates, which include the continuations, take $78\%$ of its training time ($45$ seconds per iteration, against $6$ for outcome-only GRPO). The VinePPO-style variant trails by $1.1$ $[0.4,1.7]$ ($p=.013$). This deficit lies within $\pm2$ points (TOST $p=.005$), its crossed interval reaches zero, and at the variant's own, shorter training time it is even with outcome-only GRPO ($+0.3$, secondary). The $k=1$ reward's deficit, $1.1$ $[-0.5,+2.8]$ ($p=.15$), is neither significant nor shown to lie within $\pm2$. Given still more time, the $k=1$ reward reaches outcome-only GRPO's accuracy at $B$ after $1.25\,T_c$ on the mean curves, and the other three do not within their runs, which end at $1.27$ to $2.5\,T_c$\expl{} (\Cref{app:p39}).

\answer{No. Per generated token, every Monte-Carlo variant loses $3$ to $15$ points. Given as much training time as outcome-only GRPO, none outperforms it, and the two closest trail by $1.1$ points.}

\subsection{Can a Gradient Test Predict the Result?}
\label{sec:q2}

\claim{Per generated token, the efficiency test anticipates the reversal.} The $k=4$ reward is $1.56\times$ as efficient as outcome-only GRPO per update but only $0.145\times$ $[0.120,0.190]$ per generated token. Per token, the test ranks the three Monte-Carlo rewards as $k=1$ ($0.45\times$), $k=2$ ($0.27\times$), and $k=4$ ($0.15\times$), the same order as training (\Cref{fig:matched}b). We had preregistered this order, together with the prediction that every Monte-Carlo reward would trail outcome-only GRPO, and both predictions held. Because the three are about equally efficient per update ($1.55$ to $1.63\times$), this order follows their token costs, and what the test adds is the size of each deficit.

\claim{But it overestimates every reward it rates.} Whether a reward is worth training depends on the size of its effect, not only its rank. To turn efficiency into a predicted accuracy, we use a heuristic extrapolation, not a derived prediction: we suppose that a reward with per-token efficiency $e$ reaches, after $t$ tokens, the accuracy that outcome-only GRPO reaches after $et$ tokens. This predicts $35.9\%$, $31.9\%$, and $27.5\%$ accuracy at $B$ for $k=1$, $2$, and $4$; the rewards actually reach $34.3\%$, $29.9\%$, and $26.3\%$\expl. A learned critic tests the extrapolation where it is most favorable to dense rewards. Instead of sampling continuations, which dominate the cost of the Monte-Carlo reward, it predicts each prefix's value with a ridge regression on the policy's hidden states, refit each iteration on earlier rollouts; its forward passes and fits still add training time. Its temporal-difference (TD) form credits each segment with the change in predicted value, and its baseline form subtracts the predicted value from the final reward. Per token, the critic is $1.36\times$ $[1.20,1.62]$ as efficient as outcome-only GRPO in its TD form and $1.15\times$ $[0.97,1.39]$ in its baseline form. On $24$ fresh Qwen2.5-1.5B seeds, both forms are equivalent to outcome-only GRPO within $\pm2$ points (TOST, Holm $p<10^{-4}$). The TD form changes accuracy by $-0.4$ $[-1.2,+0.3]$ points, and the baseline form ends $0.7$ $[0.2,1.2]$ points ahead (Holm $p=.016$), a lead that appears only at the final evaluation. Fitting the critic adds $23$ to $24\%$ to training time, so at equal training time both forms trail outcome-only GRPO, by $1.2$ $[0.7,1.7]$ and $0.7$ $[0.1,1.2]$ points\expl. Where the extrapolation predicts average leads of $2.0$ and $0.9$ points, the critics lead by $0.2$ and $0.0$, shortfalls we had preregistered ($p<10^{-5}$ each). Across all five rewards, the extrapolation overestimates accuracy by $0.8$ to $1.8$ points on average, and every interval excludes zero (exploratory for the Monte-Carlo rewards). The test therefore cannot certify a gain of a point or two. It describes a single SGD step, with its step size tuned to each update, at the initial policy, whereas training takes many Adam steps at a fixed learning rate as the policy changes.

\answer{It predicts the order, not the size. The test ranks the Monte-Carlo rewards as training does but overestimates all five rewards by $0.8$ to $1.8$ points, as large as the gains we measured.}

\section{Case Study: Prompt Selection}
\label{sec:selection}

Dense rewards show what the protocol rejects; prompt selection shows what it confirms. Rather than spending generated tokens to improve the update on prompts already sampled, prompt selection spends one forward pass to decide which prompts to sample (\Cref{sec:interventions}). The protocol confirms its gain, shows where the gain comes from (\Cref{sec:q3}), predicts where it appears (\Cref{sec:q4}), and measures what it is worth (\Cref{sec:q5}).

\subsection{Does Prompt Selection Help?}
\label{sec:q3}

\begin{figure*}[t]
\centering
\includegraphics[width=0.86\textwidth]{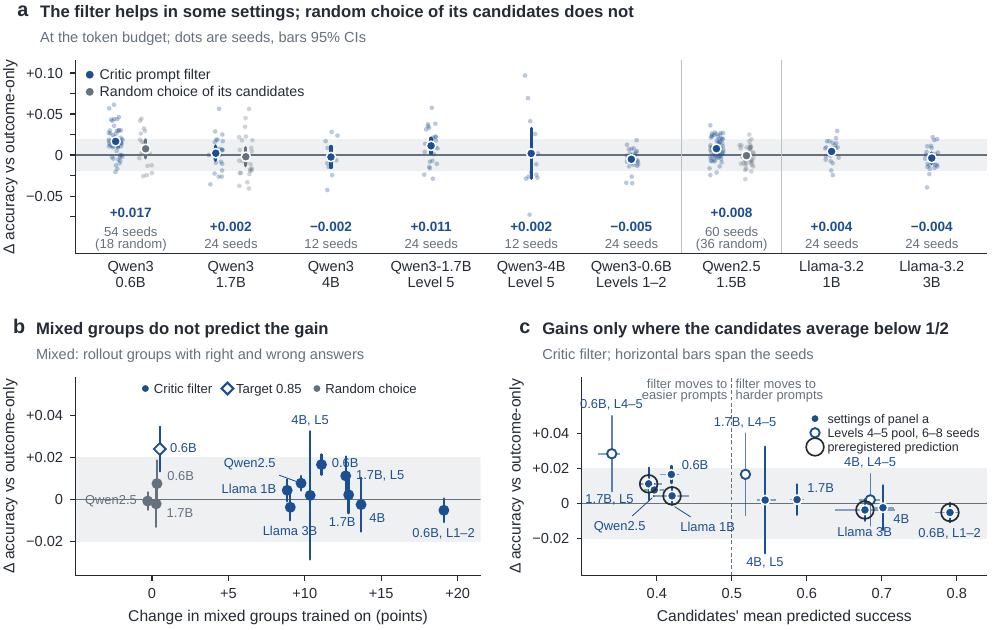}
\vspace{-6pt}
\caption{\textbf{Where prompt selection helps is predicted by how hard its candidates are, not by how many informative groups it adds.} At matched generated tokens: \textbf{(a)} Accuracy at the budget minus outcome-only GRPO on the same seeds, for the critic filter and for random choice of $8$ of its $24$ candidates (dots: seeds; bold: the filter's mean). The pool is MATH levels 3--5 unless labeled; Llama models train on the strict verifier. \textbf{(b)} Change in the share of mixed-outcome groups trained on before $B$ against the change in accuracy, both relative to outcome-only GRPO; the diamond is the filter aimed at predicted success $0.85$. \textbf{(c)} The critic's mean predicted success of the candidates against the change in accuracy; the dashed line marks $1/2$. Hollow points are the first studies, on a level 4--5 pool, and rings mark the four settings with a preregistered prediction. Bars are $95\%$ CIs over seeds; the gray band marks $\pm0.02$.}
\label{fig:selection}
\end{figure*}

\claim{It gains on two of six models with the level 3--5 pool, and both gains replicate.} On Qwen3-0.6B, the critic filter raises accuracy by $1.7$ $[1.2,2.1]$ points over $54$ seeds from four studies (and is higher in $42$ of them), including a gain of $1.2$ $[0.5,1.9]$ points in a preregistered replication on $24$ fresh seeds (Holm $p=.003$). On Qwen2.5-1.5B, the gain is smaller but holds in three preregistered studies ($12$, $24$, and $24$ seeds; Holm $p=.039$, $.006$, and $.034$); over all $60$ seeds, it is $0.8$ $[0.4,1.1]$ points. The other four models show no significant gain on this pool (\Cref{fig:selection}a), and \Cref{sec:q4} explains where the filter helps.

\claim{It carries over to problems that no study had evaluated on.} Every study evaluated on the same $200$ test problems, and on them the crossed bootstrap interval for the Qwen2.5-1.5B gain reached zero each time, so we tested transfer directly. The last study on each model also scored, at $B$, the $1{,}543$ numeric-answer MATH test problems of levels 3--5 that lie outside all evaluation sets, and the $1{,}319$ GSM8K test problems. On these MATH problems, the filter gains $0.8$ $[0.5,1.1]$ points on Qwen2.5-1.5B and $1.1$ $[0.6,1.5]$ on Qwen3-0.6B (Holm $p<10^{-4}$ each), with crossed intervals excluding zero. On GSM8K, which neither configuration trains on, it gains $1.1$ and $1.3$ points (secondary).

\claim{It comes from which prompts are chosen.} Because it draws $24$ candidates per iteration and trains on $8$, the filter also moves through the prompt stream three times as fast as outcome-only GRPO, which could help on its own. Random choice among the same candidates isolates this side effect. In preregistered primary contrasts, random choice trails the filter on both models where the filter helps: by $1.7$ $[0.5,2.9]$ points on Qwen3-0.6B ($p=.008$, twelve fresh seeds), and on Qwen2.5-1.5B by $1.0$ $[0.4,1.6]$ (Holm $p=.011$) and again by $0.9$ $[0.3,1.5]$ on $24$ fresh seeds (Holm $p=.007$). Random choice itself is equivalent to outcome-only GRPO within $\pm2$ points on all three models tested.

\claim{Most of it survives when training time is counted.} Scoring the candidates costs a forward pass, so the filter's runs take $9$ to $14\%$ longer to reach $B$. At equal training time, the filter still leads, by $1.0$ $[0.6,1.5]$ points on Qwen3-0.6B and $0.5$ $[0.2,0.8]$ on Qwen2.5-1.5B\expl{} (\Cref{fig:time}b); on the seeds of the Qwen3-0.6B replication alone, this lead is $0.4$ $[-0.1,+1.0]$.

\claim{A per-prompt success estimate carries the gain, and cheaper signals do not.} Simpler rules test which part of the filter carries the gain. A history filter, close to GRESO \citep{zheng2025greso}, uses the reward of each prompt's most recent group and needs no forward pass. On Qwen2.5-1.5B, it gains $0.6$ $[0.1,1.1]$ points (Holm $p=.035$, $36$ seeds), also at equal training time, while the learned critic adds an early lead of $1.9$ $[1.6,2.1]$ points over it during the first half of the budget (secondary). On Qwen3-0.6B and 1.7B, filtering by MATH difficulty level gains nothing ($+0.2$, TOST $p=.005$), and DAPO's dynamic sampling \citep{yu2025dapo} loses $1.8$ $[0.8,2.7]$ points, because the groups it discards after generating them count toward the budget.

\answer{Yes, on two of six models: by $1.7$ and $0.8$ points at equal tokens and by $1.0$ and $0.5$ at equal training time (exploratory). The gains replicate on fresh seeds and unseen problems and come from the choice of prompts.}

\subsection{When Does Prompt Selection Help?}
\label{sec:q4}

\claim{Not where it adds the most informative groups.} The rationale for training on prompts near $1/2$, and for dynamic sampling \citep{gao2025pcl,yu2025dapo}, is that only mixed-outcome groups have nonzero advantages. The filter raises the share of these groups by $8$ to $19$ points in every setting at learning rate $10^{-6}$. Accuracy, however, rises in only some settings, and not in those where the share rises most (\Cref{fig:selection}b), so this rationale does not predict it.

\claim{It helps where the candidates are predicted to fail more often than not.} We formed this account post hoc from five settings. In them, the critic's mean predicted success over the $24$ candidates separates the two settings where the filter helped ($0.42$ and $0.40$) from the rest ($0.54$ to $0.70$), with no overlap between seeds, and the predictions are calibrated (\Cref{tab:calib}). Random choice, which trains on a uniform draw of the candidates, realizes the predicted success to within $0.2$ points on all three models. The rule follows from how the filter chooses prompts. When the mean prediction is below $1/2$, keeping the candidates nearest $1/2$ shifts training toward easier prompts and away from groups that are all wrong; when it is above $1/2$, the filter shifts training toward harder prompts. Outcome-only GRPO's own training reward separates the first eleven settings in the same way (\Cref{tab:account}), so a short pilot run is enough to apply the rule.

\begin{table}[t]
\vspace{-1ex}
\centering
\caption{\textbf{Preregistered tests of the account that prompt selection helps when the candidates' mean predicted success (Cand.) is below $1/2$.} Result: accuracy at $B$ minus outcome-only GRPO, in points ($24$ seeds, $95\%$ CIs), except $^a$the gain on levels 1--2 minus the gain on levels 3--5 ($24$ and $18$ seeds) and $^b$the filter aimed at $p^\star=0.25$ minus the filter aimed at $1/2$. L5, L1--2: level-5 and level 1--2 pools. A check marks a prediction that held.}
\label{tab:tests}
\footnotesize
\setlength{\tabcolsep}{2.5pt}
\resizebox{\columnwidth}{!}{%
\begin{tabular}{@{}lcll@{}}
\toprule
Setting & Cand. & Predicted & Result \\
\midrule
Qwen3-1.7B, L5 & $0.39$ & gain \textcolor{pvGreen}{\checkmark} & $+1.1\,[+0.2,+2.0]$ \\
Qwen3-0.6B, L1--2 & $0.79$ & none \textcolor{pvGreen}{\checkmark} & $-3.0\,[-3.9,-2.0]^a$ \\
Llama-3.2-3B & $0.68$ & none \textcolor{pvGreen}{\checkmark} & $-0.4\,[-1.0,+0.2]$ \\
Llama-3.2-1B & $0.42$ & gain \textcolor{pvRed}{$\times$} & $+0.4\,[-0.1,+1.0]$ \\
Qwen3-0.6B, $p^\star{=}0.25$ & $0.42$ & loss \textcolor{pvGreen}{\checkmark} & $-2.8\,[-3.6,-2.0]^b$ \\
Qwen2.5-1.5B, $p^\star{=}0.25$ & $0.40$ & loss \textcolor{pvGreen}{\checkmark} & $-1.0\,[-1.6,-0.4]^b$ \\
\bottomrule
\end{tabular}}
\end{table}

\claim{Preregistered tests confirm the account, including a direct manipulation.} Five of six preregistered tests agree with it (\Cref{tab:tests}). The two most direct tests hold the model and pool fixed and change only the filter's target, to predicted success $0.25$, so that the filter moves training toward prompts harder than its candidates: on both models where the gain had replicated, it disappears. Two tests change the training pool: the filter helps on Qwen3-1.7B with level 5 (mean prediction $0.39$), although its crossed interval reaches zero, and not on Qwen3-0.6B with levels 1--2 ($0.79$). Two change the model family: it does not help on Llama-3.2-3B ($0.68$), as predicted, nor on Llama-3.2-1B ($0.42$), the one failed prediction. Because four of the five confirmed predictions are of no gain, the more common outcome, the two direct manipulations weigh most (\Cref{app:limitations}).

\answer{Where candidates are predicted to fail more often than not. The filter helps significantly in four of the five such settings and in none of the seven others (\Cref{fig:selection}c); on a fixed model and pool, aiming it at harder prompts removes the gain.}

\begin{figure*}[t]
\centering
\includegraphics[width=0.88\textwidth]{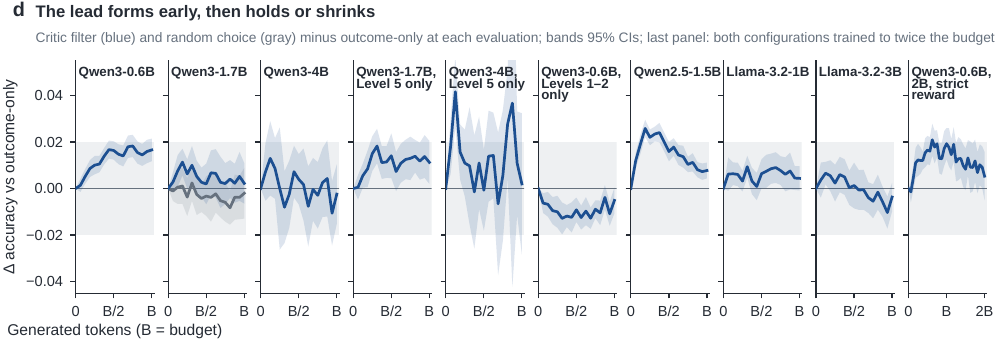}
\caption{\textbf{The prompt filter leads early, then holds or shrinks.} \textbf{(d)} The differences of \Cref{fig:selection}a at every evaluation: the critic filter (blue) and random choice (gray) minus outcome-only GRPO on the same seeds. In the last panel, Qwen3-0.6B runs trained for twice the budget with the strict verifier as the reward of both configurations ($24$ seeds). Bands are $95\%$ CIs over seeds; the gray band marks $\pm0.02$.}
\label{fig:selection-time}
\end{figure*}

\begin{figure*}[t]
\centering
\includegraphics[width=0.88\textwidth]{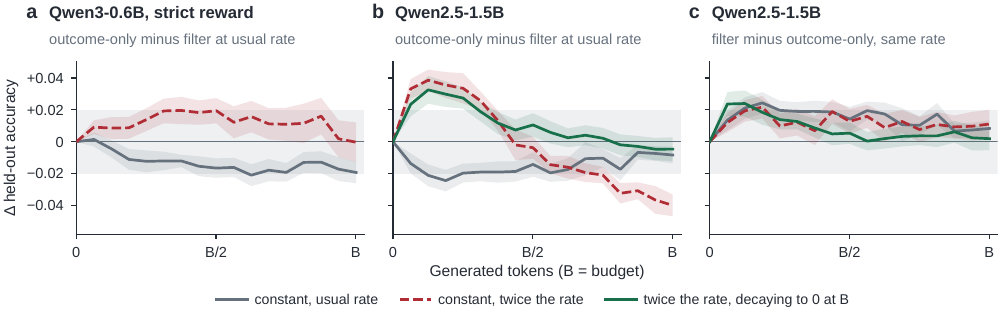}
\vspace{-14pt}
\caption{\textbf{A retuned learning rate brings outcome-only GRPO within half a point of the critic filter in mean accuracy.} Accuracy at every evaluation, paired by seed (preregistered studies on the level 3--5 pool, $24$ seeds each). \textbf{(a, b)} Outcome-only GRPO at each learning rate minus the filter at the default constant rate $10^{-6}$; on Qwen3-0.6B every configuration trains on the strict verifier. \textbf{(c)} On Qwen2.5-1.5B, the filter minus outcome-only GRPO at the same rate: under the decaying rate its gain fades by the budget. Bands are $95\%$ CIs; the gray band marks $\pm0.02$.}
\label{fig:lr}
\end{figure*}

\subsection{What Is the Gain Worth?}
\label{sec:q5}

\claim{It is a head start that saves about a fifth of the token budget.} The lead forms early: on Qwen2.5-1.5B, it peaks at $2.6$ $[2.2,3.0]$ points at three-sixteenths of the budget ($3B/16$) and falls to $0.8$ at $B$ (\Cref{fig:selection-time}). On both models, the filter reaches outcome-only GRPO's final accuracy with $0.78$ of the token budget, and with $0.85$ and $0.87$ of its training time, including the time spent scoring candidates\expl. On Qwen3-0.6B, the gain also holds at pass@8, by $0.9$ $[0.2,1.7]$ points, and at every test level, whereas on Qwen2.5-1.5B it shrinks as $k$ grows\expl{} (\Cref{tab:passk}). With longer training, outcome-only GRPO catches up. A preregistered study trained both Qwen3-0.6B configurations to twice the budget (strict verifier, $24$ seeds): the filter leads by $1.9$ $[1.3,2.6]$ points at $B$ but by only $0.5$ $[-0.5,+1.6]$ at $2B$, which is equivalent to zero within $\pm2$ points (TOST $p=.004$).

\claim{A retuned baseline gains about as much, so the baseline sets the benchmark.} A head start could also reflect a baseline that learns too slowly, so we retuned its learning rate. On Qwen3-0.6B, a preregistered study (strict verifier, $24$ seeds) doubled the constant learning rate of $10^{-6}$, which raised outcome-only GRPO's accuracy by $1.9$ $[0.5,3.3]$ points (Holm $p=.018$); the faster baseline minus the filter is $-0.03$ $[-1.3,+1.2]$ points, equivalent within $\pm2$ (TOST $p=.002$; \Cref{fig:lr}a). On Qwen2.5-1.5B, a doubled rate lowers both by about three points, and the filter's gain holds ($1.1$, Holm $p=.018$). A rate that decays linearly from $2\times10^{-6}$ to $0$ at $B$ keeps the faster early rise without that loss (\Cref{fig:lr}b). With it, outcome-only GRPO ends $0.5$ $[-0.3,+1.2]$ points below the filter at the constant rate (Holm $p=.41$; equivalent within $\pm2$ but not shown within $\pm1$), and the filter's own gain under the schedule, $0.2$ $[-0.5,+0.9]$, is equivalent to zero within $\pm1$ point (TOST $p=.015$; \Cref{fig:lr}c).

\answer{A head start worth about a fifth of the budget. It fades by $2B$, and a retuned baseline comes within half a point of it.}

\section{Related Work}
\label{sec:related}

\textbf{The structure of policy-gradient updates.} Writing an outcome update as a weighted sum of the completions' summed scores follows from the score-function estimator \citep{williams1992reinforce,sutton1999pg}, and it applies to the completion-level advantages of GRPO and RLOO \citep{shao2024deepseekmath,ahmadian2024rloo}; \citet{mishra2026foundations} study this factorization for GRPO. Analyses of GRPO's dynamics in expectation \citep{mroueh2026grpo} and of which credit schemes give proportional expected gradients \citep{fu2026pact} leave open whether the updates computed on a finite batch are equal, which \Cref{sec:theory} answers exactly. The low-dimensional gradient subspaces measured by \citet{schneider2024subspaces} also differ from the span $V$ of one batch's completion scores. HAPO and EAPO redistribute advantages within completions while preserving their signs \citep{he2026hapo,yeo2026eapo}, so their updates, too, can leave $V$. The sampled gradients of potential-based shaping \citep{ng1999shaping} need not lie in $V$ either, and work on reward hacking \citep{skalse2022hacking,gao2023overoptimization} shows that optimizing a proxy reward too hard can lower the true objective.

\textbf{Process rewards and how they are combined with outcomes.} Step-level supervision \citep{lightman2023verify,wang2023mathshepherd}, progress-based advantages \citep{setlur2024pav}, Monte-Carlo value estimates \citep{kazemnejad2024vineppo}, and implicit reward models trained online from outcome labels \citep{cui2024implicitprm,cui2025prime} report gains from feedback on intermediate steps. VinePPO reaches PPO's peak accuracy in up to $3\times$ less wall-clock time and outperforms GRPO and RLOO, with $9$ rollouts at every reasoning step, models fine-tuned on the task, and KL-regularized PPO steps; our adaptation keeps its value-difference advantage and clipped reuse but not its recipe (\Cref{tab:adapt}). SPO-tree \citep{guo2025spo} and TreeRL \citep{hou2025treerl} also train on the sampled continuations, so that their tokens are not spent on value estimates alone. How advantages are normalized, the level at which credit is assigned, and how uniform groups are handled affect how process and outcome rewards combine \citep{wang2026pass,agrawal2026verigate}.

\textbf{Choosing training prompts.} Because a uniform group gives GRPO no advantage, several methods aim training at prompts with success near $1/2$. Some measure success with rollouts first: DAPO discards uniform groups after generating them \citep{yu2025dapo}, online difficulty filtering keeps prompts whose pass rates fall in a band around $1/2$ \citep{bae2025online}, and LILO ranks prompts by learnability \citep{foster2025learnability}. Others decide before generating: GRESO skips prompts whose recent groups were uniform \citep{zheng2025greso}, MoPPS keeps a Bayesian estimate of each prompt's success \citep{qu2025mopps}, DOTS predicts difficulty from similar questions \citep{sun2025dots}, and PCL trains on the prompts whose value-model prediction is nearest $1/2$ \citep{gao2025pcl}. Our critic filter uses PCL's rule and our history filter resembles GRESO and MoPPS; rather than propose a new rule, we test such rules at matched compute.

\section{Lessons and Conclusion}
\label{sec:discussion}

Each lesson comes from a result that a cheaper protocol would have missed. (1)~\emph{Count every generated token and every GPU-hour, and report updates alongside them.} The Monte-Carlo reward ranked first at equal updates and last at equal tokens, and a critic's small gain at equal tokens disappeared at equal training time. (2)~\emph{Treat gradient tests as heuristics, not certificates.} The efficiency test overestimated every reward it rated by one to two points, as large as the gains we measured. (3)~\emph{Control for side effects.} Random choice among the same candidates separated which prompts were chosen from how fast training moved through the prompt stream. (4)~\emph{Test the mechanism, not only the method.} The share of informative groups did not predict where prompt selection helped; the candidates' predicted success did, in five of six preregistered tests. (5)~\emph{Retune the baseline.} Doubling the initial learning rate brought outcome-only GRPO within half a point of the critic filter on both models where its gain replicated. (6)~\emph{Fix the verifier first.} A lenient verifier let runs in either configuration drift away from boxing their answers. (7)~\emph{Replicate on fresh seeds and on more than one model, and preregister.} A six-seed gain on Qwen3-1.7B did not replicate on eighteen fresh seeds, and the one gain we predicted outside the Qwen family did not appear. \Cref{app:limitations} discusses the scope of these lessons and the limitations of our studies.

Dense rewards can produce update directions that no reweighting of whole completions can, but only training at matched compute can show whether they help. None of our dense-reward variants beat outcome-only GRPO by more than a point, and the Monte-Carlo reward fell $14.3$ points behind at equal tokens, while choosing which prompts to train on beat it by up to $1.7$ points. Because that gain comes early, a faster-learning baseline nearly matches it: outcome-only GRPO with twice the initial learning rate comes within half a point. A claim that a training intervention helps should state its token and GPU-hour budgets, its verifier, the models on which it replicated, and how its baseline was tuned.

\label{page:mainbodyend}
\label{page:references}

\bibliographystyle{mlsys2025}
\bibliography{references-1}

\section*{AI Usage Disclosure}
We used generative AI tools for grammar check, spelling mistakes and help us to visualize our results efficiently as well as debugging help during codebase development.

\clearpage
\appendix
\label{page:supplement}
\section*{Supplementary Material}
\section*{Reproducibility and Responsible Reporting}
The supplementary material gives the proofs, training and evaluation settings, seed counts, and statistical methods. The artifact at {\urlstyle{same}\url{https://github.com/vicky157/grpo-validation}} contains the trainer and analysis code, every configuration and preregistration file with its dated amendments, each study's report and submission record, and every run's configuration, evaluation log (with training time), seconds per training stage, and compute record, from which every contrast at matched tokens or matched training time can be recomputed; one script rebuilds the training-time tables and figures. The per-iteration logs are omitted for size. We use public mathematical-reasoning benchmarks and released language-model checkpoints. The study involves no human participants or new personal-data collection. We report accuracy losses and inconclusive comparisons alongside positive results.

\textbf{Conventions.} The appendix uses the terms defined in \Cref{sec:setup}, and reports accuracy and differences in accuracy as fractions, as the figures do ($0.01$ is one percentage point of the main text).

\section{Additional Background}
\label{app:background}

This section gives the details of the protocol and of the gradient tests that \Cref{sec:setup} summarizes.

\textbf{Recipe.} A preregistered stability rule set the learning rate to $10^{-6}$, after $10^{-5}$ degraded every configuration over the matched horizon. The completion limit matters: at $380$ tokens, $73\%$ of the initial samples are cut off, so part of what every run learns is to finish within the limit; at $768$ tokens, $20\%$ are. Accuracy is averaged over held-out test problems of levels 3--5, or of the training pool's own levels for the level-5 and level 1--2 pools. Both reuse variants keep the published form of credit but not the published step boundaries, sample counts, or recipes (\Cref{tab:adapt}).

\textbf{Statistics.} A study's Holm families are the groups of tests corrected together. The plan files are internal rather than in a public registry; \Cref{app:matched} audits their timing. Because every study evaluates on the same $200$ test problems, a crossed bootstrap, which resamples seeds and test problems together, checks every primary contrast.

\textbf{Verifier drift.} Some runs trained with the default verifier gradually stop boxing their answers, which shows up as accuracy under the default verifier exceeding that under the strict one by more than $10$ points. This drift occurred in up to $49$ of a study's $408$ evaluations and in either configuration. Preregistered studies that retrained the drifting configurations with the strict verifier found no drift ($0$ of $2{,}196$ evaluations) and no loss in accuracy (\Cref{app:m-verifier}).

\textbf{Alignment.} Let $J(\theta)$ be the expected accuracy of $\pi_\theta$, $h=\nabla J(\theta)$ its gradient, and $\widehat h$ an estimate of $h$ from separately sampled continuations. Every update that weights whole completions lies in $V$. A dense update $d$ can therefore improve on its projection $\PV d$ onto $V$, to first order, only through its component outside $V$:
\begin{equation}
\label{eq:projection-rate}
\langle h,d-\PV d\rangle=\langle\Pperp h,\Pperp d\rangle,\qquad \Pperp=I-\PV
\end{equation}
(\Cref{thm:nns}; \Cref{sec:theory} characterizes exactly which dense rewards produce updates outside $V$). The alignment test reports the cosine between $\Pperp d$ and $\Pperp\widehat h$, the components of the dense update and of the estimated gradient that lie outside $V$.

\textbf{Efficiency.} Alignment ignores how noisy an update is and what it costs; efficiency accounts for both. For an objective with smoothness constant $L$, the gradient efficiency $(\E\langle u,\widehat h\rangle)^2/\E\|u\|^2$ of an update direction $u$ equals $2L$ times the best expected one-step gain of an SGD step along $u$, $\max_\eta\,\eta\E\langle u,\widehat h\rangle-\tfrac L2\eta^2\E\|u\|^2$; the expected inner product $\E\langle u,\widehat h\rangle$ was positive for every method we measured (\Cref{app:matched}).

\section{Scope and Open Validation Questions}
\label{app:limitations}

This section expands on the limitations in \Cref{sec:discussion}. It first covers the matched-compute studies of \Cref{sec:dense,sec:selection}, and then the equal-iteration study of \Cref{sec:equalround} and the fixed-batch theory of \Cref{sec:theory}.

\subsection{Matched-compute studies}

\textbf{Scope.} These studies use an independent reimplementation of the trainer and a learning rate of $10^{-6}$, rather than the equal-iteration study's $10^{-5}$, because the faster rate degraded every configuration over the matched horizon. They cover numeric-answer MATH problems, evaluation sets of $200$ problems (on Qwen2.5-1.5B and Qwen3-0.6B, also $1{,}543$ further MATH test problems and the GSM8K test set), models of up to $4$B parameters, and budgets of $8$--$35$M generated tokens. Most contrasts pool $6$--$60$ seed pairs, but some, including primary contrasts of the first studies, have only $n=3$--$5$. Four of the six models are from the Qwen family, in which RLVR can behave differently from other families \citep{shao2025spurious}. We tested the other two, Llama-3.2-1B and 3B, only with prompt selection and on one pool. On 3B, the candidates' mean predicted success was above $1/2$. On 1B it was below $1/2$, but the predicted gain did not appear, so we have shown no gain outside the Qwen family.

\textbf{Learning rate.} Most runs use a constant learning rate of $10^{-6}$. Three studies double it, and one decays it linearly from $2\times10^{-6}$ to zero over the budget; we did not tune it for each model or configuration. On Qwen3-0.6B, doubling the rate gave outcome-only GRPO as large a gain as the filter, also without drift, so a baseline tuned for speed may leave less for prompt selection to add. On Qwen2.5-1.5B, doubling the rate lowered both configurations at the budget, and the filter's gain held. With the decaying rate, however, outcome-only GRPO ended $0.005$ below the filter at $10^{-6}$, a difference that is neither significant nor shown to lie within $\pm0.01$, and the filter's gain under that schedule was equivalent to zero within $\pm0.01$, a margin wider than the gain at the constant rate. We tried no warmup, no other peak rates, and no other schedules, and none of the schedules we tried was tuned for the filter. A schedule tuned for the filter might restore its gain, and one tuned for the baseline might close the remaining half point.

\textbf{Where prompt selection helps.} Prompt selection helped Qwen3-0.6B (on levels 3--5 and 4--5), Qwen2.5-1.5B on levels 3--5, and Qwen3-1.7B on level 5. It did not help Qwen3-1.7B on levels 3--5, Qwen3-4B, Qwen3-0.6B on levels 1--2, or Llama-3.2-3B, and it did not help Llama-3.2-1B significantly. We formed the account that it helps when the candidates' mean predicted success is below $1/2$ post hoc, from five settings.

\textbf{Tests of the account.} Two preregistered studies agreed with the account, one in each direction: they found a small gain ($+0.011$) whose crossed interval reaches zero, and no gain where the candidates averaged $0.79$. Both moved the candidates across $1/2$ by changing the training pool, which also changes the test set, so they do not separate the candidates' difficulty from other properties of a pool. A third study kept the pool and changed the model. It agreed with the prediction of no gain, but the candidates fell on the other side of $1/2$ from the one its design anticipated. Two of these three predictions were of no gain, which is the more common outcome here (seven of twelve settings), so they put the account at less risk than a predicted gain does. A fourth study tested a predicted gain on the smaller Llama-3.2-1B with the level 3--5 pool, chosen by a pilot because outcome-only GRPO's training reward on it was below $0.45$. That prediction failed ($+0.004$ $[-0.001,+0.010]$, $p=.095$, equivalent to zero within $\pm0.02$), although the filter's mean lead was positive at every evaluation after the start. Of the account's two predicted gains, therefore, one was found. A fifth study kept the model and pool fixed (Qwen3-0.6B on levels 3--5) and changed only the filter's target, to $0.25$. As predicted, the filter then trained on prompts harder than its candidates and lost its gain. A sixth study repeated this on Qwen2.5-1.5B with the same pool and found the same result, although the deficit there ($-0.010$, as large as the gain with target $1/2$) has a crossed interval that just reaches zero, and the filter kept a small head start that its preregistration did not predict. These two studies separate the direction in which the filter moves training from the pool, but only on one pool and on the two models where the filter gained.

\textbf{Applying the account.} Both of the account's predictors come from training runs. The base model's accuracy on its first training prompts, which is known before training, does not separate the settings (\Cref{tab:account}), so applying the account requires at least a short pilot run of outcome-only GRPO. On Llama-3.2-3B, a training reward far above held-out accuracy may reflect exposure to MATH training problems during pretraining; we did not test for contamination. The first studies, on a level 4--5 pool, were not used to form the account. They agree with it on Qwen3-0.6B and 4B ($6$ and $8$ seeds) and are inconclusive on Qwen3-1.7B, whose candidates averaged $0.52$. Starting accuracy alone does not predict a gain: training on level 5 gives Qwen3-4B the starting accuracy of 0.6B, but no gain. The composition of the trained groups does not predict a gain either.

\textbf{Compared methods.} Prior work also uses the selection rule itself. We compared it at matched tokens with dynamic sampling, the history filter, the level filter, and random choice, but not with methods that rank prompts by first sampling rollouts for them, such as LILO or online difficulty filtering \citep{foster2025learnability,bae2025online}. For dense rewards, we tested at matched tokens Monte-Carlo rewards with one, two, and four continuations at each of five cuts, whose continuations are scored and discarded; a ridge critic; and, with $768$-token completions, a variant that also trains on its continuations and a VinePPO-style advantage with rollout reuse. These variants adapt published methods to our trainer and recipe rather than reproduce them (\Cref{tab:adapt}). VinePPO estimates values at every reasoning step with nine rollouts under a KL-regularized PPO recipe; SPO-tree trains on a three-level tree with a probability mask; and TreeRL branches at uncertain tokens and trains on every path from the root to a leaf. Faithful reproductions, fewer cuts, other tree-structured estimators, or a separately trained implicit reward model may trade off cost differently.

\textbf{Gradient test.} The per-token gradient test ranked the three Monte-Carlo rewards in the same order as training, but only on one model. When we extrapolated it to training as a heuristic, it overestimated all five rewards we checked by about $0.01$ to $0.02$ (exploratory for the Monte-Carlo rewards and preregistered for the two critics; every interval excludes zero).

\textbf{Duration and transfer of the gain.} We trained beyond the budget only on Qwen3-0.6B, where most of the gain was gone by $2B$ ($+0.005$ $[-0.005,+0.016]$, compared with $+0.019$ at $B$). On Qwen2.5-1.5B, the gain replicated twice on fresh seeds, but in all three studies the crossed seed-by-problem bootstrap intervals on the $200$ problems include zero. The third study therefore also scored the $1{,}543$ MATH test problems of levels 3--5 that lie outside every other evaluation set. There, the crossed interval excludes zero ($+0.008$ $[0.004,0.012]$), so the gain carries over to other problems from the same distribution. On GSM8K, an easier task that neither configuration trains on, the gain was $+0.011$ $[0.006,0.016]$ (secondary); we tested no other task. Measured with the strict verifier, both contrasts are unresolved, because one filter run (two on GSM8K) had drifted at $B$; without the drifted runs, both are positive (exploratory). On Qwen3-0.6B, a fourth study on fresh seeds replicated the gain on the $200$ problems, at $+0.012$ compared with the first studies' $+0.021$, and found it on the further MATH problems (crossed interval $[0.006,0.016]$) and on GSM8K ($+0.013$). At equal training time, however, its lead on those seeds alone was not significant ($+0.004$ $[-0.001,+0.010]$).

\textbf{Multiplicity.} Holm's correction applies within each study, not across studies. Under a correction across studies, the replicated gains (Qwen3-0.6B and Qwen2.5-1.5B on levels 3--5) remain stronger evidence than single-study results with $p$ between $.01$ and $.05$, such as the level-5 gain on Qwen3-1.7B, the history filter's gain, and the baseline-form critic's gain at the budget.

\textbf{Truncation.} On Qwen2.5-1.5B, with a $380$-token limit, $73\%$ of the initial samples are cut off, so part of what every run learns there is to finish within the limit. The Monte-Carlo reward's deficit is smaller with $768$-token completions and a budget calibrated for them ($-0.061$ on three seeds and $-0.066$ on eight, compared with $-0.143$). With $768$-token completions and as much training time as outcome-only GRPO, the $k=4$ reward and continuation reuse still trailed ($-0.033$ and $-0.052$), and the VinePPO-style variant trailed within $\pm0.02$ ($-0.011$). The $k=1$ reward's deficit ($-0.011$ $[-0.028,+0.005]$), however, was neither significant nor shown to lie within $\pm0.02$. We therefore claim neither that a cheap Monte-Carlo reward loses per GPU-hour in that setting nor that it matches outcome-only GRPO.

\textbf{Hardware and accounting.} Every equal-time contrast compares runs on one GPU type. On other hardware, the ratio of generation time to update time may differ, and the equal-time results with it. Behavior at larger scales and on other tasks remains open. The token budget counts generated tokens. It does not count the prefill pass of the prompt critic, which we report as training time instead. Measured in training time, the head start in the three settings with a gain shrinks from $22$--$23\%$ to $13$--$17\%$ (\Cref{tab:tokens}).

\subsection{Equal-iteration study and theory}

\textbf{Sampled geometry.} The score map describes one policy and one sampled batch, and the scores and the span of whole-completion updates change during training. Also, $V$ contains every reweighting of whole completions, a larger set than the one coefficient vector that a single outcome update uses. The geometry describes dense rewards used as training coefficients. Using process rewards to rank or search among solutions at inference time is a separate use, which it does not cover.

\textbf{Optimizer and controls.} The neural runs use Adam with gradient-norm clipping at $1.0$. Per-coordinate scaling and optimizer state can change the direction of a raw gradient, so the geometric decomposition describes the gradient before these transformations. Permuting token weights preserves their magnitudes, but the norm of the projected gradient, the norm of the total update, and the policy KL can still change. To separate the effect of coherent credit from the effect of update size, controls must therefore be matched at the stage of the update being studied.

\textbf{Reference and evaluation target.} The progress reference relies on a finite number of Monte-Carlo continuations, a segmentation of each prefix, a finite continuation horizon, and a verifier. Splitting samples avoids reusing them directly, but the reference keeps these estimator choices. The strongest check, with disjoint samples and full-parameter gradients, uses four Qwen2.5-1.5B probes with mean alignment $0.84$; we report references that partially overlap with the training samples separately. Training samples at temperature $1.0$ with top-$p$ $0.95$, whereas validation uses greedy decoding. Evaluating accuracy on the same fixed prompts ties the local objective more directly to measured performance, which is what the matched-compute studies do (\Cref{sec:protocol}).

\textbf{Statistical resolution.} The primary comparison, with $n=14$, establishes equivalence within $\pm0.03$ greedy accuracy, so smaller benefits and harms remain compatible with the data. These seed intervals are conditional on the chosen validation set. A larger fixed test set, separate sampled and greedy evaluations, and replication with independent seeds would each address a different source of uncertainty.

\textbf{Breadth and horizon.} The amount of replication varies across the five-family probe panel, and most secondary training comparisons use $n\le5$. The SmolLM3 endpoints and the positive single-seed Phi-4 pilots are descriptive evidence only, and the expansion of the gate study to $n=32$ includes eight exploratory seeds. The experiments use six segments, a constant dense weight, and short training horizons. Other segmentations, schedules, datasets, initial policies, and longer horizons remain open.

\textbf{Reward integration.} Our implicit-reward control uses the log ratio between the actor and the reference policy, whereas PRIME uses a separately trained reward model in the numerator. The learned-PRM study uses one particular segmentation and a reconstructed scoring head. Finally, masking tokens and then centering credit within each completion redistributes credit to the tokens that were not masked. Gates should therefore be evaluated after every transformation the trainer applies, and a controlled allocation experiment must also match the chosen credit or update budget.

\section{Matched-Compute Studies}
\label{app:matched}

This appendix documents the matched-compute studies of \Cref{sec:dense,sec:selection}. \Cref{app:m-setup} describes the setup that all studies share, and \Cref{app:m-overview} lists every contrast and every study. \Cref{app:m-dense,app:m-help,app:m-where,app:m-last,app:m-verifier} then follow the questions of the main text. Each opens with what the studies show together and continues with one record per study: its design, its preregistered contrasts, its checks, and the secondary results that the tables do not show.

\subsection{Setup}
\label{app:m-setup}

\textbf{Why a separate implementation.} The equal-iteration comparisons of \Cref{sec:equalround} match seeds and training iterations, so the Monte-Carlo reward uses more compute there than outcome-only GRPO: every scored prefix needs continuation rollouts. To compare the methods at the compute they consume, we reimplemented the training pipeline independently of the equal-iteration code, counted every generated token toward a matched budget $B$ (in this appendix, $B$ denotes this budget, not the matrix of \Cref{sec:theory}), and extended the comparison to a learned critic and to prompt selection on six models. Together the studies comprise $\nruns$ runs and $\gpuh$ GPU-hours, with one GPU per run. All ran on H200 GPUs except S35 ($105$ GPU-hours) and $22$ GPU-hours of task studies, which ran on A100 GPUs.

\textbf{Implementation.} Rollouts are sampled with vLLM in bf16; the trainer keeps float32 master weights and Adam states with bf16 autocast compute. With pure bf16 weights, Adam's per-element step at these learning rates is smaller than half a bf16 unit in the last place for typical weights, so most weights would not change: in a check on Qwen2.5-0.5B, $87\%$ of parameters were unchanged after one step. The equal-iteration study's recipe at learning rate $10^{-5}$ degraded or collapsed every configuration, outcome-only included, over the matched horizon. A preregistered stability probe, on a development set disjoint from the training pools and test sets, selected $10^{-6}$; a rate of $3\times10^{-6}$ failed the probe. For Qwen3, a preregistered truncation rule raised the generation limit from $380$ to $768$ tokens. Qwen3-4B and Llama-3.2 runs use a microbatch size of $1$ to fit in memory, which changes neither the update nor the token count. Llama-3.2 runs also fix the date that the chat template inserts. Each iteration trains on $8$ prompts with $8$ completions each, sampled at temperature $1$ and top-$p$ $1$, and takes one optimizer step on them, with no ratio clipping or reuse of rollouts. The group advantage is the reward minus the group mean, divided by the group's population standard deviation; the loss sums the token terms and divides by the number of completions in the iteration. The optimizer is Adam ($\beta=(0.9,0.999)$, $\epsilon=10^{-8}$) without weight decay, with gradient clipping at norm $1$, no KL penalty, and a constant learning rate without warmup, so a run's result at $B$ does not depend on whether it continues past $B$ (only S32 decays the rate, linearly to zero at $B$). The token budget $B$ counts all tokens generated during training: rollouts, continuations, and the discarded groups of dynamic sampling. A preregistered calibration rule set it for each model and task before the studies: $48$ times the mean number of tokens per iteration that the $k=4$ Monte-Carlo reward generates over iterations 0--2 of one calibration run, rounded up to a multiple of $250{,}000$. The densest reward thus gets $50$ to $57$ updates, and outcome-only GRPO gets $584$ on Qwen2.5-1.5B and $755$ on Qwen3-4B.

\textbf{Models and budgets.} The models are Qwen2.5-1.5B-Instruct ($380$-token completions, $B=12.25$M); Qwen3 0.6B, 1.7B, and 4B in non-thinking mode ($768$ tokens, $B=17.25$M, $21.0$M, and $22.25$M); and, for prompt selection only, Llama-3.2-1B- and 3B-Instruct ($768$ tokens, $B=10.0$M and $8.25$M). Each iteration trains on $8$ prompts with $8$ completions each.

\textbf{Pools and evaluation.} All problems have plain numeric answers, so the last-number rule is well defined. Training uses MATH train problems of levels 4--5 ($562$ prompts after removing overlaps with the evaluation sets) or levels 3--5 ($3{,}486$ prompts, $3{,}458$ used). Evaluation uses $200$ MATH test problems of levels 3--5, every $B/16$ tokens. The level-5 (S14, S18, S19) and level 1--2 (S21) studies train and evaluate on those levels only, with their own $200$-problem test sets. S33 and S34 also score two further test sets, only at the initial evaluation and at the first evaluation at or after $B$. These scores are computed after the primary evaluation and with their own common random numbers, so they cannot affect training. The first holds the $1{,}543$ MATH test problems of levels 3--5 with numeric answers of at most $1{,}000$ characters that are in no other evaluation set. Of $2{,}295$ such problems of any length, $623$ are in earlier evaluation sets, $12$ are longer, and $117$ are near-duplicates of the level 3--5 training pool or the $200$ (exact match after normalization, or $8$-gram Jaccard similarity of at least $0.6$, the check used to clean the training pools). The second is the GSM8K test set ($1{,}319$ problems), on which neither configuration trains. At each evaluation the policy draws $8$ completions per problem at temperature $1$ and top-$p$ $1$, with common random numbers across configurations of the same seed. The primary metric is accuracy at the first evaluation at or after $B$, with no checkpoint selection. Greedy accuracy and strict-verifier accuracy, which counts only boxed answers, are secondary metrics.

\textbf{Statistics.} The unit is a seed, or a (model size, seed) pair when sizes are pooled. We use paired $t$-tests, Holm adjustment within each preregistered primary family, and two one-sided tests for equivalence within $\pm0.02$ ($\pm0.01$ in S28 and S31--S33, where the gain under test is $0.008$). A crossed seed-by-problem bootstrap, which resamples seeds and test problems together, serves as a robustness check. Secondary contrasts are not corrected for multiple comparisons. Non-inferiority means that the lower bound of the $90\%$ interval exceeds $-0.02$. Studies S15, S25, S26, and S32--S35 ran a smoke stage before their main runs, which checked that each configuration differed from its comparator only in the intended setting and that paired configurations shared their first iteration. Each preregistration also fixed how each possible outcome would be read, and the last sentence of each study below gives that reading. Because some of the gains we find are smaller than these margins, we report intervals throughout, alongside the equivalence tests.

\textbf{Preregistration.} Before its runs, each study preregistered its hypotheses, primary contrasts, Holm families, equivalence margin, budgets, and seeds. The preregistration files are kept in the project repository, so their timing rests on file-system records. For $27$ of the $34$ files whose studies have main training runs, the last modification precedes the study's first main training run. Of the other seven, six changed only in comments, and S2's two dated amendments added rules for parts of the study before those parts ran. Calibration, pilot, and smoke stages may start before a file is final (S24, S27, and S35), and dated amendments, such as S1's change of learning rate after its pilot, are recorded in the files.

\textbf{Monte-Carlo reward (S1, S25).} The completion is cut into six segments at fixed token offsets. Continuations start at the five interior cuts, with $k=4$ draws each ($k=1$ and $2$ in two S25 configurations), and each continuation may use the rest of the completion's token limit. The segment credit $\hat p_s-\hat p_{s-1}$ is added, without centering, to the group advantage. Because the cuts are at fixed offsets and the continuations run to the token limit, the credit targets the same reward as outcome-only GRPO, accuracy at the end of the completion: its expected contribution for a prompt is the gradient of that prompt's success probability. Normalized GRPO's expected update weights that gradient by $\lambda_G(p)$, which varies across prompts (\Cref{prop:population}), so adding the credit can change how the gradients of different prompts are weighted relative to each other. The equal-iteration study's rule, which centers the credit and normalizes it by its maximum, serves as a diagnostic configuration with the same continuations.

\textbf{Learned critic (S2--S5, S26).} A ridge regression on the policy's hidden states predicts the success probability of a prefix. It is refit every iteration on the rollouts of earlier iterations, so each iteration is scored by a critic fit on other iterations. The temporal-difference form credits a segment with the change in predicted value across it. The baseline form subtracts the predicted value at the start of the segment from the outcome. Because the prediction depends only on the prefix and is fit on earlier iterations, this baseline does not change the expectation of the score-function estimator for the final reward. Both forms are added to the group advantage, which, as for the Monte-Carlo reward, can change how the gradients of different prompts are weighted relative to each other. At the initial Qwen2.5-1.5B policy, over $29{,}875$ prefix cuts, the cross-fitted critic's squared error against the prefix value is $0.057$, where the value's variance across cuts is $0.076$ and the $k=4$ Monte-Carlo estimate's error is $0.010$; its Brier score against outcomes is $0.101$, against $0.122$ for a constant predictor. The critic is thus far noisier than the Monte-Carlo reward, but it needs no continuations.

\textbf{Prompt filters (S6--S24, S27, S28, S30, S31).} Each iteration draws $24$ candidates from the seed's prompt stream, $3$ times the $8$ prompts it trains on. The stream is the pool shuffled once per epoch and consumed in order, so a filter passes through it three times as fast as outcome-only GRPO. A candidate's predicted success $p$ comes from one of three sources. The critic filter uses a ridge critic on the hidden state at the end of the prompt, refit on the group mean rewards of the previous $32$ iterations. The history filter uses the prompt's own mean group reward from the last iteration in which it was trained, or the recent mean reward for prompts not yet trained. The level filter uses the recent mean reward of the prompt's MATH level. The trainer keeps the $6$ candidates with $p$ nearest the target, $1/2$ by default, plus $2$ random others. Until $32$ prompts have been trained, it takes the first $8$ candidates. The random-choice control (S11--S13, S15, S20) fits and logs the critic filter but trains on $8$ of the $24$ candidates drawn uniformly at random. Because the stream depends only on the seed and the iteration, the control and the filter of a seed score the same candidates in every iteration. DAPO-style dynamic sampling generates groups in stream order and keeps those with mixed outcomes until it has $8$; every discarded group still counts toward $B$.

\textbf{Length reward (S6--S9).} Some early configurations added a length term relative to the group. Every token of a correct completion has its credit changed by $-\lambda(L_i-m_g)/L_{\max}$, where $L_i$ is the completion's length, $m_g$ the mean length of the correct completions in its group, and $L_{\max}$ the token limit. Wrong completions are unchanged, and the term sums to zero over each group. We report this term only because of the drift it caused.

\textbf{Verifiers.} The default verifier follows the equal-iteration study's rule. It reads the last number in the final boxed expression, or, if there is no boxed expression, the last number in the completion. The strict verifier requires a boxed answer. An evaluation counts as drifted when default-verifier accuracy exceeds strict-verifier accuracy by more than $0.1$.

\subsection{Overview}
\label{app:m-overview}

\begin{figure*}[p]
\centering
\includegraphics[width=0.92\textwidth]{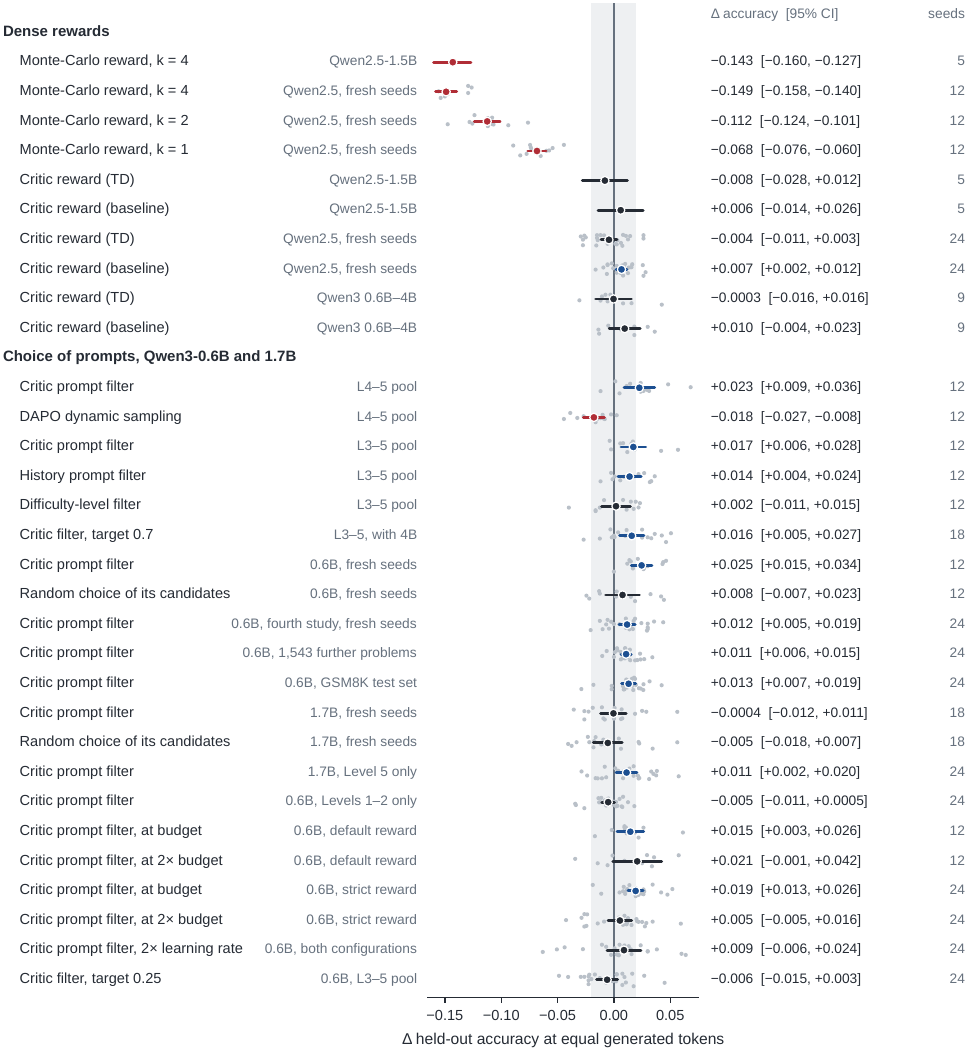}
\caption{\textbf{Every matched-compute comparison, part 1: dense rewards, and the choice of prompts on Qwen3-0.6B and 1.7B.} Held-out accuracy at the token budget minus outcome-only GRPO on the same seeds and evaluation draws, on the $200$ level 3--5 test problems (level-5 or level 1--2 problems for those pools), except in the rows marked ``further test problems'' and ``GSM8K'' (part 2). Gray dots are seeds; bars are $95\%$ paired-$t$ CIs, blue when above zero and red when below; the band marks $\pm0.02$. ``Fresh seeds'' rows come from studies that repeated an earlier comparison on new seeds. ``Default reward'' and ``strict reward'' rows train both configurations for twice the budget with that verifier as the reward. \Cref{tab:matched,tab:matched-b} mark which rows are preregistered primary contrasts.}
\label{fig:overview}
\end{figure*}

\begin{figure*}[p]
\centering
\includegraphics[width=0.92\textwidth]{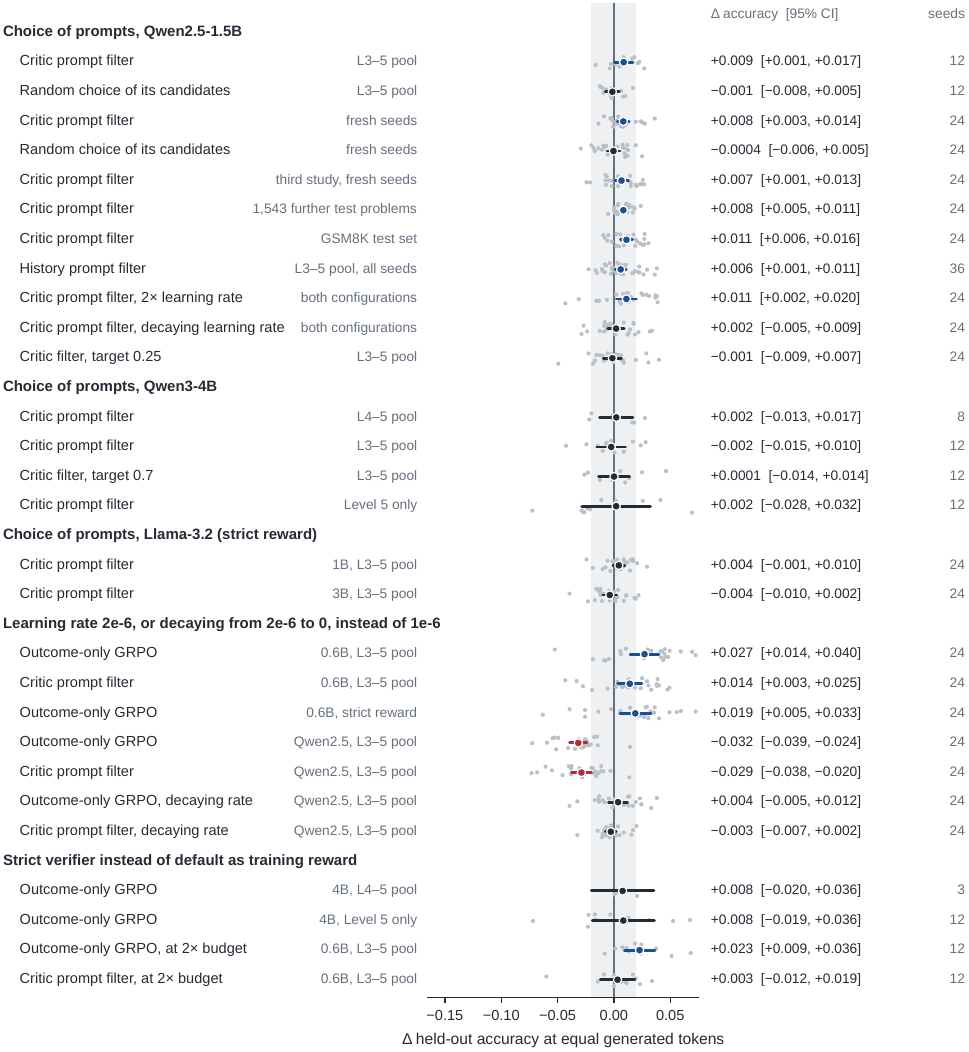}
\caption{\textbf{Every matched-compute comparison, part 2: the choice of prompts on Qwen2.5-1.5B, Qwen3-4B, and the Llama models, the learning rate, and the verifier.} As in \Cref{fig:overview}. In the last two blocks, a run at twice the learning rate or with a learning rate decaying linearly from $2\times10^{-6}$ to $0$ at the budget, or trained on the strict verifier, minus the same configuration at $10^{-6}$ or trained on the default verifier.}
\label{fig:overview-b}
\end{figure*}

\Cref{tab:matched,tab:matched-b} give every matched-compute contrast, \Cref{fig:overview,fig:overview-b} show the same contrasts with their per-seed values, and \Cref{tab:studies} lists the studies. The studies are numbered in the order they were preregistered; the artifact names each by its original label.

\begin{table}[t]
\centering
\caption{\textbf{Which interventions were tested at which model.} $\bullet$: preregistered study with $8$ or more seeds; $\circ$: pilot or pooled estimate with $3$--$9$ seeds per model; --: not tested. Completion limit in tokens; on Qwen2.5-1.5B, a preregistered study also compared the Monte-Carlo reward and both kinds of reuse at $768$ tokens, to equal training time.}
\label{tab:coverage}
\footnotesize
\setlength{\tabcolsep}{3.2pt}
\resizebox{\columnwidth}{!}{%
\begin{tabular}{@{}lcccccc@{}}
\toprule
& \multicolumn{4}{c}{Qwen} & \multicolumn{2}{c}{Llama-3.2} \\
\cmidrule(lr){2-5}\cmidrule(l){6-7}
Intervention & 2.5-1.5B & 3-0.6B & 3-1.7B & 3-4B & 1B & 3B \\
\midrule
Completion limit & $380$ & $768$ & $768$ & $768$ & $768$ & $768$ \\
\midrule
Monte-Carlo reward & $\bullet$ & $\circ$ & $\circ$ & $\circ$ & -- & -- \\
Learned-critic reward & $\bullet$ & $\circ$ & $\circ$ & $\circ$ & -- & -- \\
Rollout reuse, VinePPO-style & $\bullet$ & -- & -- & -- & -- & -- \\
Continuation reuse & $\bullet$ & -- & -- & -- & -- & -- \\
Critic filter & $\bullet$ & $\bullet$ & $\bullet$ & $\bullet$ & $\bullet$ & $\bullet$ \\
Random choice of candidates & $\bullet$ & $\bullet$ & $\bullet$ & -- & -- & -- \\
History, level, dynamic sampling & $\bullet$ & $\circ$ & $\circ$ & -- & -- & -- \\
Retuned learning rate & $\bullet$ & $\bullet$ & -- & -- & -- & -- \\
Strict verifier as reward & -- & $\bullet$ & -- & $\bullet$ & $\bullet$ & $\bullet$ \\
\bottomrule
\end{tabular}}
\end{table}

\begin{table*}[p]\centering
\caption{\textbf{Every matched-compute contrast, part 1: dense rewards within a completion, and the critic filter on the two models where it gained.} Held-out accuracy at the first evaluation at or after the token budget $B$, or after $2B$ or at outcome-only GRPO's training time at $B$ where marked, minus outcome-only GRPO on the same seeds and evaluation draws. Evaluation uses the setting's $200$ test problems ($8$ samples each), except in the rows marked ``further test problems'' (the $1{,}543$ numeric-answer MATH test problems of levels 3--5 outside every other evaluation set) and ``GSM8K'' (its $1{,}319$ test problems). Intervals are $95\%$ paired-$t$ CIs over seeds; the bars draw them on the scale above each block, over a gray band at $\pm0.02$, blue when the interval lies above zero and red when below. The per-size Monte-Carlo rows give three means without intervals. ``Holm'' marks a preregistered primary contrast with its Holm-adjusted $p$; ``TOST'' gives the equivalence $p$ at $\pm0.02$ (studies on Qwen2.5-1.5B that compare against its gain of $0.008$ preregistered $\pm0.01$, reported in the text). Secondary, estimate (seeds pooled over several studies of the same configurations), and exploratory rows carry no multiplicity correction. The Monte-Carlo reward scores six segments from $k$ continuations at each cut; learned critics are ridge regressions on hidden states that credit the change in predicted value (TD form) or subtract it from the outcome (baseline form); the critic filter keeps the $6$ of $24$ candidates whose predicted success is nearest $1/2$, plus $2$ at random. \Cref{tab:matched-b} continues the table.}
\label{tab:matched}
\setlength{\tabcolsep}{3pt}\scriptsize\renewcommand{\arraystretch}{0.94}
\begin{tabularx}{\textwidth}{@{}>{\raggedright\arraybackslash}X>{\raggedright\arraybackslash}p{2.75cm}c@{\hspace{7pt}}l@{\hspace{6pt}}c@{\hspace{4pt}}>{\raggedright\arraybackslash}p{3.3cm}@{}}\toprule
Method & Model, training pool & $n$ & $\Delta$ accuracy $[95\%$ CI$]$ & & Test \\\midrule
\noalign{\setciscale{-0.17}{0.12}}
\multicolumn{4}{@{}l}{\emph{Dense rewards within a completion}} & \ciaxis{-0.1/\textminus0.10,0/0,0.1/+0.10} & \\
Monte-Carlo reward, $k{=}4$ & Qwen2.5-1.5B, L4--5 & $5$ & $-0.143\,[-0.160,-0.127]$ & \ci{-0.143}{-0.160}{-0.127} & Holm $p{=}5{\times}10^{-5}$ \\
\quad same, at matched updates & Qwen2.5-1.5B, L4--5 & $5$ & $+0.015\,[+0.010,+0.021]$ & \ci{+0.015}{+0.010}{+0.021} & exploratory \\
\quad fresh seeds & Qwen2.5-1.5B, L4--5 & $12$ & $-0.149\,[-0.158,-0.140]$ & \ci{-0.149}{-0.158}{-0.140} & secondary \\
Monte-Carlo reward, $k{=}2$ & Qwen2.5-1.5B, L4--5 & $12$ & $-0.112\,[-0.124,-0.101]$ & \ci{-0.112}{-0.124}{-0.101} & secondary \\
Monte-Carlo reward, $k{=}1$ & Qwen2.5-1.5B, L4--5 & $12$ & $-0.068\,[-0.076,-0.060]$ & \ci{-0.068}{-0.076}{-0.060} & Holm $p{<}10^{-8}$ \\
\quad minus $k{=}4$ & Qwen2.5-1.5B, L4--5 & $12$ & $+0.081\,[+0.071,+0.090]$ & \ci{+0.081}{+0.071}{+0.090} & Holm $p{<}10^{-8}$ \\
\quad $k{=}1$ / $2$ / $4$, at matched updates & Qwen2.5-1.5B, L4--5 & $12$ & $+0.014$ / $+0.018$ / $+0.010$ & \cidots{+0.014}{+0.018}{+0.010} & exploratory \\
Monte-Carlo reward, $k{=}4$ & Qwen3 0.6B/1.7B/4B, L4--5 & $3$ each & $-0.080$ / $-0.091$ / $-0.100$ & \cidots{-0.080}{-0.091}{-0.100} & each $p\le.010$ \\
Learned critic, TD form & Qwen2.5-1.5B, L4--5 & $24$ & $-0.004\,[-0.012,+0.003]$ & \ci{-0.004}{-0.012}{+0.003} & Holm TOST $p{<}10^{-4}$ \\
\quad first study & Qwen2.5-1.5B, L4--5 & $5$ & $-0.008\,[-0.028,+0.012]$ & \ci{-0.008}{-0.028}{+0.012} & TOST $p{=}.083$ \\
Learned critic, baseline form & Qwen2.5-1.5B, L4--5 & $24$ & $+0.007\,[+0.002,+0.012]$ & \ci{+0.007}{+0.002}{+0.012} & Holm $p{=}.016$, TOST $p{<}10^{-5}$ \\
\quad first study & Qwen2.5-1.5B, L4--5 & $5$ & $+0.006\,[-0.014,+0.026]$ & \ci{+0.006}{-0.014}{+0.026} & TOST $p{=}.062$ \\
Learned critic, TD form & Qwen3, 3 sizes, L4--5 & $9$ & $-0.0003\,[-0.016,+0.016]$ & \ci{-0.0003}{-0.016}{+0.016} & TOST $p{=}.011$ \\
Learned critic, baseline form & Qwen3, 3 sizes, L4--5 & $9$ & $+0.010\,[-0.004,+0.023]$ & \ci{+0.010}{-0.004}{+0.023} & $p{=}.15$, non-inferior \\
\multicolumn{4}{@{}l}{\emph{At outcome-only GRPO's training time at $B$ ($768$ tokens, strict reward, A100)}} & & \\
Monte-Carlo reward, $k{=}4$ & Qwen2.5-1.5B, L4--5 & $8$ & $-0.033\,[-0.043,-0.024]$ & \ci{-0.033}{-0.043}{-0.024} & Holm $p{=}2{\times}10^{-4}$ \\
\quad at matched tokens & Qwen2.5-1.5B, L4--5 & $8$ & $-0.066\,[-0.079,-0.052]$ & \ci{-0.066}{-0.079}{-0.052} & secondary \\
Monte-Carlo reward, $k{=}1$ & Qwen2.5-1.5B, L4--5 & $8$ & $-0.011\,[-0.028,+0.005]$ & \ci{-0.011}{-0.028}{+0.005} & Holm $p{=}.15$ \\
VinePPO-style, $4$ steps per batch & Qwen2.5-1.5B, L4--5 & $8$ & $-0.011\,[-0.017,-0.004]$ & \ci{-0.011}{-0.017}{-0.004} & Holm $p{=}.013$, TOST $p{=}.005$ \\
Continuation reuse, $k{=}2$ & Qwen2.5-1.5B, L4--5 & $8$ & $-0.052\,[-0.063,-0.041]$ & \ci{-0.052}{-0.063}{-0.041} & Holm $p{<}10^{-4}$ \\\midrule
\noalign{\setciscale{-0.05}{0.05}}
\multicolumn{4}{@{}l}{\emph{Critic filter on Qwen3-0.6B and Qwen2.5-1.5B}} & \ciaxis{-0.04/\textminus0.04,0/0,0.04/+0.04} & \\
Critic filter & Qwen3-0.6B, L3--5 & $54$ & $+0.017\,[+0.012,+0.021]$ & \ci{+0.017}{+0.012}{+0.021} & estimate, $42$ of $54$ higher \\
\quad fresh seeds of one study & Qwen3-0.6B, L3--5 & $12$ & $+0.025\,[+0.015,+0.034]$ & \ci{+0.025}{+0.015}{+0.034} & secondary \\
\quad fresh seeds of a fourth study & Qwen3-0.6B, L3--5 & $24$ & $+0.012\,[+0.005,+0.019]$ & \ci{+0.012}{+0.005}{+0.019} & Holm $p{=}.003$ \\
\quad same, $1{,}543$ further test problems & Qwen3-0.6B, L3--5 & $24$ & $+0.011\,[+0.006,+0.015]$ & \ci{+0.011}{+0.006}{+0.015} & Holm $p{<}10^{-4}$, $21$ of $24$ higher \\
\quad same, GSM8K test set & Qwen3-0.6B, L3--5 & $24$ & $+0.013\,[+0.007,+0.019]$ & \ci{+0.013}{+0.007}{+0.019} & secondary \\
\quad default reward, at $2B$ & Qwen3-0.6B, L3--5 & $12$ & $+0.021\,[-0.001,+0.042]$ & \ci{+0.021}{-0.001}{+0.042} & Holm $p{=}.059$ \\
\quad strict reward, at $B$ & Qwen3-0.6B, L3--5 & $24$ & $+0.019\,[+0.013,+0.026]$ & \ci{+0.019}{+0.013}{+0.026} & secondary, $22$ of $24$ higher \\
\quad strict reward, at $2B$ & Qwen3-0.6B, L3--5 & $24$ & $+0.005\,[-0.005,+0.016]$ & \ci{+0.005}{-0.005}{+0.016} & Holm $p{=}.302$, TOST $p{=}.004$ \\
\quad both at learning rate $2{\times}10^{-6}$ & Qwen3-0.6B, L3--5 & $24$ & $+0.009\,[-0.006,+0.024]$ & \ci{+0.009}{-0.006}{+0.024} & Holm $p{=}.46$ \\
\quad at $10^{-6}$ vs.\ outcome at $2{\times}10^{-6}$ & Qwen3-0.6B, L3--5 & $24$ & $-0.005\,[-0.018,+0.007]$ & \ci{-0.005}{-0.018}{+0.007} & Holm $p{=}.46$, TOST $p{=}.010$ \\
\quad same, strict reward & Qwen3-0.6B, L3--5 & $24$ & $+0.000\,[-0.012,+0.013]$ & \ci{+0.000}{-0.012}{+0.013} & Holm $p{=}.96$, TOST $p{=}.002$ \\
Critic filter & Qwen3-0.6B, levels 1--2 & $24$ & $-0.005\,[-0.011,+0.000]$ & \ci{-0.005}{-0.011}{+0.000} & secondary, TOST $p{<}.001$ \\
\quad minus its gain on L3--5 & Qwen3-0.6B & $24$, $18$ & $-0.030\,[-0.039,-0.020]$ & \ci{-0.030}{-0.039}{-0.020} & Holm $p{<}10^{-6}$ (Welch) \\
Critic filter & Qwen2.5-1.5B, L3--5 & $60$ & $+0.008\,[+0.004,+0.011]$ & \ci{+0.008}{+0.004}{+0.011} & estimate, $42$ of $60$ higher \\
\quad first study & Qwen2.5-1.5B, L3--5 & $12$ & $+0.009\,[+0.001,+0.017]$ & \ci{+0.009}{+0.001}{+0.017} & Holm $p{=}.039$ \\
\quad fresh seeds of one study & Qwen2.5-1.5B, L3--5 & $24$ & $+0.008\,[+0.003,+0.014]$ & \ci{+0.008}{+0.003}{+0.014} & Holm $p{=}.006$ \\
\quad fresh seeds of a third study & Qwen2.5-1.5B, L3--5 & $24$ & $+0.007\,[+0.001,+0.013]$ & \ci{+0.007}{+0.001}{+0.013} & Holm $p{=}.034$ \\
\quad same, $1{,}543$ further test problems & Qwen2.5-1.5B, L3--5 & $24$ & $+0.008\,[+0.005,+0.011]$ & \ci{+0.008}{+0.005}{+0.011} & Holm $p{<}10^{-4}$, $23$ of $24$ higher \\
\quad same, GSM8K test set & Qwen2.5-1.5B, L3--5 & $24$ & $+0.011\,[+0.006,+0.016]$ & \ci{+0.011}{+0.006}{+0.016} & secondary \\
\quad both at learning rate $2{\times}10^{-6}$ & Qwen2.5-1.5B, L3--5 & $24$ & $+0.011\,[+0.002,+0.020]$ & \ci{+0.011}{+0.002}{+0.020} & Holm $p{=}.018$ \\
\quad at $10^{-6}$ vs.\ outcome at $2{\times}10^{-6}$ & Qwen2.5-1.5B, L3--5 & $24$ & $+0.040\,[+0.033,+0.047]$ & \ci{+0.040}{+0.033}{+0.047} & Holm $p{<}10^{-6}$ \\
\quad both at decaying learning rate & Qwen2.5-1.5B, L3--5 & $24$ & $+0.002\,[-0.005,+0.009]$ & \ci{+0.002}{-0.005}{+0.009} & Holm $p{=}.59$, TOST $p{<}10^{-4}$ \\
\quad at $10^{-6}$ vs.\ outcome, decaying rate & Qwen2.5-1.5B, L3--5 & $24$ & $+0.005\,[-0.003,+0.012]$ & \ci{+0.005}{-0.003}{+0.012} & Holm $p{=}.41$, TOST $p{<}.001$ \\
\bottomrule\end{tabularx}
\end{table*}

\begin{table*}[p]\centering
\caption{\textbf{Every matched-compute contrast, part 2: the critic filter on the other models, controls, other selection rules, learning rate, and verifier.} As in \Cref{tab:matched}; both Llama models train on the strict verifier. Random-choice rows give the difference from outcome-only GRPO over all seeds, and their test column gives the preregistered primary contrast against the critic filter, one $p$ per study with its number of seeds; random choice trains on $8$ of the filter's $24$ candidates drawn at random. History and level filters predict success from a prompt's last group reward or MATH level; $p^\star$ is the filter's target. Learning-rate rows give a configuration at $2\times10^{-6}$, or at a rate decaying linearly with generated tokens from $2\times10^{-6}$ to $0$ at $B$, minus the same configuration and seeds at $10^{-6}$. Verifier rows give a configuration trained on the strict verifier, which requires a boxed answer, minus the same configuration and seed trained on the default verifier, which falls back to the last number; both are evaluated with the default verifier, and ``NI'' marks a one-sided non-inferiority test at $-0.02$. All runs in both parts used H200 GPUs (\Cref{app:matched}), except those of the $768$-token rows at outcome-only GRPO's training time, which used A100 GPUs (\Cref{app:p39}).}
\label{tab:matched-b}
\setlength{\tabcolsep}{3pt}\scriptsize\renewcommand{\arraystretch}{0.94}
\begin{tabularx}{\textwidth}{@{}>{\raggedright\arraybackslash}X>{\raggedright\arraybackslash}p{2.75cm}c@{\hspace{7pt}}l@{\hspace{6pt}}c@{\hspace{4pt}}>{\raggedright\arraybackslash}p{3.3cm}@{}}\toprule
Method & Model, training pool & $n$ & $\Delta$ accuracy $[95\%$ CI$]$ & & Test \\\midrule
\noalign{\setciscale{-0.05}{0.05}}
\multicolumn{4}{@{}l}{\emph{Critic filter at the other four models}} & \ciaxis{-0.04/\textminus0.04,0/0,0.04/+0.04} & \\
Critic filter & Qwen3-1.7B, L3--5 & $24$ & $+0.002\,[-0.006,+0.011]$ & \ci{+0.002}{-0.006}{+0.011} & estimate, TOST $p{<}.001$ \\
\quad fresh seeds of one study & Qwen3-1.7B, L3--5 & $18$ & $-0.0004\,[-0.012,+0.011]$ & \ci{-0.0004}{-0.012}{+0.011} & secondary, TOST $p{=}.001$ \\
Critic filter & Qwen3-1.7B, level 5 & $24$ & $+0.011\,[+0.002,+0.020]$ & \ci{+0.011}{+0.002}{+0.020} & Holm $p{=}.019$ \\
Critic filter & Qwen3-4B, L4--5 & $8$ & $+0.002\,[-0.013,+0.017]$ & \ci{+0.002}{-0.013}{+0.017} & Holm $p{=}.746$ \\
Critic filter & Qwen3-4B, L3--5 & $12$ & $-0.002\,[-0.015,+0.010]$ & \ci{-0.002}{-0.015}{+0.010} & TOST $p{=}.006$ \\
Critic filter & Qwen3-4B, level 5 & $12$ & $+0.002\,[-0.028,+0.032]$ & \ci{+0.002}{-0.028}{+0.032} & Holm $p{=}.889$ \\
Critic filter & Llama-3.2-3B, L3--5 & $24$ & $-0.004\,[-0.010,+0.002]$ & \ci{-0.004}{-0.010}{+0.002} & Holm $p{=}.23$, TOST $p{<}10^{-5}$ \\
Critic filter & Llama-3.2-1B, L3--5 & $24$ & $+0.004\,[-0.001,+0.010]$ & \ci{+0.004}{-0.001}{+0.010} & Holm $p{=}.095$, TOST $p{<}10^{-5}$ \\\midrule
\multicolumn{4}{@{}l}{\emph{Random choice of the filter's candidates}} & \ciaxis{-0.04/\textminus0.04,0/0,0.04/+0.04} & \\
Random choice & Qwen3-0.6B, L3--5 & $18$ & $+0.008\,[-0.003,+0.019]$ & \ci{+0.008}{-0.003}{+0.019} & Holm $p{=}.168$ ($6$), $.008$ ($12$) \\
Random choice & Qwen2.5-1.5B, L3--5 & $36$ & $-0.001\,[-0.005,+0.003]$ & \ci{-0.001}{-0.005}{+0.003} & Holm $p{=}.011$ ($12$), $.007$ ($24$) \\
Random choice & Qwen3-1.7B, L3--5 & $24$ & $-0.002\,[-0.013,+0.009]$ & \ci{-0.002}{-0.013}{+0.009} & Holm $p{=}.184$ ($18$) \\\midrule
\multicolumn{4}{@{}l}{\emph{Other selection rules}} & \ciaxis{-0.04/\textminus0.04,0/0,0.04/+0.04} & \\
Critic filter & Qwen3 0.6B+1.7B, L4--5 & $12$ & $+0.023\,[+0.009,+0.036]$ & \ci{+0.023}{+0.009}{+0.036} & Holm $p{=}.006$ \\
Critic filter & Qwen3 0.6B+1.7B, L3--5 & $12$ & $+0.017\,[+0.006,+0.028]$ & \ci{+0.017}{+0.006}{+0.028} & Holm $p{=}.016$ \\
History prompt filter & Qwen3 0.6B+1.7B, L3--5 & $12$ & $+0.014\,[+0.004,+0.024]$ & \ci{+0.014}{+0.004}{+0.024} & secondary \\
History prompt filter & Qwen2.5-1.5B, L3--5 & $36$ & $+0.006\,[+0.001,+0.011]$ & \ci{+0.006}{+0.001}{+0.011} & Holm $p{=}.035$ \\
\quad critic filter minus history filter & Qwen2.5-1.5B, L3--5 & $36$ & $+0.002\,[-0.002,+0.007]$ & \ci{+0.002}{-0.002}{+0.007} & Holm $p{=}.27$, TOST $p{<}.001$ \\
Difficulty-level filter & Qwen3 0.6B+1.7B, L3--5 & $12$ & $+0.002\,[-0.011,+0.015]$ & \ci{+0.002}{-0.011}{+0.015} & TOST $p{=}.005$ \\
\quad critic filter minus level filter & Qwen3 0.6B+1.7B, L3--5 & $12$ & $+0.015\,[+0.004,+0.027]$ & \ci{+0.015}{+0.004}{+0.027} & Holm $p{=}.046$ \\
DAPO dynamic sampling & Qwen3 0.6B+1.7B, L4--5 & $12$ & $-0.018\,[-0.027,-0.008]$ & \ci{-0.018}{-0.027}{-0.008} & secondary \\
Critic filter, target $p^\star{=}0.7$ & Qwen3, three sizes, L3--5 & $18$ & $+0.016\,[+0.005,+0.027]$ & \ci{+0.016}{+0.005}{+0.027} & secondary \\
Critic filter, target $p^\star{=}0.7$ & Qwen3-4B, L3--5 & $12$ & $+0.0001\,[-0.014,+0.014]$ & \ci{+0.0001}{-0.014}{+0.014} & TOST $p{=}.004$ \\
Critic filter, target $p^\star{=}0.85$ & Qwen3-0.6B, L3--5 & $18$ & $+0.024\,[+0.013,+0.035]$ & \ci{+0.024}{+0.013}{+0.035} & estimate \\
Critic filter, target $p^\star{=}0.25$ & Qwen3-0.6B, L3--5 & $24$ & $-0.006\,[-0.015,+0.003]$ & \ci{-0.006}{-0.015}{+0.003} & Holm $p{=}.20$, TOST $p{=}.003$ \\
\quad minus target $1/2$ & Qwen3-0.6B, L3--5 & $24$ & $-0.028\,[-0.036,-0.020]$ & \ci{-0.028}{-0.036}{-0.020} & Holm $p{<}10^{-6}$ \\
Critic filter, target $p^\star{=}0.25$ & Qwen2.5-1.5B, L3--5 & $24$ & $-0.001\,[-0.010,+0.007]$ & \ci{-0.001}{-0.010}{+0.007} & Holm $p{=}.73$, TOST $p{<}10^{-4}$ \\
\quad minus target $1/2$ & Qwen2.5-1.5B, L3--5 & $24$ & $-0.010\,[-0.016,-0.004]$ & \ci{-0.010}{-0.016}{-0.004} & Holm $p{=}.008$ \\
\midrule
\multicolumn{4}{@{}l}{\emph{Learning rate $2{\times}10^{-6}$, or decaying from it to $0$ at $B$, minus $10^{-6}$, same configuration and seeds}} & \ciaxis{-0.04/\textminus0.04,0/0,0.04/+0.04} & \\
Outcome-only GRPO & Qwen3-0.6B, L3--5 & $24$ & $+0.027\,[+0.014,+0.040]$ & \ci{+0.027}{+0.014}{+0.040} & secondary \\
Outcome-only GRPO, strict reward & Qwen3-0.6B, L3--5 & $24$ & $+0.019\,[+0.005,+0.033]$ & \ci{+0.019}{+0.005}{+0.033} & Holm $p{=}.018$ \\
Outcome-only GRPO & Qwen2.5-1.5B, L3--5 & $24$ & $-0.032\,[-0.039,-0.024]$ & \ci{-0.032}{-0.039}{-0.024} & secondary \\
Critic filter & Qwen2.5-1.5B, L3--5 & $24$ & $-0.029\,[-0.038,-0.020]$ & \ci{-0.029}{-0.038}{-0.020} & secondary \\
Outcome-only GRPO, decaying rate & Qwen2.5-1.5B, L3--5 & $24$ & $+0.004\,[-0.005,+0.012]$ & \ci{+0.004}{-0.005}{+0.012} & secondary \\
Critic filter, decaying rate & Qwen2.5-1.5B, L3--5 & $24$ & $-0.003\,[-0.007,+0.002]$ & \ci{-0.003}{-0.007}{+0.002} & exploratory \\\midrule
\multicolumn{4}{@{}l}{\emph{Reward verification}} & \ciaxis{-0.04/\textminus0.04,0/0,0.04/+0.04} & \\
Strict verifier as training reward & Qwen3-4B, L4--5 & $3$ & $+0.008\,[-0.020,+0.036]$ & \ci{+0.008}{-0.020}{+0.036} & non-inferior \\
Strict verifier as training reward & Qwen3-4B, level 5 & $12$ & $+0.008\,[-0.019,+0.036]$ & \ci{+0.008}{-0.019}{+0.036} & Holm $p{=}.022$, NI \\
Strict verifier as training reward & Qwen3-0.6B, L3--5, $2B$ & $12$ & $+0.023\,[+0.009,+0.036]$ & \ci{+0.023}{+0.009}{+0.036} & Holm $p{<}.001$, NI \\
\quad with the critic filter & Qwen3-0.6B, L3--5, $2B$ & $12$ & $+0.003\,[-0.012,+0.019]$ & \ci{+0.003}{-0.012}{+0.019} & secondary, NI \\
\bottomrule\end{tabularx}
\end{table*}

\begin{table*}[t]\centering
\caption{\textbf{The matched-compute studies}, in the order they were preregistered. Artifact: the study's label in the artifact, whose configuration files are named after it (S35, for example, is P39, configuration \texttt{phase39}). $n$: seeds per configuration in the study's primary contrasts (several values for several models or configurations). Where: the section that reports the study.}
\label{tab:studies}
\setlength{\tabcolsep}{3pt}\scriptsize
\begin{tabularx}{\textwidth}{@{}ll>{\raggedright\arraybackslash}X>{\raggedright\arraybackslash}p{4.6cm}lc@{}}\toprule
Study & Artifact & Question & Model, pool & $n$ & Where \\\midrule
S1 & P3 & Does the Monte-Carlo reward pay for its continuations? & Qwen2.5-1.5B, L4--5; Qwen3, three sizes; GSM8K & $5$; $3$ each & \ref{app:m-dense} \\
S2 & P4 & Longer completions; gradient efficiency per update and per token & Qwen2.5-1.5B, L4--5 & $3$ & \ref{app:m-dense} \\
S3 & P5 & Pilot: reusing samples & Qwen2.5-1.5B, L4--5 & $3$ & \ref{app:pilots} \\
S4 & P6 & Learned critic; pilot of other estimators & Qwen2.5-1.5B, L4--5 & $5$; $3$ & \ref{app:m-dense} \\
S5 & P9 & Learned critic on other models & Qwen3, three sizes, L4--5 & $9$ & \ref{app:m-dense} \\
S6 & P10 & Does the critic filter help? & Qwen3 0.6B+1.7B, L4--5 & $12$ & \ref{app:m-help} \\
S7 & P11 & History filter, dynamic sampling, a larger model & Qwen3, L4--5 & $12$; $8$ & \ref{app:m-help} \\
S8 & P12 & Critic filter and history filter on a wider pool & Qwen3 0.6B+1.7B, L3--5 & $12$ & \ref{app:m-help} \\
S9 & P13 & Level filter; strict verifier as reward & Qwen3 0.6B+1.7B, L3--5; Qwen3-4B & $12$; $3$ & \ref{app:m-help} \\
S10 & P14 & Filter aimed at predicted success $0.7$ & Qwen3, three sizes, L3--5 & $18$ & \ref{app:m-help} \\
S11 & P15 & Random choice among the candidates; 4B replication & Qwen3-0.6B, Qwen3-4B, L3--5 & $6$; $6$ & \ref{app:m-help} \\
S12 & P16 & Random choice on fresh seeds & Qwen3-0.6B, L3--5 & $12$ & \ref{app:m-help} \\
S13 & P17 & Replication with random choice & Qwen3-1.7B, L3--5 & $18$ & \ref{app:m-help} \\
S14 & P18 & Is the 4B null a ceiling? & Qwen3-4B, level 5 & $12$ & \ref{app:m-help} \\
S15 & P19 & Critic filter and random choice & Qwen2.5-1.5B, L3--5 & $12$ & \ref{app:m-help} \\
S16 & P20 & Twice the budget & Qwen3-0.6B, L3--5 & $12$ & \ref{app:m-last} \\
S17 & P21 & Twice the budget, strict reward & Qwen3-0.6B, L3--5 & $24$ & \ref{app:m-last} \\
S18 & P22 & Strict verifier as reward & Qwen3-4B, level 5 & $12$ & \ref{app:m-verifier} \\
S19 & P23 & Account: a gain predicted & Qwen3-1.7B, level 5 & $24$ & \ref{app:m-where} \\
S20 & P24 & Replication on fresh seeds & Qwen2.5-1.5B, L3--5 & $24$ & \ref{app:m-help} \\
S21 & P25 & Account: no gain predicted & Qwen3-0.6B, levels 1--2 & $24$ & \ref{app:m-where} \\
S22 & P26 & A filter without a critic & Qwen2.5-1.5B, L3--5 & $36$ & \ref{app:m-help} \\
S23 & P27 & Twice the learning rate & Qwen3-0.6B, L3--5 & $24$ & \ref{app:m-last} \\
S24 & P28 & Account outside the Qwen family: no gain predicted & Llama-3.2-3B, L3--5 & $24$ & \ref{app:m-where} \\
S25 & P29 & Fewer continuations per cut & Qwen2.5-1.5B, L4--5 & $12$ & \ref{app:m-dense} \\
S26 & P30 & Learned critic on fresh seeds & Qwen2.5-1.5B, L4--5 & $24$ & \ref{app:m-dense} \\
S27 & P31 & Account outside the Qwen family: a gain predicted & Llama-3.2-1B, L3--5 & $24$ & \ref{app:m-where} \\
S28 & P32 & Twice the learning rate & Qwen2.5-1.5B, L3--5 & $24$ & \ref{app:m-last} \\
S29 & P33 & Twice the learning rate without drift & Qwen3-0.6B, L3--5 & $24$ & \ref{app:m-last} \\
S30 & P34 & Filter aimed at harder prompts & Qwen3-0.6B, L3--5 & $24$ & \ref{app:m-where} \\
S31 & P35 & Filter aimed at harder prompts & Qwen2.5-1.5B, L3--5 & $24$ & \ref{app:m-where} \\
S32 & P36 & Decaying learning rate & Qwen2.5-1.5B, L3--5 & $24$ & \ref{app:m-last} \\
S33 & P37 & Further test problems & Qwen2.5-1.5B, L3--5 & $24$ & \ref{app:m-help} \\
S34 & P38 & Further test problems & Qwen3-0.6B, L3--5 & $24$ & \ref{app:m-help} \\
S35 & P39 & Equal training time past the budget & Qwen2.5-1.5B, L4--5, $768$ tokens & $8$ & \ref{app:p39} \\
\bottomrule\end{tabularx}
\end{table*}

\textbf{Training time per run.} Scoring the $24$ candidates adds one prefill pass per iteration, and the configurations complete different numbers of iterations within the budget because their completions differ in length. The mean training time per run, in hours, for the critic filter and for outcome-only GRPO is: $0.57$ vs.\ $0.52$ (Qwen3-0.6B, eighteen seeds), $0.93$ vs.\ $0.89$ (1.7B, $24$), $2.45$ vs.\ $2.52$ (4B, level 3--5 pool, $12$), $2.29$ vs.\ $2.14$ (4B, level 5, $12$), $0.85$ vs.\ $0.78$ (1.7B, level 5, $24$), $0.67$ vs.\ $0.70$ (0.6B, level 1--2, $24$), $0.52$ vs.\ $0.45$ (Qwen2.5-1.5B, $36$), $0.73$ vs.\ $0.70$ (Llama-3.2-3B, $24$), and $0.525$ vs.\ $0.450$ (Llama-3.2-1B, $24$), a change of $-5\%$ to $+17\%$. The random-choice control, which also scores every candidate, takes $0.56$, $0.96$, and $0.50$ hours on Qwen3-0.6B, Qwen3-1.7B, and Qwen2.5-1.5B.

\subsection{Dense rewards at matched compute}
\label{app:m-dense}

These studies ask whether the per-update advantage of dense rewards survives once their continuations are paid for, and, if not, whether a cheaper estimate closes the gap. The Monte-Carlo reward spends $91\%$ of its generated tokens on continuations, so by the time it reaches $B$ it has made $50$ updates to outcome-only GRPO's $584$. That cost decides the comparison on every model we trained (\Cref{tab:early}). Fewer continuations per cut shrink the deficit, in the order the per-token gradient test predicts, because continuation sampling adds little noise to the update (S25). The heuristic extrapolation from that test overestimates all five rewards we checked by $0.008$ to $0.018$ on average over the evaluations, and every interval excludes zero. This error is small next to the Monte-Carlo rewards' extrapolated deficits, so their order survives, but it is as large as the critics' extrapolated leads. A learned critic, which needs no continuations, matches outcome-only GRPO within $\pm0.02$ (S26); its temporal-difference form shortens completions, by $16\%$ $[6,24]$ on Qwen3 and $3\%$ $[1,6]$ on Qwen2.5-1.5B. The equal-iteration study's rule for combining credit, which centers it and normalizes it by its maximum, is equivalent to ours with the same continuations (\Cref{tab:early}), so the result does not hinge on that choice. No Monte-Carlo reward we can afford, with $k\ge1$ and six segments, matches outcome-only GRPO at this budget, and the designs that reuse samples or train past the budget leave the ranking unchanged (\Cref{app:pilots,app:p39}).

\subsubsection{First studies (S1--S10)}
\Cref{tab:early} gives the source of every number from S1--S10 that \Cref{sec:dense,sec:selection} quote and that later sections of this appendix do not report. Gradient efficiencies were measured on $96$ fixed batches of $8\times8$ rollouts at the initial Qwen2.5-1.5B policy, relative to outcome-only GRPO. The expected alignment $\mathbb{E}\langle u,\nabla J\rangle$ was positive for every method in every screen, at least three standard errors above zero, so squaring it in the efficiency hides no sign. Of these studies, the GSM8K study ran on A100 GPUs and the others on H200 GPUs.

\begin{table*}[t]\centering
\caption{\textbf{Results of the first matched-compute studies} quoted in \Cref{sec:dense,sec:selection}. Differences are in held-out accuracy at $B$ against outcome-only GRPO on the same seeds unless stated, with $95\%$ paired-$t$ CIs; efficiencies are ratios to outcome-only GRPO with $95\%$ bootstrap CIs. ``Holm'' marks a preregistered primary contrast; other rows are secondary or, where marked, exploratory.}
\label{tab:early}
\setlength{\tabcolsep}{3pt}\small
\begin{tabular}{@{}lllcll@{}}\toprule
 & Comparison & Model, pool & $n$ & Estimate & Test \\\midrule
S1 & Monte-Carlo reward, $k=4$ & Qwen2.5-1.5B, L4--5 & $5$ & $-0.143\,[-0.160,-0.127]$ & Holm $p=5{\times}10^{-5}$ \\
 & \quad at matched updates ($50$ vs.\ $584$) & & $5$ & $+0.015\,[+0.010,+0.021]$ & exploratory \\
 & \quad centered, max-normalized rule minus it & & $5$ & $-0.007\,[-0.013,+0.000]$ & TOST $p=.003$ \\
 & Monte-Carlo reward, $k=4$ & Qwen3-0.6B, L4--5 & $3$ & $-0.080\,[-0.093,-0.067]$ & $p=.001$ \\
 & & Qwen3-1.7B, L4--5 & $3$ & $-0.091\,[-0.118,-0.065]$ & $p=.005$ \\
 & & Qwen3-4B, L4--5 & $3$ & $-0.100\,[-0.142,-0.057]$ & $p=.010$ \\
 & & Qwen2.5-1.5B, GSM8K & $3$ & $-0.098\,[-0.134,-0.062]$ & $p=.007$ \\
S2 & Monte-Carlo reward, $k=4$, $768$ tokens & Qwen2.5-1.5B, L4--5 & $3$ & $-0.061\,[-0.074,-0.048]$ & Holm $p=.003$ \\
 & efficiency, $k=4$ reward, per update & Qwen2.5-1.5B & -- & $1.56\,[1.28,2.03]$ & -- \\
 & \quad per generated token & & -- & $0.145\,[0.120,0.190]$ & -- \\
 & efficiency per token, TD critic & & -- & $1.36\,[1.20,1.62]$ & -- \\
 & \quad baseline critic & & -- & $1.15\,[0.97,1.39]$ & -- \\
S4 & learned critic, TD form & Qwen2.5-1.5B, L4--5 & $5$ & $-0.008\,[-0.028,+0.012]$ & TOST $p=.083$ \\
 & \quad baseline form & & $5$ & $+0.006\,[-0.014,+0.026]$ & TOST $p=.062$ \\
S5 & learned critic, TD form & Qwen3, 3 sizes, L4--5 & $9$ & $-0.0003\,[-0.016,+0.016]$ & TOST $p=.011$ \\
 & \quad baseline form & & $9$ & $+0.010\,[-0.004,+0.023]$ & $p=.15$, non-inferior \\
 & \quad TD form, log length ($16\%$ shorter) & & $9$ & $-0.169\,[-0.280,-0.058]$ & Holm $p=.016$ \\
S6 & critic filter & Qwen3 0.6B+1.7B, L4--5 & $12$ & $+0.023\,[+0.009,+0.036]$ & Holm $p=.006$ \\
 & \quad per size & Qwen3-0.6B, L4--5 & $6$ & $+0.028\,[+0.007,+0.050]$ & secondary \\
 & & Qwen3-1.7B, L4--5 & $6$ & $+0.017\,[-0.007,+0.040]$ & secondary \\
S7 & critic minus history filter & Qwen3 0.6B+1.7B, L4--5 & $12$ & $+0.005\,[-0.004,+0.015]$ & TOST $p=.003$ \\
 & critic filter & Qwen3-4B, L4--5 & $8$ & $+0.002\,[-0.013,+0.017]$ & Holm $p=.746$ \\
 & DAPO dynamic sampling & Qwen3 0.6B+1.7B, L4--5 & $12$ & $-0.018\,[-0.027,-0.008]$ & secondary \\
S8 & critic filter & Qwen3 0.6B+1.7B, L3--5 & $12$ & $+0.017\,[+0.006,+0.028]$ & Holm $p=.016$ \\
 & critic minus history filter & & $12$ & $+0.003\,[-0.007,+0.014]$ & TOST $p=.003$ \\
 & history filter & & $12$ & $+0.014\,[+0.004,+0.024]$ & secondary \\
S9 & critic minus level filter & Qwen3 0.6B+1.7B, L3--5 & $12$ & $+0.015\,[+0.004,+0.027]$ & Holm $p=.046$ \\
 & level filter & & $12$ & $+0.002\,[-0.011,+0.015]$ & TOST $p=.005$ \\
 & strict verifier as reward & Qwen3-4B, L4--5 & $3$ & $+0.008\,[-0.020,+0.036]$ & non-inferior \\
S10 & critic filter, target $0.7$ & Qwen3, 3 sizes, L3--5 & $18$ & $+0.016\,[+0.005,+0.027]$ & secondary \\
\bottomrule\end{tabular}
\end{table*}

\subsubsection{Fewer continuations per cut (S25)}
S25 repeats the S1 protocol (Qwen2.5-1.5B, level 4--5 pool, $B=12.25$M) on fresh seeds 5--16 with four configurations: outcome-only GRPO and the Monte-Carlo reward with $k=4$, $2$, and $1$ continuations per interior cut.

\textbf{The prediction.} A second gradient measurement scored rewards with $k=1$ to $16$ on $96$ batches at the initial policy, against the same reference gradient as the first measurement but with separately sampled rollouts; its values for $k=4$ ($1.63\times$ per update and $0.15\times$ per token) differ from the first measurement's ($1.56\times$ and $0.145\times$) because of this resampling. Per update, the rewards with $k=1$, $2$, $4$, $8$, and $16$ are $1.55\times$, $1.60\times$, $1.63\times$, $1.67\times$, and $1.69\times$ as efficient as outcome-only GRPO; per generated token, they are $0.45\times$ $[0.35,0.63]$, $0.27\times$ $[0.21,0.38]$, $0.15\times$ $[0.12,0.21]$, $0.08\times$, and $0.04\times$. Fitting the update's second moment as $a+b/k$ attributes $3.8\%$ $[3.4,4.2]$ of it at $k=4$ to continuation sampling. Therefore, no allocation of the $k=4$ continuations across cuts can raise the per-update efficiency by more than a factor of $1.04$. The preregistration predicted that every Monte-Carlo reward trails outcome-only GRPO and that the rewards order as $k=1>2>4$.

\textbf{Primary contrasts} (Holm over two; $n=12$).
\begin{itemize}
\item $k=1$ minus outcome-only GRPO: $-0.068$ $[-0.076,-0.060]$, Holm $p=1.5\times10^{-9}$, lower in all $12$ seeds; crossed bootstrap $[-0.087,-0.051]$.
\item $k=1$ minus $k=4$: $+0.081$ $[0.071,0.090]$, Holm $p=1.5\times10^{-9}$, higher in all $12$; crossed $[0.063,0.100]$.
\end{itemize}
\textbf{Secondary contrasts.} $k=4$ minus outcome-only GRPO: $-0.149$ $[-0.158,-0.140]$, which replicates S1's $-0.143$. $k=2$ minus outcome-only: $-0.112$ $[-0.124,-0.101]$. Between rewards, $k=2$ leads $k=4$ by $0.036$ $[0.025,0.048]$ and $k=1$ leads $k=2$ by $0.044$ $[0.036,0.052]$, each in all twelve seeds. On strict-verifier accuracy, $k=1$ trails outcome-only GRPO by $0.096$ $[0.087,0.104]$, and averaged over the first half of the budget it trails by $0.060$ $[0.055,0.065]$.

\textbf{Descriptive.} At $B$, the configurations have made $599$, $163$, $92$, and $50$ updates (outcome-only, $k=1$, $2$, $4$). Continuations take $70\%$, $83\%$, and $91\%$ of the Monte-Carlo rewards' tokens. The share of cut-off samples falls from $74\%$ at the start to $22\%$, $46\%$, $58\%$, and $67\%$, so the rewards with fewer updates keep more of the initial policy's long completions. Training takes $0.43$, $0.20$, $0.16$, and $0.14$ hours per run. The $k=1$ deficit grows to $0.094$ at $10B/16$ and is $0.068$ at $B$. Its lead over $k=4$ grows to $0.081$ at $B$.

\textbf{A heuristic extrapolation (exploratory).} If a reward's per-token efficiency $e$ held throughout training, training with it would reach, after $t$ tokens, the accuracy that outcome-only GRPO reaches after $et$ tokens. This is a heuristic, not a consequence of the analysis. Read off outcome-only GRPO's curve on the same seed, it gives $0.359$, $0.319$, and $0.275$ at $B$ for $k=1$, $2$, and $4$. The rewards reached $0.343$, $0.299$, and $0.263$, against outcome-only GRPO's $0.412$. Averaged over all $16$ evaluations, observed minus extrapolated accuracy is $-0.017$ $[-0.023,-0.012]$, $-0.013$ $[-0.020,-0.006]$, and $-0.008$ $[-0.013,-0.002]$. Applied to the critics of \Cref{sec:q2} (seeds 0--4), the same extrapolation uses only the evaluations at which $et$ stays within the length of the outcome-only run: up to $11B/16$ for the temporal-difference form and up to $13B/16$ for the baseline form. It gives leads of $+0.020$ and $+0.008$; the critics led by $+0.003$ $[-0.012,+0.018]$ and $+0.000$. Observed minus extrapolated is $-0.017$ $[-0.031,-0.003]$ and $-0.008$ $[-0.020,+0.005]$. The extrapolation thus overestimates all five rewards by similar amounts. S26 then tested the critics' shortfall as a preregistered prediction on $24$ fresh seeds (next subsection), and it held. The efficiency is measured at the initial policy and describes the first-order gain of an SGD step tuned to each reward, whereas training uses Adam at one fixed learning rate.

\subsubsection{Learned critic on fresh seeds (S26)}
S26 repeated the S4 critic comparison on Qwen2.5-1.5B (level 4--5 pool, $380$ tokens, $B=12.25$M, the same configuration files) on seeds 5--28, after S25 suggested that the extrapolation overestimates the critics. Outcome-only seeds 5--16 are the S25 runs, finished before S26 was designed; seeds 17--28 were trained in S26. The primary family was equivalence within $\pm0.02$ for both forms (TOST, Holm over the two). The preregistration also specified each primary contrast's two-sided paired $t$, Holm-adjusted over the two, and that a form significantly above zero would count as a gain at matched tokens.

\textbf{Primary contrasts.} Both forms are equivalent to outcome-only GRPO.
\begin{itemize}
\item Temporal-difference form minus outcome-only: $-0.0044$ $[-0.0115,+0.0027]$, TOST Holm $p=6.7\times10^{-5}$, higher in $11$ of $24$ seeds; crossed bootstrap $[-0.016,+0.007]$.
\item Baseline form minus outcome-only: $+0.0067$ $[+0.0020,+0.0115]$, TOST Holm $p=7.9\times10^{-6}$; two-sided $p=.008$ (Holm $.016$), higher in $17$ of $24$; crossed bootstrap $[-0.002,+0.016]$.
\end{itemize}
By that rule, the baseline form gains. Its lead appears only at $B$, however. At $14B/16$ and $15B/16$ it was $-0.003$ $[-0.009,+0.003]$ and $+0.005$ $[-0.003,+0.012]$.

\textbf{Secondary contrasts.} On strict-verifier accuracy, the temporal-difference form minus outcome-only GRPO is $-0.021$ $[-0.049,+0.007]$. Two of its runs drifted: seed 5 from $13B/16$ on, by $0.29$ to $0.36$, and seed 15 at $B$, by $0.14$. Without them it is $-0.004$ $[-0.011,+0.004]$. On strict-verifier accuracy the baseline form leads by $0.004$ $[-0.002,+0.011]$. Averaged over the first half of the budget, the temporal-difference form leads by $0.004$ $[0.001,0.008]$ ($p=.017$, predicted no difference) and the baseline form by $0.000$ $[-0.003,+0.004]$. The temporal-difference form trails the baseline form by $0.011$ $[0.005,0.017]$ at $B$ ($p<.001$, predicted no difference).

\textbf{Shortfall from the extrapolation (preregistered, F1 and F2).} The extrapolation gave the temporal-difference form ($e=1.36$) a lead of $+0.0205$ $[0.0195,0.0216]$ on average over the evaluations up to $11B/16$. It led by $+0.0023$ $[-0.0013,+0.0059]$, a shortfall of $0.018$ $[0.015,0.022]$ ($p=9\times10^{-11}$). It gave the baseline form ($e=1.15$) $+0.0088$ up to $13B/16$. That form led by $+0.0001$ $[-0.003,+0.003]$, a shortfall of $0.009$ $[0.006,0.012]$ ($p=6\times10^{-6}$). Pooled with seeds 0--4 ($n=29$, estimates), the forms differ from outcome-only GRPO by $-0.005$ $[-0.011,+0.001]$ and $+0.007$ $[0.002,0.011]$ at $B$, and fall short of the extrapolation by $0.018$ and $0.009$.

\textbf{Descriptive.} At $B$, accuracy is $0.411$ (outcome-only), $0.407$ (temporal-difference), and $0.418$ (baseline). The temporal-difference form's greedy completions are $3\%$ $[1,6]$ shorter, and $18.7\%$ of its samples are truncated against $22.4\%$. Both critics are in use from iteration 4 on.

\subsection{Does prompt selection help?}
\label{app:m-help}

Prompt selection spends compute differently from dense rewards. One forward pass over $24$ candidate prompts per iteration decides which $8$ prompts to sample, and scoring generates no tokens. Curriculum methods that train on prompts with predicted success near $1/2$ use the same rule \citep{gao2025pcl,qu2025mopps,sun2025dots}, so we test the rule rather than propose it. It gains on two of six models with the level 3--5 pool, Qwen3-0.6B and Qwen2.5-1.5B, and both gains replicate on fresh seeds (\Cref{tab:matched}). The studies below probe the gain from four sides. Fresh seeds and further test sets ask whether it is real (S12, S20, S33, S34). Random choice among the same candidates asks whether it comes from the choice of prompts or from moving faster through the prompt stream (S11--S13, S15, S20). Simpler estimates of a prompt's success ask whether the learned critic is needed (S7--S9, S22). Other models and pools ask how far the gain extends (S13, S14, and \Cref{app:m-where}).

The answers are consistent. On Qwen2.5-1.5B, the gain carries over to $1{,}543$ MATH test problems that no study had evaluated on, with an interval that also resamples the problems, and to GSM8K, which neither configuration trains on; the same holds on Qwen3-0.6B. On both models, the choice of prompts carries the gain: random choice among the same candidates is equivalent to outcome-only GRPO (on Qwen3-0.6B, $+0.008$ $[-0.003,+0.019]$ over $18$ seeds from two studies, TOST $p=.014$), and it trails the filter in every preregistered primary contrast. A per-prompt success estimate near $1/2$ carries most of the gain at $B$ even without a learned critic, and the critic adds an early lead. A filter built from the MATH difficulty level alone gains nothing, and dynamic sampling loses at matched tokens, because the groups it discards still count toward the budget.

\subsubsection{Random choice among the candidates (S11, S12) and a Qwen3-4B replication (S11)}
\Cref{tab:random} lists the Qwen3-0.6B configurations that share a seed's prompt stream and evaluation draws. Random choice among the filter's candidates reproduces outcome-only GRPO's mix of uniform and mixed groups, while training on fewer distinct prompts. The three critic-filter configurations train on fewer all-wrong groups than either. Target $0.85$ trains on the same share of mixed-outcome groups as random choice and still leads it by $0.016$ $[0.009,0.024]$ over eighteen seeds (higher in $15$). The share of mixed-outcome groups, which dynamic sampling maximizes, therefore does not by itself account for the differences in accuracy.

\begin{table}[h]\centering
\caption{\textbf{Qwen3-0.6B configurations on one prompt stream} (L3--5 pool; seeds 0--17, $n=18$, except target $0.7$, seeds 0--5, $n=6$). Accuracy at $B$; shares of trained groups that were all-wrong, all-correct, or mixed; and distinct prompts trained, over iterations before $B$. Filter rows are the critic filter at target $p^\star$; random choice draws from the same candidates.}
\label{tab:random}
\setlength{\tabcolsep}{3pt}\small
\begin{tabular}{lccccc}\toprule
Configuration & accuracy & wrong & correct & mixed & prompts \\\midrule
Outcome only & $0.440$ & $0.359$ & $0.140$ & $0.500$ & $3{,}458$ \\
Random choice & $0.447$ & $0.353$ & $0.144$ & $0.503$ & $2{,}557$ \\
Filter, $p^\star{=}1/2$ & $0.464$ & $0.235$ & $0.153$ & $0.612$ & $2{,}248$ \\
Filter, $p^\star{=}0.7$ & $0.466$ & $0.167$ & $0.267$ & $0.566$ & $2{,}215$ \\
Filter, $p^\star{=}0.85$ & $0.464$ & $0.151$ & $0.344$ & $0.505$ & $2{,}130$ \\
\bottomrule\end{tabular}
\end{table}

The first random-choice study (S11, seeds 0--5) left its primary contrast unresolved: random choice minus the filter was $-0.017$ $[-0.037,+0.003]$ (Holm $p=.168$). Adding seeds to that contrast after seeing the result would have been optional stopping. S12 therefore preregistered the same contrast as its only primary test on twelve fresh seeds (6--17), repeating random choice, the filter at targets $1/2$ and $0.85$, and outcome-only GRPO. Pooled values over seeds 0--17 are reported as estimates.

The manipulation check passed in every run. The control's chosen prompts matched the candidates' predicted success to within $0.005$, and it trained on $0.118$ $[0.109,0.127]$ more all-wrong groups than the filter of the same seed. The control and the filter scored identical candidates in every iteration that both runs reached. On the fresh seeds, random choice minus the filter is $-0.017$ $[-0.029,-0.005]$ ($p=.008$; per seed $-0.029$, $-0.027$, $-0.008$, $-0.038$, $+0.002$, $-0.020$, $+0.023$, $-0.039$, $-0.014$, $-0.001$, $-0.028$, $-0.024$). The crossed seed-by-problem bootstrap gives $[-0.032,-0.002]$. Strict-verifier accuracy moves the same way ($-0.017$ $[-0.036,+0.002]$). The secondary contrasts on the fresh seeds are: filter minus outcome $+0.025$ $[0.015,0.034]$ ($p<10^{-4}$); random choice minus outcome $+0.008$ $[-0.007,+0.023]$ (TOST $p=.048$); target $0.85$ minus random choice $+0.015$ $[0.005,0.024]$ ($p=.007$); target $0.85$ minus target $1/2$ $-0.002$ (TOST $p=.005$); and target $0.85$ minus outcome $+0.022$ $[0.007,0.038]$ ($p=.010$).

Per seed, the filter's gain over outcome-only GRPO splits into the critic's choice (filter minus random choice) and the stream (random choice minus outcome). Over the eighteen seeds, the gain is $+0.024$ $[0.016,0.033]$, the critic's choice $+0.017$ $[0.008,0.026]$, and the stream $+0.008$ $[-0.003,+0.019]$; the parts sum to the gain up to rounding.

On Qwen3-4B, fresh seeds 6--11 repeated target $0.7$, target $1/2$, and outcome-only GRPO. Per seed, target $0.7$ minus target $1/2$ was $-0.017$, $-0.016$, $-0.040$, $+0.016$, $+0.012$, and $-0.022$. Pooled with seeds 0--5 ($n=12$, secondary), the three pairwise differences are $+0.003$ (target $0.7$ minus $1/2$), $+0.0001$ (target $0.7$ minus outcome), and $-0.002$ ($1/2$ minus outcome). Each is equivalent within $\pm0.02$ (TOST $p=.008$, $.004$, $.006$). One fresh filter run stopped boxing its answers late in training and ended with strict-verifier accuracy $0.004$ at default-verifier accuracy $0.814$. Over the twelve seeds, one outcome-only run (strict $0.312$) and no target-$0.7$ run ended drifted.

\subsubsection{Qwen3-1.7B replication with random choice (S13)}
On seeds 0--5 (S8), the Qwen3-1.7B filter led outcome-only GRPO by $+0.010$ $[0.001,0.019]$ ($p=.033$). S13 preregistered random choice minus the filter as its only primary contrast on eighteen fresh seeds (6--23), with the filter and outcome-only configurations repeated on those seeds and random choice added on seeds 0--5. Random choice minus the filter was $-0.005$ $[-0.013,+0.003]$ ($p=.184$; crossed bootstrap $[-0.017,+0.006]$), equivalent to zero within $\pm0.02$ (TOST $p<.001$). The filter's gain did not replicate: filter minus outcome was $-0.0004$ $[-0.012,+0.011]$ (TOST $p=.001$), and random choice minus outcome $-0.005$ $[-0.018,+0.007]$. Over all $24$ seeds, the three differences are $+0.002$ $[-0.006,+0.011]$ (filter minus outcome), $-0.004$ $[-0.011,+0.003]$ (random minus filter), and $-0.002$ $[-0.013,+0.009]$ (random minus outcome). The manipulation check passed in every run. The control's chosen prompts matched the candidates' predicted success to within $0.007$. It trained on $2.5$ $[2.1,2.9]$ points more all-wrong groups than the filter, and the two configurations scored identical candidates in all $11{,}102$ shared iterations before $B$. On this size the filter removes mainly all-correct groups. Over the fresh seeds, the all-correct share falls from $38\%$ to $27\%$ and the all-wrong share from $25\%$ to $22\%$, while the mixed share rises from $38\%$ to $51\%$.

\subsubsection{Qwen3-4B on level-5 problems (S14)}
On Qwen3-4B, outcome-only accuracy on the level 3--5 test set rises from $0.62$ to $0.76$ within $3B/16$ and then plateaus near $0.81$. S14 asked whether the 4B null reflects this ceiling. Both configurations trained on the level-5 problems of the training pool ($1{,}460$ after removing near-duplicates of the new test set) and were evaluated on $200$ level-5 test problems disjoint from the level 3--5 set. There, 4B starts at $0.350$, and outcome-only GRPO reaches $0.629$ at $B$, below the preregistered ceiling of $0.80$. The primary contrast, filter minus outcome at $B$, was $+0.002$ $[-0.028,+0.032]$ ($p=.889$; crossed $[-0.027,+0.033]$; TOST $p=.109$; per-seed SD $0.048$). Secondary results:
\begin{itemize}
\item The filter led early: $+0.018$ at $B/16$, $+0.042$ at $B/8$ ($p<10^{-4}$), and $+0.016$ at $3B/16$. Its mean lead over the first half of the budget was $+0.013$ $[0.001,0.025]$ ($p=.032$).
\item Strict-verifier accuracy differed by $+0.099$ $[-0.051,+0.250]$, carried by two drifted outcome-only runs.
\item The filter shortened completions (log length $-0.134$ $[-0.222,-0.046]$), with $11\%$ of samples truncated at $B$ against $17\%$.
\item It trained on $10.4$ $[9.0,11.7]$ points more mixed groups.
\end{itemize}
Its chosen prompts were nearer $1/2$ than its candidates in all twelve runs ($0.514$ against $0.544$).

\subsubsection{Qwen2.5-1.5B with random choice (S15)}
S15 repeated the random-choice design on the equal-iteration study's Qwen2.5-1.5B-Instruct. It used the level 3--5 pool, $380$-token completions, $B=12.25$M, and seeds 0--11. The filter and random-choice configurations used the Qwen3 filter settings. Holm adjustment covered two primary contrasts:
\begin{itemize}
\item Filter minus outcome: $+0.0086$ $[0.0005,0.0168]$ (Holm $p=.039$; higher in $9$ of $12$; crossed bootstrap $[-0.004,+0.022]$).
\item Random choice minus the filter: $-0.0101$ $[-0.0165,-0.0036]$ (Holm $p=.011$; lower in $10$ of $12$; crossed $[-0.023,+0.002]$).
\end{itemize}
Random choice minus outcome was $-0.001$ $[-0.008,+0.005]$ (TOST $p<.001$). Strict-verifier accuracy moves the same way: $+0.008$ $[0.001,0.015]$ for filter minus outcome and $-0.019$ $[-0.034,-0.003]$ for random choice minus the filter. Per seed, the gain splits into the critic's choice, $+0.010$ $[0.004,0.016]$, and the stream, $-0.001$ $[-0.008,+0.005]$. The manipulation check passed. The control's chosen prompts matched the candidates' predicted success to within $0.006$, and it trained on $12.3$ points more all-wrong groups than the filter. The two configurations scored identical candidates in all $7{,}527$ shared iterations. The filter's lead was $+0.030$ at $3B/16$ and $B/4$ and declined to $+0.009$ at $B$.

\subsubsection{Replication on Qwen2.5-1.5B (S20)}
S20 repeated the three S15 configurations on fresh seeds 12--35, with no other change. Holm adjustment covered two primary contrasts:
\begin{itemize}
\item Filter minus outcome: $+0.0083$ $[0.0031,0.0135]$ (Holm $p=.006$; higher in $18$ of $24$; crossed bootstrap $[-0.001,+0.018]$).
\item Random choice minus the filter: $-0.0087$ $[-0.0148,-0.0026]$ (Holm $p=.007$; lower in $18$ of $24$; crossed $[-0.019,+0.001]$).
\end{itemize}
Random choice minus outcome was $-0.0004$ $[-0.006,+0.005]$ (TOST $p<.001$). Per seed, the gain splits into the critic's choice, $+0.009$ $[0.003,0.015]$, and the stream, $-0.000$ $[-0.006,+0.005]$. The manipulation check passed. The control's chosen prompts matched the candidates' predicted success to within $0.011$, it trained on $12.6$ points more all-wrong groups than the filter, the filter's chosen prompts were nearer $1/2$ than its candidates in every run, and the two configurations scored identical candidates in all $14{,}885$ shared iterations. The candidates' mean predicted success was $0.395$ (S15: $0.398$), and the filter raised mixed groups by $9.7$ points $[9.1,10.3]$. Over the $36$ seeds of S15 and S20, filter minus outcome is $+0.0084$ $[0.0043,0.0126]$ (higher in $27$), random choice minus the filter $-0.0091$ $[-0.0135,-0.0048]$ (lower in $28$), and random choice minus outcome $-0.0007$ $[-0.0049,+0.0034]$. Two outcome-only runs (seeds 30 and 32) drifted, the first drift on this model. In strict-verifier accuracy, those two seeds put the filter ahead by $+0.247$ and $+0.223$, which makes the strict-verifier difference $+0.028$ $[0.000,0.055]$ ($p=.049$, secondary). In default-verifier accuracy, which the primary contrast uses, the same seeds differ by $+0.006$ and $+0.003$.

\subsubsection{Further test problems on Qwen2.5-1.5B (S33)}
In S15 and S20 the crossed bootstrap intervals of the gain reached zero, so S33 asked whether the gain carries over to other test problems. It repeated the S20 filter and outcome-only configurations on fresh seeds 36--59 with one addition: at the initial evaluation and at the first evaluation at or after $B$, each run also scored the two further test sets described under Pools and evaluation. The margin was $\pm0.01$. Holm adjustment covered two primary contrasts, filter minus outcome-only GRPO at $B$:
\begin{itemize}
\item On the $1{,}543$ further MATH test problems: $+0.0084$ $[0.0054,0.0114]$ (Holm $p=1.3\times10^{-5}$; higher in $23$ of $24$ seeds; crossed bootstrap $[0.004,0.012]$).
\item On the $200$ problems: $+0.0067$ $[0.0006,0.0129]$ (Holm $p=.034$; higher in $15$ of $24$; crossed $[-0.004,+0.017]$).
\end{itemize}
Neither is equivalent to zero within $\pm0.01$ (TOST $p=.14$ for each). Secondary results:
\begin{itemize}
\item GSM8K, with no directional prediction: $+0.0112$ $[0.0059,0.0164]$ ($p<.001$; higher in $19$ of $24$; crossed $[0.005,0.017]$). From $0.596$ for both configurations at the start, the filter reached $0.717$ and outcome-only GRPO $0.706$.
\item Strict-verifier accuracy: $-0.001$ $[-0.021,+0.018]$ on the further MATH problems, $-0.004$ $[-0.024,+0.017]$ on the $200$, and $-0.020$ $[-0.066,+0.027]$ on GSM8K. One filter run (seed 57) was drifted at $B$ on both MATH sets, with default-verifier accuracy $0.40$ and strict-verifier accuracy $0.19$ on the further problems, and on GSM8K so was a second (seed 40); no outcome-only run drifted. Without the drifted pairs (exploratory), the strict contrasts are $+0.008$ $[0.003,0.013]$, $+0.006$ $[-0.002,+0.014]$, and $+0.012$ $[0.006,0.017]$, and the default ones $+0.009$, $+0.007$, and $+0.012$.
\item Over the first half of the budget, the filter led by $0.020$ $[0.017,0.023]$ on the $200$, in every seed, and it raised mixed groups by $9.7$ points $[9.4,10.1]$, in every seed.
\end{itemize}
Descriptive: at the initial evaluation, both configurations of every seed scored the same on every set. On the further MATH problems, accuracy rose from $0.212$ to $0.407$ for outcome-only GRPO and $0.415$ for the filter. The gain was $+0.009$ $[0.004,0.013]$ on level 3 ($586$ problems), $+0.009$ $[0.005,0.013]$ on level 4 ($555$), and $+0.007$ $[0.003,0.011]$ on level 5 ($402$). Per seed, the gain on the further problems minus that on the $200$ was $+0.002$ $[-0.005,+0.008]$. The candidates' mean predicted success was $0.398$, and the chosen prompts' $0.460$. Over the $60$ seeds of S15, S20, and S33, the filter minus outcome-only GRPO on the $200$ is $+0.0078$ $[0.0043,0.0112]$ (higher in $42$; exploratory). The gain therefore carries over to other test problems from the same distribution.

\begin{figure}[t]
\centering
\includegraphics[width=\columnwidth]{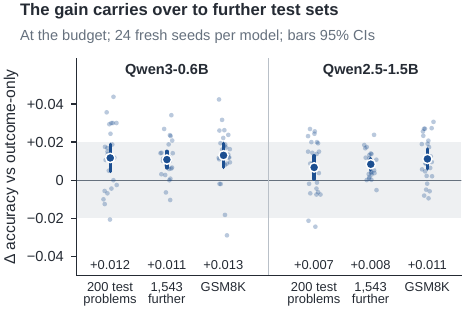}
\caption{\textbf{The prompt filter's gain carries over to further test sets.} Critic filter minus outcome-only GRPO at the token budget, on the same runs: the $200$ test problems of every study, the $1{,}543$ numeric-answer MATH test problems of levels 3--5 that no study had evaluated on, and the $1{,}319$ GSM8K test problems, on Qwen3-0.6B (S34) and Qwen2.5-1.5B (S33), $24$ fresh seeds each. Dots are seeds, bars $95\%$ paired-$t$ CIs, and the gray band marks $\pm0.02$.}
\label{fig:transfer}
\end{figure}

\subsubsection{Further test problems on Qwen3-0.6B (S34)}
S34 is the companion of S33 on the other model with a replicated gain. It repeated the S12 Qwen3-0.6B filter and outcome-only configurations on the level 3--5 pool, on fresh seeds 42--65 (earlier Qwen3-0.6B studies used seeds 0--41), with S33's one addition: at the initial evaluation and at the first evaluation at or after $B$ ($17.25$M), each run also scored the two further test sets. The margin was $\pm0.02$. Holm adjustment covered two primary contrasts, filter minus outcome-only GRPO at $B$:
\begin{itemize}
\item On the $1{,}543$ further MATH test problems: $+0.0108$ $[0.0064,0.0152]$ (Holm $p=7\times10^{-5}$; higher in $21$ of $24$ seeds and lower in $2$; crossed bootstrap $[0.006,0.016]$).
\item On the $200$ problems: $+0.0117$ $[0.0045,0.0188]$ (Holm $p=.003$; higher in $16$ of $24$ and lower in $7$; crossed $[0.0022,0.0215]$).
\end{itemize}
Both predictions held (\Cref{fig:transfer}). Secondary results:
\begin{itemize}
\item GSM8K, with no directional prediction: $+0.0131$ $[0.0067,0.0194]$ ($p<.001$; higher in $20$ of $24$; crossed $[0.006,0.019]$). From $0.579$ for both configurations at the start, the filter reached $0.673$ and outcome-only GRPO $0.660$.
\item Strict-verifier accuracy: $+0.017$ $[0.000,0.034]$ on the further MATH problems ($p=.049$), $+0.017$ $[0.002,0.031]$ on the $200$ ($p=.027$), and $+0.041$ $[-0.017,+0.098]$ on GSM8K ($p=.16$). One run of each configuration was drifted at $B$ on both MATH sets (outcome-only seed 46 and filter seed 52), and on GSM8K two of each.
\item Over the first half of the budget, the filter led by $0.009$ $[0.006,0.013]$ on the $200$ (higher in $20$ of $24$), and it raised mixed groups by $11.1$ points $[10.6,11.6]$, in every seed.
\end{itemize}
Descriptive: at the initial evaluation, both configurations of every seed scored the same on every set. On the further MATH problems, accuracy rose from $0.375$ to $0.469$ for outcome-only GRPO and $0.480$ for the filter. The gain was $+0.008$ $[0.003,0.013]$ on level 3 ($586$ problems), $+0.012$ $[0.008,0.017]$ on level 4 ($555$), and $+0.013$ $[0.007,0.018]$ on level 5 ($402$). Per seed, the gain on the further problems minus that on the $200$ was $-0.001$ $[-0.007,+0.005]$. The candidates' mean predicted success was $0.42$, and the chosen prompts' $0.47$. Pooled with seeds 0--23 as preregistered, filter minus outcome-only GRPO on the $200$ is $+0.017$ $[0.012,0.022]$ ($48$ seeds, higher in $38$; exploratory), and over all $54$ seeds on this model, $+0.017$ $[0.012,0.021]$ (higher in $42$). The replication's gain is smaller than the first studies' $+0.021$ on seeds 0--29, and at equal training time it is $+0.004$ $[-0.001,+0.010]$, not significant, against $+0.015$ $[0.008,0.022]$ on seeds 0--29 (exploratory; \Cref{tab:time}). The gain on this model also carries over to the further problems.

\subsubsection{A filter without a critic on Qwen2.5-1.5B (S22)}
S22 asked whether the Qwen2.5-1.5B gain needs a learned critic. Its one new configuration is the S15/S20 filter configuration with the history estimate in place of the critic: the same candidates, rule, warm-up, and random picks. It ran on the same $36$ seeds, paired with the S15 and S20 runs of each seed. The history filter has no per-prompt estimate until a prompt has been trained. A filter run scores about $15{,}200$ candidates before $B$, $4.4$ passes through the $3{,}458$ prompts of the stream, so its first pass ends near $0.23B$. Holm adjustment covered two primary contrasts:
\begin{itemize}
\item History filter minus outcome, with no directional prediction: $+0.0060$ $[0.0011,0.0110]$ (Holm $p=.035$; higher in $22$ of $36$; crossed bootstrap $[-0.002,+0.014]$).
\item Critic filter minus history filter, predicted positive: $+0.0024$ $[-0.0020,+0.0068]$ (Holm $p=.27$; higher in $21$ of $36$; crossed $[-0.006,+0.010]$), equivalent to zero within $\pm0.02$ (TOST $p<10^{-9}$).
\end{itemize}
Secondary results:
\begin{itemize}
\item Over the first half of the budget, the critic filter led the history filter by $0.019$ $[0.016,0.021]$ on average, in all $36$ seeds. By evaluation, its lead was $+0.020$ at $B/8$, peaked at $+0.026$ at $B/4$, and fell to $+0.002$ at $B$. Over the same half, the history filter led outcome-only GRPO by $+0.002$ $[-0.001,+0.004]$.
\item The history filter led random choice from its candidates by $+0.007$ $[0.002,0.012]$ ($p=.010$).
\item Strict-verifier accuracy: $+0.017$ $[-0.001,+0.035]$ over outcome-only, carried by the two drifted outcome-only runs ($+0.236$ and $+0.204$ on seeds 30 and 32).
\item The history filter raised mixed groups by $4.8$ points $[4.4,5.2]$, higher in every seed, about half the critic filter's $9.8$ on the same seeds.
\end{itemize}
Descriptive: its candidates' estimated success averaged $0.33$, and its chosen prompts' $0.37$. Its mean training reward before $B$ was $0.40$, against $0.37$ for outcome-only GRPO and $0.46$ for the critic filter. It trained on $3{,}444$ distinct prompts per run on average, against the critic filter's $2{,}605$. A per-prompt success estimate near $1/2$ thus carries the gain at $B$ without a learned critic. The interval on the critic filter's advantage at $B$, $[-0.002,+0.007]$, does not exclude a difference nearly as large as the critic filter's own gain ($+0.008$). On this model, the critic's clearest contribution is the early lead.

\subsection{Where does prompt selection help?}
\label{app:m-where}

The filter changes the trained groups in every setting: it raises the share of mixed-outcome groups, the only groups with a nonzero GRPO advantage, by $8$ to $19$ points. Accuracy, however, rises in only some settings, so the composition of the groups does not predict the gain. On Qwen3-0.6B, the filter aimed at predicted success $0.85$ trains on about the same share of mixed groups as outcome-only GRPO ($+0.5$ points, compared with $+0.3$ for random choice) and still gains as much as the filter aimed at $1/2$ (\Cref{tab:matched-b}). The share of all-wrong groups does not predict the gain either: on Qwen3-0.6B the target $0.85$ cuts it from $36\%$ to $15\%$ and gains, while on Qwen3-1.7B it cuts it from $25\%$ to $10\%$ and gains nothing ($+0.001$, $n=6$).

What separates the settings is how hard the candidates are. In the five settings from which we formed this account, post hoc, the critic's mean predicted success over the $24$ candidates was $0.42$ and $0.40$ where the filter helped and $0.54$ to $0.70$ where it did not, with no overlap between seeds (\Cref{tab:account}). When the candidates' mean prediction is below $1/2$, keeping those nearest $1/2$ moves training toward easier prompts and away from all-wrong groups; when it is above $1/2$, the filter moves training toward harder prompts. The training reward shows the move. Wherever the filter helped, its mean training reward exceeded outcome-only GRPO's ($0.43$--$0.47$ against $0.30$--$0.39$), and in the six settings without a gain whose candidates averaged above $0.53$, it fell below. Outcome-only GRPO's own training reward, which needs no critic, separates the first eleven settings in the same way, whereas the base model's accuracy on its first training prompts, known before training, does not.

Six preregistered studies then tested the account (\Cref{tab:tests}). Two moved the candidates across $1/2$ by changing the pool (S19, S21), two changed the model family (S24, S27), and two kept model and pool fixed and reversed only the direction of the move, by aiming the filter at $0.25$ (S30, S31). Five of the six predictions held. The gain predicted on Llama-3.2-1B did not appear at $B$, although the filter led at every evaluation after the first. Across all twelve settings, the filter gained significantly in four of the five settings whose candidates averaged below $1/2$ and in none of the seven above. Candidates below $1/2$ thus mark where a gain is possible, and on a fixed model and pool the direction of the move decides whether it appears.

\subsubsection{Candidate difficulty and a test on Qwen3-1.7B with level 5 (S19)}
Averaged over the iterations before $B$ in which the critic was in use, the critic's mean predicted success over the $24$ candidates separated, in the five settings from which the account was formed, the settings in which the filter helped from those in which it did not (S27 later found a setting below $1/2$ without a significant gain). It is $0.419$ (Qwen3-0.6B, $54$ seeds, range $0.404$--$0.430$) and $0.397$ (Qwen2.5-1.5B, $60$ seeds, $0.383$--$0.409$) where the filter helped, against $0.587$ (Qwen3-1.7B, $0.578$--$0.596$), $0.544$ (Qwen3-4B on level 5, $0.533$--$0.558$), and $0.701$ (Qwen3-4B on levels 3--5, $0.688$--$0.717$) where it did not. Below $1/2$, the filter moves training toward easier prompts and away from all-wrong groups ($36\%\to24\%$ on Qwen3-0.6B, $39\%\to27\%$ on Qwen2.5-1.5B). Above $1/2$, it moves training toward harder prompts and mainly removes all-correct groups. The level-5 study on 4B did not test this account, because 4B's candidates there were still above $1/2$ on average. S19, preregistered after this observation, trains Qwen3-1.7B, the model without a gain on levels 3--5, on the level-5 pool and test set ($24$ seeds). Its precondition is a candidate mean below $1/2$, and its primary prediction is a gain.

The precondition held: the candidates' mean predicted success was $0.389$ (seed range $0.377$--$0.407$), and the chosen prompts' $0.457$. The single primary contrast, filter minus outcome-only at $B$, was $+0.0112$ $[0.0020,0.0205]$ ($p=.019$; higher in $17$ of $24$; crossed bootstrap $[-0.0003,+0.024]$). Its $90\%$ interval, $[0.004,0.019]$, also lies inside $\pm0.02$ (TOST $p=.031$), so the gain is significant but smaller than two points. Strict-verifier accuracy moved the same way ($+0.010$ $[0.001,0.019]$, $p=.025$, secondary). As on Qwen3-0.6B, the filter removed mainly all-wrong groups ($45\%\to30\%$; all-correct $17\%\to19\%$) and raised mixed groups by $12.7$ points $[12.1,13.3]$, higher in every seed. The lead rose to $+0.018$ at $5B/16$ and stayed between $+0.007$ and $+0.014$ from $B/2$ to $B$. Training started from accuracy $0.294$ and reached $0.409$ (outcome-only) and $0.420$ (filter) at $B$. The account thus predicted a gain in a model that had none on levels 3--5, a confirmatory test of an account formed post hoc on five settings.

\subsubsection{A test in the other direction: Qwen3-0.6B with levels 1--2 (S21)}
S21, preregistered after S19 and S20, tests the account where it predicts no gain, at the model where the filter helped most. Both Qwen3-0.6B configurations trained on the $1{,}367$ MATH train problems of levels 1--2 with numeric answers (after removing $26$ that overlap an evaluation set) and were evaluated on $200$ level 1--2 test problems, with the same budget as on levels 3--5 ($B=17.25$M) and seeds 0--23. The precondition was a candidate mean above $1/2$. The single primary contrast compares the filter's gain on levels 1--2 with its gain on levels 3--5 (seeds 0--17 of S8 and S12, Welch $t$, unpaired because the two sides use different runs, pools, and test sets); the account predicts it to be negative.

The precondition held: the candidates' mean predicted success was $0.791$ (seed range $0.771$--$0.803$), and the chosen prompts' $0.656$. Outcome-only GRPO rose from $0.706$ to $0.809$ at $B$, below the preregistered headroom bound of $0.9$. The primary contrast was $-0.030$ $[-0.039,-0.020]$ (Welch $p=5\times10^{-7}$, $32$ df): the level 1--2 gain, $-0.005$, against $+0.024$ on levels 3--5. Secondary results:
\begin{itemize}
\item Filter minus outcome-only on levels 1--2: $-0.005$ $[-0.011,+0.000]$ ($p=.071$; lower in $14$ of $24$; crossed bootstrap $[-0.014,+0.004]$), equivalent to zero within $\pm0.02$ (TOST $p<10^{-5}$). Strict-verifier accuracy: $-0.008$ $[-0.021,+0.006]$.
\item The filter raised mixed groups by $19.1$ points $[18.4,19.8]$, higher in every seed, by removing mainly all-correct groups ($57\%\to33\%$; all-wrong $4\%\to9\%$). Its mean training reward was $0.67$ against outcome-only's $0.81$.
\end{itemize}
Exploratory: the filter trailed from the first evaluation; over the first half of the budget its mean difference was $-0.010$ $[-0.014,-0.006]$ (higher in $3$ of $24$ seeds), and at $B$ it had not reached outcome-only GRPO's accuracy (\Cref{tab:tokens}). Pass@8 differed by $+0.005$ $[-0.003,+0.012]$. The critic's predictions on the trained prompts were $1.2$ points below their realized success (\Cref{tab:calib}), which does not change the side of $1/2$ on which the candidates fall. The account's prediction thus held in its second direction: the model that gains on levels 3--5 does not gain where its candidates average above $1/2$. With S19, this gives two confirmatory tests, one in each direction, of an account formed post hoc.

\subsubsection{Outside the Qwen family: Llama-3.2-3B (S24)}
S24 trains Llama-3.2-3B-Instruct on the level 3--5 pool and evaluates it on the level 3--5 test set (seeds 0--23). A calibration stage, which ran before the preregistration and computed no contrast, applied the S1 rules: $33\%$ of the base model's samples were truncated at $380$ tokens, so the limit is $768$ tokens, and the budget rule gives $B=8.25$M. Because the default verifier's fallback had let outcome-only GRPO drift on three of the four Qwen models, both configurations train with the strict verifier, as S17 and S18 support; held-out evaluation is unchanged. The preregistered prediction depends on the candidates' mean predicted success. Below $1/2$, the account predicts a gain, tested by the single primary contrast, filter minus outcome-only at $B$. Above $1/2$, it predicts none, and a gain would contradict it.

The candidates averaged $0.678$ (seed range $0.638$--$0.718$), and the chosen prompts $0.585$, so the account predicts no gain. The model did much better on the training pool than on the test set. Outcome-only GRPO's training reward was $0.378$ in iterations 0--2 and averaged $0.698$ ($0.663$--$0.722$) over the iterations before $B$, while its held-out accuracy rose from $0.288$ (strict $0.238$) to $0.434$. On the four Qwen models with the same pool, the training reward of iterations 0--2 was $0.01$ to $0.11$ below the initial held-out accuracy. The primary contrast was $-0.0037$ $[-0.0098,+0.0025]$ ($p=.23$; higher in $10$ of $24$ seeds; crossed bootstrap $[-0.013,+0.006]$), equivalent to zero within $\pm0.02$ (TOST $p=7\times10^{-6}$). Secondary results:
\begin{itemize}
\item Strict-verifier accuracy: $-0.003$ $[-0.009,+0.003]$. Mean over the first half of the budget: $+0.004$ $[-0.001,+0.008]$ ($p=.088$).
\item The filter raised mixed groups by $9.1$ points $[8.3,9.8]$, higher in every seed, mainly by removing all-correct groups ($35\%\to22\%$; all-wrong $8\%\to12\%$). Its mean training reward was $0.591$.
\end{itemize}
Exploratory: the difference peaked at $+0.007$ at $B/8$ and was negative from $9B/16$ on. Pass@8 was $0.011$ $[0.0003,0.023]$ lower (\Cref{tab:passk}). The critic's predictions on the trained prompts were $0.9$ points below their realized success (\Cref{tab:calib}). The account predicted no gain on this side of $1/2$, and none was found; S27 tests whether the filter can help outside the Qwen family.

\subsubsection{The gain prediction outside the Qwen family: Llama-3.2-1B (S27)}
S27 trains Llama-3.2-1B-Instruct, the smaller model of the same family, with both configurations on the strict verifier as in S24 (seeds 0--23). Two stages ran before the preregistration and computed no contrast. A calibration stage applied the S1 rules: $37.5\%$ of the base model's samples were truncated at $380$ tokens, so the limit is $768$ tokens, and the budget rule gives $B=10.0$M. At $768$ tokens the base model's accuracy on the level 3--5 test set was $0.147$ (strict $0.075$). A pilot then chose the training pool by a rule fixed with the calibration. Outcome-only GRPO was trained to $B$ with a seed outside the study, on the level 3--5 pool and on the level-5 pool. The study would use the level 3--5 pool if that pilot's mean training reward over the iterations before $B$ was below $0.45$, and otherwise the level-5 pool under the same condition. The level 3--5 pilot's training reward was $0.387$, so the study uses that pool and its test set. The precondition is a candidate mean below $1/2$; the single primary contrast, filter minus outcome-only at $B$, is then predicted to be positive, and a contrast equivalent to zero or negative counts against the account.

The candidates averaged $0.420$ (seed range $0.395$--$0.443$), and the chosen prompts $0.467$, so the precondition held. Outcome-only GRPO's training reward, $0.394$ ($0.364$--$0.423$), was also below $1/2$. The primary contrast was $+0.0044$ $[-0.0008,+0.0096]$ ($p=.095$; higher in $16$ of $24$ seeds; crossed bootstrap $[-0.0045,+0.0136]$), equivalent to zero within $\pm0.02$ (TOST $p=1\times10^{-6}$). At the per-pair spread observed ($0.012$), the study had power $0.86$ for a gain of $+0.008$, the size found on Qwen2.5-1.5B. Secondary results:
\begin{itemize}
\item Strict-verifier accuracy: $+0.003$ $[-0.003,+0.009]$. Mean over the first half of the budget: $+0.005$ $[0.003,0.008]$ ($p<.001$; higher in $18$ of $24$), as predicted.
\item The filter raised mixed groups by $8.8$ points $[8.2,9.5]$, higher in every seed, mainly by removing all-wrong groups ($28\%\to18\%$; all-correct $10\%\to11\%$). Its mean training reward was $0.464$.
\end{itemize}
Exploratory: the difference lay between $+0.001$ and $+0.009$ at every evaluation after the first. Averaged over the second half of the budget it was $+0.007$ $[0.002,0.011]$ ($p=.004$), and over all sixteen evaluations $+0.006$ $[0.003,0.009]$. Outcome-only GRPO rose from $0.148$ to $0.261$ at $B$, by only $0.009$ in the second half, so the filter reached its accuracy at $B$ after $0.59$ of the budget (\Cref{tab:tokens}). Pass@8 differed by $+0.003$ $[-0.011,+0.016]$ (\Cref{tab:passk}). The critic's predictions on the trained prompts were $0.08$ points below their realized success (\Cref{tab:calib}). The account's prediction of a gain at $B$ thus failed outside the Qwen family. The filter's mean lead was small but positive at every evaluation, significant only when averaged over half the budget, and the interval at $B$ includes the Qwen2.5-1.5B gain.

\subsubsection{The filter aimed at harder prompts on a fixed pool (S30)}
The tests of the account above changed the pool (S19, S21) or the model (S24, S27). S30 keeps the model and pool where the filter gains most, Qwen3-0.6B on levels 3--5, and changes only the filter's target, from $1/2$ to $0.25$, so that the filter trains on prompts harder than its candidates while still avoiding those predicted near $0$ (seeds 0--23, $10^{-6}$, default reward, as the comparators). The comparators are the filter at $1/2$ and outcome-only GRPO on the same seeds (S8, S12, and S16 read at $B$). Two primary contrasts, Holm-adjusted, were preregistered: T1, the filter at $0.25$ minus the filter at $1/2$, predicted negative, and T2, the filter at $0.25$ minus outcome-only GRPO, predicted to show no gain. A manipulation check required the chosen prompts' predicted success to lie at least $0.05$ below the candidates' on average and the filter's training reward to lie below outcome-only GRPO's.

The manipulation check passed. The chosen prompts' predicted success was $0.287$ against the candidates' $0.359$, between $0.064$ and $0.080$ lower in every seed (at target $1/2$: $0.468$ against $0.419$). The training reward was $0.287$ ($0.279$--$0.292$), against $0.386$ for outcome-only GRPO and $0.468$ for the filter at $1/2$, lower in every seed. All-wrong groups rose from $35.9\%$ to $41.5\%$ and all-correct groups fell from $14.1\%$ to $6.8\%$, so mixed groups rose by only $1.7$ points $[1.3,2.2]$. Both primary predictions held. T1 was $-0.0280$ $[-0.0359,-0.0201]$ (Holm $p=4\times10^{-7}$; lower in $22$ of $24$ seeds; crossed bootstrap $[-0.041,-0.015]$). T2 was $-0.0060$ $[-0.0154,+0.0033]$ (Holm $p=.20$; crossed bootstrap $[-0.019,+0.007]$), equivalent to zero within $\pm0.02$ (TOST $p=.003$). Secondary results: strict-verifier accuracy $-0.003$ $[-0.019,+0.012]$ against outcome-only GRPO; over the first half of the budget, $-0.002$ $[-0.004,+0.001]$ against outcome-only GRPO, so there was no head start. In strict-verifier accuracy the filter at $0.25$ trailed the filter at $1/2$ by $0.028$ $[0.013,0.043]$ (exploratory). The filter at $0.25$ drifted in $9$ of $408$ evaluations, in two runs, against $1$ of $414$ (outcome-only GRPO) and $0$ of $414$ (filter at $1/2$) up to $B$ on the same seeds; one run (seed 4) was drifted at $B$. Without it, T1 is $-0.028$ $[-0.036,-0.020]$ and T2 $-0.005$ $[-0.015,+0.005]$ (exploratory). Training took $0.54$ hours per run on H200 GPUs. On a fixed model and pool, the gain thus depended on the direction of the move, as the account says.

\subsubsection{The filter aimed at harder prompts on Qwen2.5-1.5B (S31)}
S31 repeats S30 on the other model with a replicated gain, Qwen2.5-1.5B on levels 3--5 (candidates $0.40$, gain $+0.008$). The configuration is the S20 filter with the target changed from $1/2$ to $0.25$ and nothing else; the comparators are the S20 filter and outcome-only runs of the same seeds (12--35). The primary contrasts, Holm-adjusted, were those of S30: T1, the filter at $0.25$ minus the filter at $1/2$, predicted negative, and T2, the filter at $0.25$ minus outcome-only GRPO, predicted to show no gain. Because the gain under test is $0.008$, the equivalence margin was $\pm0.01$, as in S28. The manipulation check was that of S30, and a secondary contrast predicted no head start over outcome-only GRPO.

The manipulation check passed, by a smaller margin than in S30. The chosen prompts' predicted success was $0.278$ against the candidates' $0.331$, $0.053$ lower on average and between $0.039$ and $0.065$ lower per seed (at target $1/2$: $0.458$ against $0.395$). The training reward was $0.282$ ($0.274$--$0.289$), against $0.367$ for outcome-only GRPO and $0.457$ for the filter at $1/2$, lower in every seed. All-wrong groups rose from $39.1\%$ to $43.2\%$, but all-correct groups fell from $15.8\%$ to $8.0\%$, so mixed groups rose by $3.9$ points $[3.4,4.4]$, in every seed. Both primary predictions held. T1 was $-0.0097$ $[-0.0159,-0.0035]$ (Holm $p=.008$; lower in $18$ of $24$ seeds; crossed bootstrap $[-0.020,+0.0003]$), against a gain of the filter at $1/2$ of $+0.0083$ on these seeds. T2 was $-0.0014$ $[-0.0095,+0.0067]$ (Holm $p=.73$; crossed bootstrap $[-0.012,+0.009]$), equivalent to zero within $\pm0.01$ (TOST $p=.019$) and within $\pm0.02$ ($p<10^{-4}$). The secondary prediction of no head start failed: over the first half of the budget the filter at $0.25$ led outcome-only GRPO by $0.0047$ $[0.0007,0.0087]$ ($p=.023$), a quarter of the filter's lead at $1/2$ ($0.019$), and trailed the filter at $1/2$ by $0.014$ $[0.011,0.017]$, in every seed. In strict-verifier accuracy, T2 was $+0.010$ $[-0.027,+0.047]$ and T1 $-0.018$ $[-0.040,+0.004]$, both widened by drift. The filter at $0.25$ drifted in $5$ of $408$ evaluations, against $16$ (outcome-only GRPO) and $1$ (filter at $1/2$) of $408$ on the same seeds; one of its runs (seed 21) and two outcome-only runs (seeds 30 and 32) were drifted at $B$. Without the filter's drifted run, T1 is $-0.0086$ $[-0.0146,-0.0025]$ and T2 $+0.0007$ $[-0.0064,+0.0079]$; without all three, $-0.0073$ $[-0.0137,-0.0010]$ and $+0.0024$ $[-0.0050,+0.0099]$ (exploratory). Training took $0.49$ hours per run on H200 GPUs. The direction of the move thus mattered on a second model; the early lead shows that training on prompts harder than the candidates, while still avoiding those predicted near $0$, kept a small part of the filter's head start on this model, but none of its gain at $B$.

\subsubsection{Candidates, calibration, and training reward (exploratory)}
\Cref{tab:account} lists every setting with a critic-filter configuration and an outcome-only configuration on the same seeds, ordered by the candidates' mean predicted success. The first studies on the level 4--5 pool (S6, S7; $6$--$8$ seeds) were not among the five settings on which the account of \Cref{sec:q4} was formed, and S19, S21, S24, and S27 tested it; we add all of them here. Training rewards are means over the iterations started before $B$, under the verifier that trained the configuration (strict for the Llama models, default otherwise). In all five settings below $1/2$ the filter raised the training reward; it gained significantly in four, and on Llama-3.2-1B its $+0.004$ $[-0.001,+0.010]$ was not significant. In the six settings above $0.53$ it lowered the training reward (by $0.01$ on Qwen3-4B on level 5, by $0.05$ to $0.15$ in the others) and did not gain; Qwen3-1.7B on levels 4--5 sits at $0.52$, with the filter's training reward $0.01$ above outcome-only GRPO's and an inconclusive estimate. Outcome-only GRPO's training reward, which needs no critic, orders the Qwen settings as the candidates' predicted success does. In the first eleven settings it separated the seeds of the settings with and without a gain without overlap; on Llama-3.2-1B it fell in the range of the settings with a gain, so it, too, predicted the gain that S27 did not find. The rewards of the first iteration, before any update, do not separate the settings: Qwen3-4B on level 5 ($0.30$) falls inside the range of the settings with a gain ($0.18$--$0.35$), Llama-3.2-3B ($0.39$) just above it, and Llama-3.2-1B ($0.13$) below it. Over a whole budget, outcome-only GRPO's training reward reflects how quickly the model learns the pool, which its first iteration does not.

\begin{table*}[t]\centering
\caption{\textbf{Where the critic filter helps} (exploratory). Candidates: the critic's mean predicted success of the $24$ candidates per iteration (seed mean, with the range over seeds). Training reward: mean group reward over the iterations started before $B$ (strict verifier for the Llama models, whose configurations train on it). Iteration 0: training reward of the first iteration, before any update, the same in both configurations (the base model on its first $8$ training prompts per seed). Start: outcome-only GRPO's accuracy on the test set before training. Last column: filter minus outcome-only accuracy at $B$ ($95\%$ paired-$t$ CI).}
\label{tab:account}
\setlength{\tabcolsep}{4pt}\small
\begin{tabular}{lccccccc}\toprule
 & & & \multicolumn{2}{c}{training reward} & & & \\\cmidrule(lr){4-5}
Setting & $n$ & candidates & filter & outcome-only & iteration 0 & start & filter $-$ outcome \\\midrule
Qwen3-0.6B, L4--5 & $6$ & $0.34$ (0.32--0.35) & $0.43$ & $0.30$ (0.30--0.31) & $0.18$ & $0.36$ & $+0.028\,[+0.007,+0.050]$ \\
Qwen3-1.7B, level 5 & $24$ & $0.39$ (0.38--0.41) & $0.45$ & $0.35$ (0.34--0.36) & $0.28$ & $0.29$ & $+0.011\,[+0.002,+0.020]$ \\
Qwen2.5-1.5B, L3--5 & $60$ & $0.40$ (0.38--0.41) & $0.46$ & $0.37$ (0.35--0.38) & $0.21$ & $0.22$ & $+0.008\,[+0.004,+0.011]$ \\
Qwen3-0.6B, L3--5 & $54$ & $0.42$ (0.40--0.43) & $0.47$ & $0.39$ (0.37--0.40) & $0.35$ & $0.35$ & $+0.017\,[+0.012,+0.021]$ \\
Llama-3.2-1B, L3--5 & $24$ & $0.42$ (0.39--0.44) & $0.46$ & $0.39$ (0.36--0.42) & $0.13$ & $0.15$ & $+0.004\,[-0.001,+0.010]$ \\\midrule
Qwen3-1.7B, L4--5 & $6$ & $0.52$ (0.51--0.52) & $0.51$ & $0.50$ (0.49--0.51) & $0.32$ & $0.56$ & $+0.017\,[-0.007,+0.040]$ \\
Qwen3-4B, level 5 & $12$ & $0.54$ (0.53--0.56) & $0.51$ & $0.52$ (0.49--0.54) & $0.30$ & $0.35$ & $+0.002\,[-0.028,+0.032]$ \\
Qwen3-1.7B, L3--5 & $24$ & $0.59$ (0.58--0.60) & $0.53$ & $0.58$ (0.56--0.59) & $0.51$ & $0.55$ & $+0.002\,[-0.006,+0.011]$ \\
Llama-3.2-3B, L3--5 & $24$ & $0.68$ (0.64--0.72) & $0.59$ & $0.70$ (0.66--0.72) & $0.39$ & $0.29$ & $-0.004\,[-0.010,+0.002]$ \\
Qwen3-4B, L4--5 & $8$ & $0.68$ (0.67--0.70) & $0.58$ & $0.66$ (0.65--0.67) & $0.40$ & $0.62$ & $+0.002\,[-0.013,+0.017]$ \\
Qwen3-4B, L3--5 & $12$ & $0.70$ (0.69--0.72) & $0.58$ & $0.71$ (0.69--0.72) & $0.49$ & $0.62$ & $-0.002\,[-0.015,+0.010]$ \\
Qwen3-0.6B, L1--2 & $24$ & $0.79$ (0.77--0.80) & $0.67$ & $0.81$ (0.80--0.83) & $0.74$ & $0.71$ & $-0.005\,[-0.011,+0.000]$ \\
\bottomrule\end{tabular}
\end{table*}

\Cref{tab:calib} checks the critic's predictions against the groups it trained on. Random choice trains on a uniform draw of the candidates, so its realized minus predicted success measures calibration on the candidate distribution: it is within $0.2$ points on all three models. On level 1--2 problems and on Llama-3.2-3B, which have no random-choice configuration, the filter's trained prompts succeeded $1.2$ and $0.9$ points more often than predicted; there the critic, if anything, underestimated how far above $1/2$ they lay. On Llama-3.2-1B they succeeded as often as predicted ($+0.08$ points $[-0.05,+0.21]$), so its candidates' side of $1/2$ is not an artifact of calibration. A prompt's prediction correlates with its group's success at $0.67$ to $0.72$, although a group has only eight rollouts. On the filter's chosen prompts, which lie near $1/2$, the correlation is lower because their range is narrower. The threshold of $1/2$ in the account is therefore a threshold on realized success, not only on the critic's scale.

\begin{table*}[t]\centering
\caption{\textbf{Calibration of the prompt critic} (exploratory), on the prompts trained in the iterations before $B$ in which the critic was in use: mean predicted and realized group success, their difference per seed in points ($95\%$ CI), and the correlation over groups.}
\label{tab:calib}
\setlength{\tabcolsep}{2.5pt}\small
\begin{tabular}{llccccc}\toprule
Model & configuration & $n$ & pred. & real. & real.\,$-$\,pred. & $r$ \\\midrule
Qwen3-0.6B & random & $18$ & $0.394$ & $0.393$ & $-0.14\,[-0.39,+0.10]$ & $0.67$ \\
 & filter & $18$ & $0.469$ & $0.468$ & $-0.05\,[-0.25,+0.14]$ & $0.47$ \\
Qwen3-1.7B & random & $24$ & $0.579$ & $0.578$ & $-0.13\,[-0.26,0.00]$ & $0.72$ \\
 & filter & $24$ & $0.533$ & $0.530$ & $-0.24\,[-0.39,-0.09]$ & $0.50$ \\
Qwen2.5-1.5B & random & $36$ & $0.366$ & $0.368$ & $+0.17\,[+0.03,+0.31]$ & $0.68$ \\
 & filter & $36$ & $0.458$ & $0.459$ & $+0.08\,[-0.05,+0.22]$ & $0.48$ \\
Qwen3-0.6B, L1--2 & filter & $24$ & $0.656$ & $0.668$ & $+1.21\,[+1.11,+1.32]$ & $0.56$ \\
Llama-3.2-3B & filter & $24$ & $0.585$ & $0.594$ & $+0.92\,[+0.79,+1.06]$ & $0.46$ \\
Llama-3.2-1B & filter & $24$ & $0.467$ & $0.468$ & $+0.08\,[-0.05,+0.21]$ & $0.44$ \\
\bottomrule\end{tabular}
\end{table*}

\subsection{Does the gain last, and can retuning match it?}
\label{app:m-last}

Where we followed the gain over time, it was a head start (\Cref{fig:selection-time}). On Qwen3-0.6B, it peaked at $5B/8$ and had mostly faded by $2B$ (S16, S17). On Qwen2.5-1.5B ($60$ seeds), it peaked at $+0.026$ $[0.022,0.030]$ at $3B/16$ and fell to $+0.008$ at $B$. On Qwen3-4B with level 5, it peaked at $+0.042$ $[0.026,0.057]$ at $B/8$ (S14). A head start is a saving in compute: where the filter gains, it reaches outcome-only GRPO's accuracy at $B$ with $22$ to $23\%$ fewer generated tokens and $13$ to $17\%$ less training time (\Cref{tab:tokens}). The Qwen3-0.6B gain also holds at pass@8 ($+0.009$ $[0.002,0.017]$, $n=54$), which cuts the share of test problems that no sample solves from $27.1\%$ to $26.2\%$, so the filter does more than sharpen sampling on problems the policy already solves (\Cref{tab:passk}).

Because the gain is a head start, a faster baseline is its natural rival, and a doubled or decaying learning rate tests it directly. On Qwen3-0.6B, doubling outcome-only GRPO's learning rate buys as much as the filter, with and without drift (S23, S29). On Qwen2.5-1.5B, a doubled constant rate lowers both configurations and the filter's gain holds (S28), while a rate decaying from $2\times10^{-6}$ to $0$ at $B$ brings outcome-only GRPO within half a point of the filter and removes the filter's own gain at $B$ (S32). The filter's gain is thus about as large as what tuning the baseline's learning rate buys, which is why a claimed gain should be measured against a tuned baseline.

\subsubsection{Twice the budget (S16)}
On Qwen3-0.6B, the filter's lead grew over the budget ($+0.011$ at $B/4$, $+0.020$ at $B/2$, and $+0.024$ at $B$ over seeds 0--17), and outcome-only accuracy was still rising at $B$. S16 trained both configurations to $2B$ on fresh seeds 18--29. Because the learning rate is constant, the first $B$ tokens of each run are trained exactly as in the configurations with budget $B$. The primary contrast at $2B$ was $+0.021$ $[-0.001,+0.042]$ ($p=.059$; crossed $[-0.001,+0.044]$; higher in $8$ of $12$). At $B$ it was $+0.015$ $[0.003,0.026]$ ($p=.019$), and the change between $B$ and $2B$ was $+0.006$ $[-0.009,+0.021]$. From $B$ to $2B$, outcome-only accuracy rose from $0.444$ to $0.487$ and the filter's from $0.459$ to $0.507$. Strict-verifier accuracy at $2B$ differed by $+0.032$ $[-0.065,+0.130]$, because drifted runs occurred in both configurations. One outcome-only run (seed 25) drifted from $13B/16$ and ended with $99\%$ of its samples truncated. Its default-verifier accuracy at $2B$ was $0.425$, against $0.519$ for the filter on the same seed. Pooling the evaluations at $B$ of all thirty Qwen3-0.6B seeds (0--29), the filter leads by $+0.021$ $[0.014,0.027]$, higher in $26$ of $30$; adding the $24$ seeds of S34 (above), it leads by $+0.017$ $[0.012,0.021]$ over $54$ seeds.

\subsubsection{Twice the budget with the strict reward (S17)}
Because drift affected S16's outcome-only configuration more often than its filter configuration, S17 repeated both S16 configurations with the strict verifier as the training reward. Seeds 18--29 pair with the S16 runs (same prompt stream and evaluation draws), and seeds 30--41 are fresh. Holm adjustment covered two primary contrasts:
\begin{itemize}
\item Filter minus outcome at $2B$, both on the strict reward ($n=24$): $+0.0053$ $[-0.0051,+0.0157]$ ($p=.302$; crossed bootstrap $[-0.007,+0.018]$; higher in $16$ of $24$; TOST $p=.004$). Seeds 18--29 give $+0.001$ and seeds 30--41 $+0.010$.
\item Outcome-only on the strict minus on the default reward at $2B$ ($n=12$, non-inferiority at $-0.02$): $+0.0228$ $[0.0092,0.0363]$ (Holm $p<.001$; crossed $[0.006,0.040]$; higher in $11$ of $12$).
\end{itemize}
The preregistered secondary contrasts, without correction for multiple comparisons, were as follows. At $B$, the filter led by $+0.019$ $[0.013,0.026]$ ($p<10^{-5}$, higher in $22$ of $24$; crossed $[0.008,0.031]$), and the lead shrank by $0.014$ $[0.004,0.024]$ from $B$ to $2B$ ($p=.011$, lower in $17$ of $24$). Over the $33$ evaluations, the lead peaked at $+0.021$ $[0.014,0.028]$ at $5B/8$. The filter configuration on the strict reward was non-inferior to the same configuration on the default reward ($+0.003$ $[-0.012,+0.019]$, $p=.003$). The filter's $2B$ lead under the strict reward minus its lead under the default reward, on seeds 18--29, was $-0.020$ $[-0.043,+0.004]$ ($p=.099$). Strict-verifier accuracy of outcome-only GRPO at $2B$ rose from $0.420$ to $0.497$ ($+0.077$ $[0.001,0.152]$, $p=.047$), with $+0.412$ and $+0.207$ on seeds 21 and 25, both of which drifted under the default reward. The filter raised the share of mixed-outcome groups trained on before $2B$ by $10.2$ points $[9.6,10.8]$, higher in all $24$ seeds. In strict-verifier accuracy, the filter led by $+0.018$ $[0.011,0.024]$ at $B$ and $+0.004$ $[-0.006,+0.014]$ at $2B$. No strict-reward run drifted ($0$ of $792$ evaluations per configuration), and all $49{,}006$ training iterations of the two configurations used the strict reward. Truncated samples at $2B$ were $28\%$ (filter) and $23\%$ (outcome-only) over the $24$ seeds. On seeds 18--29, outcome-only GRPO's fell from $36\%$ under the default reward to $24\%$ under the strict reward, and both configurations had $36\%$ under the default reward. Training took $1.17$ hours per filter run and $1.07$ per outcome-only run on H200 GPUs. The gain is therefore not shown to persist to $2B$; \Cref{tab:matched} and \Cref{fig:overview} give S16 and S17 side by side.

\subsubsection{Twice the learning rate on Qwen3-0.6B (S23)}
S23 repeats the Qwen3-0.6B level 3--5 configurations, the critic filter and outcome-only GRPO on the default reward, at learning rate $2\times10^{-6}$ on seeds 0--23. Each configuration differs from its $10^{-6}$ counterpart only in the learning rate. The $10^{-6}$ runs of the same seeds (S8, S12, and S16 read at $B$) share the prompt streams and evaluation draws. Primary contrasts (Holm over two; $n=24$):
\begin{itemize}
\item Filter minus outcome-only, both at $2\times10^{-6}$: $+0.009$ $[-0.006,+0.024]$, Holm $p=.46$, higher in $17$ of $24$ seeds; crossed bootstrap $[-0.0076,+0.0259]$; TOST $p=.07$.
\item Outcome-only at $2\times10^{-6}$ minus the filter at $10^{-6}$: $+0.005$ $[-0.007,+0.018]$, Holm $p=.46$, higher in $14$; crossed $[-0.0105,+0.0206]$; equivalent within $\pm0.02$ (TOST $p=.010$).
\end{itemize}
\textbf{Secondary contrasts.} Doubling the learning rate raised outcome-only GRPO's accuracy by $0.027$ $[0.014,0.040]$ ($p<.001$) and the filter's by $0.014$ $[0.003,0.025]$ ($p=.013$). The filter's gain changed by $-0.013$ $[-0.029,+0.003]$ ($p=.11$). Over the first half of the budget, the filter led by $0.014$ $[0.006,0.023]$ ($p=.002$). It raised mixed groups by $9.5$ points $[8.7,10.3]$, higher in every seed. In strict-verifier accuracy, the filter minus outcome-only at $2\times10^{-6}$ was $+0.043$ $[-0.036,+0.121]$.

\textbf{Stability (descriptive and exploratory).} Up to $B$, $49$ of $408$ outcome-only evaluations and $36$ of $408$ filter evaluations at $2\times10^{-6}$ drifted, in $14$ and $8$ runs; $5$ and $3$ runs were drifted at $B$. At $10^{-6}$, the same seeds had one drifted evaluation up to $B$ (outcome-only) and no run drifted at $B$. Truncated samples at $B$ were $43\%$ and $42\%$, against $36\%$ and $35\%$ at $10^{-6}$. Strict-verifier accuracy at $B$ was $0.366$ (outcome-only) and $0.409$ (filter) at $2\times10^{-6}$, against $0.410$ and $0.435$ at $10^{-6}$. In strict-verifier accuracy, outcome-only GRPO at $2\times10^{-6}$ trails the filter at $10^{-6}$ by $0.069$ $[0.005,0.132]$ ($p=.037$). On the $17$ seeds where none of the four runs was drifted at $B$, outcome-only GRPO at $2\times10^{-6}$ minus the filter at $10^{-6}$ is $+0.003$ $[-0.010,+0.016]$ in default and $-0.000$ $[-0.020,+0.019]$ in strict-verifier accuracy. On the same seeds, the filter's gain at $2\times10^{-6}$ is $+0.011$ $[-0.004,+0.027]$, and doubling the learning rate raised outcome-only GRPO's strict-verifier accuracy by $0.026$ $[0.007,0.045]$. Training took $0.56$ hours per filter run and $0.51$ per outcome-only run on H200 GPUs. Doubling the learning rate thus buys as much as the filter on this model.

\subsubsection{Twice the learning rate without drift (S29)}
S23's match rested on default-verifier accuracy, under which most of its faster outcome-only runs drifted, and its drift-free comparison rested on a subset of seeds chosen after the fact. S29 trains outcome-only GRPO at $2\times10^{-6}$ with the strict verifier as the reward, to $B$, on the seeds of the S17 strict-reward runs (18--41), and compares it with both S17 configurations at $10^{-6}$ read at $B$. Because the learning rate is constant, a run to $2B$ read at $B$ is equivalent to a run to $B$. Its two primary contrasts, Holm-adjusted, are M1, outcome-only GRPO at $2\times10^{-6}$ minus the filter at $10^{-6}$, and M2, the same minus outcome-only GRPO at $10^{-6}$; both are two-sided.

No run drifted ($0$ of $408$ evaluations). M1 was $-0.0003$ $[-0.0127,+0.0121]$ (Holm $p=.96$; crossed bootstrap $[-0.018,+0.017]$), equivalent to zero within $\pm0.02$ (TOST $p=.002$) but not within $\pm0.01$ ($p=.059$). M2 was $+0.019$ $[0.005,0.033]$ (Holm $p=.018$, higher in $18$ of $24$ seeds; crossed bootstrap $[0.001,0.037]$), as large as the filter's gain on the same seeds ($+0.019$ $[0.013,0.026]$). The per-seed spread was larger at the faster rate ($0.029$ for M1, against $0.016$ for the filter's gain). Secondary results: in strict-verifier accuracy, M1 was $+0.001$ $[-0.015,+0.016]$ and M2 $+0.018$ $[0.003,0.034]$; over the first half of the budget, outcome-only GRPO at $2\times10^{-6}$ led the filter at $10^{-6}$ by $0.015$ $[0.010,0.019]$ (higher in $22$ of $24$ seeds), so the faster rate gives a larger head start than the filter does. It trained on slightly fewer mixed groups than outcome-only GRPO at $10^{-6}$ ($46\%$ against $47\%$, and $59\%$ for the filter). Truncated samples at $B$ were $32\%$, against $27\%$ (outcome-only, $10^{-6}$) and $32\%$ (filter). Training took $0.53$ hours per run on H200 GPUs. On Qwen3-0.6B, doubling the learning rate thus buys as much as the filter without drift.

\subsubsection{Twice the learning rate on Qwen2.5-1.5B (S28)}
S28 repeats S23 on Qwen2.5-1.5B: it runs both S20 configurations at $2\times10^{-6}$, with nothing else changed, on the S20 seeds (12--35), so that each run shares its prompt stream and evaluation draws with the S20 runs of its seed at $10^{-6}$. On this model the filter's gain is $0.008$, so the preregistered equivalence margin is $\pm0.01$. The two primary contrasts, Holm-adjusted and two-sided, are L1, the filter minus outcome-only GRPO at $2\times10^{-6}$, and L2, outcome-only GRPO at $2\times10^{-6}$ minus the filter at $10^{-6}$.

L1 was $+0.0111$ $[0.0021,0.0201]$ (Holm $p=.018$; higher in $19$ of $24$ seeds; crossed bootstrap $[-0.002,+0.025]$), and L2 was $-0.0399$ $[-0.0468,-0.0331]$ (Holm $p=4\times10^{-11}$; lower in every seed; crossed bootstrap $[-0.056,-0.025]$). Secondary results:
\begin{itemize}
\item Doubling the rate lowered accuracy at $B$ by $0.032$ $[0.024,0.039]$ for outcome-only GRPO and by $0.029$ $[0.020,0.038]$ for the filter, lower in $23$ of $24$ seeds each. The filter's gain did not change with the rate: $+0.003$ $[-0.006,+0.011]$, equivalent to zero within $\pm0.01$ (TOST $p=.046$).
\item Strict-verifier accuracy: L1 $-0.005$ $[-0.040,+0.031]$, L2 $-0.039$ $[-0.050,-0.029]$. Over the first half of the budget, the filter led outcome-only GRPO at $2\times10^{-6}$ by $0.014$ $[0.010,0.019]$, and it raised mixed groups by $7.5$ points $[7.0,8.1]$, higher in every seed.
\end{itemize}
Exploratory: at $2\times10^{-6}$ both configurations rose faster at first. Outcome-only GRPO at $2\times10^{-6}$ led the filter at $10^{-6}$ by $0.033$ to $0.039$ from $B/16$ to $B/4$, reached $0.39$ by $5B/16$, and then stayed between $0.39$ and $0.40$, falling behind from $7B/16$ on. Even its best evaluation (seed mean $0.40$, at $5B/8$) stayed below outcome-only GRPO at $10^{-6}$ at $B$ ($0.42$), so selecting its best checkpoint would not have closed the gap. Both configurations drifted at $2\times10^{-6}$, outcome-only GRPO in $11$ and the filter in $12$ of $408$ evaluations, one run each at $B$; at $10^{-6}$ the same seeds gave $16$ and $1$. Truncated samples at $B$ were $16\%$ and $17\%$, against $20\%$ and $21\%$ at $10^{-6}$. The candidates averaged $0.420$ at $2\times10^{-6}$, against $0.395$ at $10^{-6}$. Training took $0.55$ hours per filter run and $0.48$ per outcome-only run on H200 GPUs. On Qwen2.5-1.5B, the filter's gain is thus not reached by a faster baseline, and it holds at twice the learning rate.

\subsubsection{A decaying learning rate on Qwen2.5-1.5B (S32)}
S28's faster runs rose quickly and then stalled. S32 gives both S28 configurations a learning rate that decays linearly with generated tokens. An iteration that starts after a fraction $f$ of the budget steps at $2\times10^{-6}(1-f)$, so the rate falls from $2\times10^{-6}$ to $0$ at $B$ and averages $10^{-6}$. Nothing else changes. The seeds are S20's (12--35), so each run shares its prompt stream and evaluation draws with the S20 and S28 runs of its seed. The margin is $\pm0.01$. The two primary contrasts, Holm-adjusted and two-sided, are V1, outcome-only GRPO with the decaying rate minus the filter at a constant $10^{-6}$, and V2, the filter minus outcome-only GRPO, both with the decaying rate.

V1 was $-0.0046$ $[-0.0120,+0.0027]$ (Holm $p=.41$; lower in $14$ and higher in $9$ of $24$ seeds; crossed bootstrap $[-0.016,+0.006]$). It is not equivalent to zero within $\pm0.01$ (TOST $p=.073$) but is within $\pm0.02$ ($p=.0001$). V2 was $+0.0019$ $[-0.0053,+0.0091]$ (Holm $p=.59$; higher in $11$ of $24$ seeds; crossed bootstrap $[-0.009,+0.013]$), equivalent to zero within $\pm0.01$ (TOST $p=.015$). Secondary results:
\begin{itemize}
\item The schedule changed outcome-only GRPO by $+0.004$ $[-0.005,+0.012]$ against a constant $10^{-6}$ ($p=.37$) and by $+0.035$ $[0.027,0.044]$ against a constant $2\times10^{-6}$ (higher in $23$ of $24$ seeds). The filter's gain changed by $-0.006$ $[-0.016,+0.003]$ ($p=.17$).
\item Over the first half of the budget, outcome-only GRPO with the decaying rate led the filter at $10^{-6}$ by $0.020$ $[0.017,0.024]$, in every seed. Under the same schedule, the filter led outcome-only GRPO by $0.014$ $[0.010,0.018]$ ($22$ of $24$ seeds) and raised mixed groups by $8.3$ points $[7.8,8.7]$, in every seed.
\item Strict-verifier accuracy: V1 $-0.020$ $[-0.051,+0.011]$, V2 $+0.014$ $[-0.021,+0.049]$.
\end{itemize}
Exploratory results:
\begin{itemize}
\item The filter with the decaying rate minus the filter at $10^{-6}$ was $-0.003$ $[-0.007,+0.002]$, equivalent within $\pm0.01$ (TOST $p=.002$). Seed means at $B$ were $0.429$ for the filter at $10^{-6}$, $0.426$ for the filter with the decaying rate, $0.424$ for outcome-only GRPO with it, and $0.420$ for outcome-only GRPO at $10^{-6}$.
\item Outcome-only GRPO with the decaying rate led the filter at $10^{-6}$ by up to $0.033$, at $B/8$, and trailed it from $13B/16$ on (\Cref{fig:lr}b). Under the schedule, the filter's lead over outcome-only GRPO peaked at $0.024$, at $B/16$ and $B/8$, and stayed between $0.000$ and $0.006$ from $9B/16$ on.
\item Drift: outcome-only GRPO drifted in $16$ and the filter in $5$ of $408$ evaluations, with one run of each drifted at $B$. Without those two seeds, V1 is $-0.005$ and V2 is $+0.002$ (TOST $p=.020$).
\item Truncated samples at $B$: $20\%$ for each configuration. Candidates: $0.425$ on average, against $0.395$ at $10^{-6}$. Training time on H200 GPUs: $0.54$ hours per filter run and $0.47$ per outcome-only run.
\end{itemize}
Because V1 is neither significant nor equivalent to zero within $\pm0.01$, it remains unresolved whether this baseline reaches the filter at a constant rate; and because V2 is equivalent to zero, the schedule removes the filter's gain at $B$. The margin is wider than the gain at $10^{-6}$, however, and the V2 interval does not exclude a gain of that size.

\subsubsection{Pass@$k$ and test levels (exploratory)}
Each evaluation stores, per test problem, the number $c$ of correct samples out of $n=8$. \Cref{tab:passk} gives the unbiased estimate $1-\binom{n-c}{k}/\binom{n}{k}$ of pass@$k$, averaged over problems, as paired differences at $B$. Pass@8 equals one minus the share of problems that no sample solves. On Qwen3-0.6B ($54$ seeds), the filter's gain over outcome-only holds at every level of the test set. It is $+0.015$ $[0.008,0.022]$ on the $79$ level-3 problems, $+0.024$ $[0.016,0.032]$ on the $62$ level-4 problems, and $+0.011$ $[0.006,0.017]$ on the $59$ level-5 problems. With both configurations on the strict reward, the gain at $B$ again holds at pass@8, and by $2B$ it is gone at every $k$. On Qwen2.5-1.5B, the gain shrinks as $k$ grows, and pass@8 does not differ. On Qwen3-0.6B on levels 1--2, where the filter trains on harder prompts, its small pass@1 deficit becomes a small pass@8 surplus; neither is significant. On Llama-3.2-3B, where it also trains on harder prompts, the deficit instead grows with $k$, and at pass@8 its interval just excludes zero. On Llama-3.2-1B, as on Qwen2.5-1.5B, the small pass@1 lead shrinks at larger $k$.

\begin{table*}[t]\centering
\caption{\textbf{Pass@$k$ at the token budget} (exploratory): differences against outcome-only GRPO on the same seeds, with $95\%$ paired-$t$ CIs for pass@1 and pass@8; the last column is outcome-only GRPO's pass@8.}
\label{tab:passk}
\setlength{\tabcolsep}{2.5pt}\small
\begin{tabular}{llccccc}\toprule
Model & Configuration ($n$) & pass@1 & pass@2 & pass@4 & pass@8 & base \\\midrule
Qwen3-0.6B & filter ($54$) & $+0.017\,[+0.012,+0.021]$ & $+0.016$ & $+0.014$ & $+0.009\,[+0.002,+0.017]$ & $0.729$ \\
 & random ($18$) & $+0.008\,[-0.003,+0.019]$ & $+0.009$ & $+0.007$ & $+0.003\,[-0.010,+0.017]$ & \\
 & filter, default reward, $2B$ ($12$) & $+0.021\,[-0.001,+0.042]$ & $+0.019$ & $+0.015$ & $+0.010\,[-0.009,+0.028]$ & $0.747$ \\
 & filter, strict reward ($24$) & $+0.019\,[+0.013,+0.026]$ & $+0.018$ & $+0.015$ & $+0.013\,[+0.002,+0.024]$ & $0.734$ \\
 & same, $2B$ ($24$) & $+0.005\,[-0.005,+0.016]$ & $+0.006$ & $+0.004$ & $-0.002\,[-0.013,+0.009]$ & $0.765$ \\
Qwen2.5-1.5B & filter ($60$) & $+0.008\,[+0.004,+0.011]$ & $+0.006$ & $+0.006$ & $+0.005\,[-0.001,+0.012]$ & $0.687$ \\
 & random ($36$) & $-0.001\,[-0.005,+0.003]$ & $-0.001$ & $0.000$ & $+0.005\,[-0.003,+0.013]$ & \\
Qwen3-1.7B & filter ($24$) & $+0.002\,[-0.006,+0.011]$ & $+0.002$ & $+0.001$ & $+0.001\,[-0.006,+0.009]$ & $0.855$ \\
Qwen3-4B & filter ($12$) & $-0.002\,[-0.015,+0.010]$ & $-0.001$ & $+0.001$ & $+0.004\,[-0.009,+0.017]$ & $0.922$ \\
Qwen3-4B, L5 & filter ($12$) & $+0.002\,[-0.028,+0.032]$ & $+0.002$ & $+0.001$ & $+0.000\,[-0.017,+0.018]$ & $0.821$ \\
Qwen3-1.7B, L5 & filter ($24$) & $+0.011\,[+0.002,+0.020]$ & $+0.012$ & $+0.010$ & $+0.005\,[-0.005,+0.015]$ & $0.705$ \\
Qwen3-0.6B, L1--2 & filter ($24$) & $-0.005\,[-0.011,+0.000]$ & $+0.003$ & $+0.005$ & $+0.005\,[-0.003,+0.012]$ & $0.941$ \\
Llama-3.2-3B & filter ($24$) & $-0.004\,[-0.010,+0.002]$ & $-0.005$ & $-0.006$ & $-0.011\,[-0.023,-0.000]$ & $0.698$ \\
Llama-3.2-1B & filter ($24$) & $+0.004\,[-0.001,+0.010]$ & $+0.002$ & $+0.001$ & $+0.003\,[-0.011,+0.016]$ & $0.568$ \\
\bottomrule\end{tabular}
\end{table*}

\subsubsection{Tokens to reach outcome-only accuracy (exploratory)}
To express the head start in tokens, we take outcome-only GRPO's mean accuracy over seeds at $B$ as the target and interpolate the filter's mean curve over the same seeds (evaluations every $B/16$) to its first crossing of the target. \Cref{tab:tokens} gives that crossing as a fraction of the budget, with $95\%$ intervals from $10{,}000$ resamples of seeds that keep each seed's two configurations together. A resample whose curve never reaches the target counts as not reached. In the three settings of \Cref{tab:tokens} with a gain, the filter reaches outcome-only GRPO's accuracy at $B$ with $22$ to $23\%$ fewer generated tokens. The same crossing on the training-time axis, which counts the critic's prefill and refit, saves $13$ to $17\%$ of outcome-only GRPO's training time. At $2B$ under the strict reward, the saving on Qwen3-0.6B is $8\%$ and its interval includes no saving. The measure depends on the slope of outcome-only GRPO's curve near $B$, not only on the lead: on Llama-3.2-1B, whose outcome-only curve had nearly flattened, a lead that was not significant at $B$ gives the largest saving in the table, $41\%$ of the tokens.

\begin{table*}[t]\centering
\caption{\textbf{Share of the budget, and of outcome-only GRPO's training time, that the critic filter needs to reach outcome-only GRPO's accuracy at $B$} (exploratory; $1$ means no saving). The history-filter row uses the critic-free filter of S22 on the same seeds. Training time includes the critic's scoring and excludes evaluation.}
\label{tab:tokens}
\setlength{\tabcolsep}{3pt}\small
\begin{tabular}{lccc}\toprule
Setting ($n$) & target & tokens & training time \\\midrule
Qwen3-0.6B, L3--5 ($54$) & $0.446$ & $0.78\,[0.71,0.84]$ & $0.85\,[0.78,0.92]$ \\
\quad strict reward ($24$) & $0.448$ & $0.74\,[0.68,0.85]$ & -- \\
\quad strict reward, target at $2B$ ($24$) & $0.504$ & $0.92\,[0.77,\text{n.r.}]$ & -- \\
Qwen2.5-1.5B, L3--5 ($60$) & $0.421$ & $0.78\,[0.72,0.84]$ & $0.87\,[0.81,0.94]$ \\
\quad history filter ($36$) & $0.421$ & $0.89\,[0.81,0.95]$ & $0.88\,[0.80,0.94]$ \\
Qwen3-1.7B, level 5 ($24$) & $0.409$ & $0.77\,[0.70,0.93]$ & $0.83\,[0.76,1.01]$ \\
Qwen3-1.7B, L3--5 ($24$) & $0.667$ & $0.90\,[0.65,\text{n.r.}]$ & $0.93\,[0.67,\text{n.r.}]$ \\
Qwen3-4B, L3--5 ($12$) & $0.817$ & n.r.\ $[0.54,\text{n.r.}]$ & n.r.\ $[0.51,\text{n.r.}]$ \\
Qwen3-4B, level 5 ($12$) & $0.629$ & $0.79\,[0.53,\text{n.r.}]$ & $0.82\,[0.54,\text{n.r.}]$ \\
Qwen3-0.6B, L1--2 ($24$) & $0.809$ & n.r.\ $[\text{n.r.},\text{n.r.}]$ & n.r.\ $[\text{n.r.},\text{n.r.}]$ \\
Llama-3.2-3B, L3--5 ($24$) & $0.434$ & n.r.\ $[0.40,\text{n.r.}]$ & n.r.\ $[0.39,\text{n.r.}]$ \\
Llama-3.2-1B, L3--5 ($24$) & $0.261$ & $0.59\,[0.48,0.73]$ & $0.69\,[0.56,0.85]$ \\
\bottomrule\end{tabular}
\par\smallskip\footnotesize n.r.: not reached within the run.
\end{table*}

\subsection{Reward verification and drift}
\label{app:m-verifier}

When no answer is boxed, the default verifier falls back to the last number in a completion, so it also credits unboxed answers. Some runs trained on it drift away from boxing their answers; we count an evaluation as drifted when default-verifier accuracy exceeds strict-verifier accuracy by more than $0.1$. \Cref{tab:drift} counts drift by study and configuration. Drift appeared mainly in long or fast training: on Qwen3-4B with level 5, on Qwen3-0.6B over twice the budget or at twice the learning rate, and on Qwen2.5-1.5B in the replication and under both faster learning rates. It affected outcome-only GRPO more often than the filter, which may have inflated the filter's lead at $2B$ in S16. Elsewhere it was rare. The history filter of S22 drifted in $1$ of $612$ evaluations; on the Qwen2.5-1.5B level 4--5 pool, no Monte-Carlo, outcome-only, or baseline-critic run drifted (S25, S26), while the temporal-difference critic drifted at $5$ of $408$ evaluations, and $2$ of its $24$ runs ended drifted. On Qwen3-1.7B, drift appeared only under a length-reward term ($5$ of $3{,}717$ evaluations over all finished runs), and under that term two of three Qwen3-4B runs stopped boxing their answers, with strict-verifier accuracy of $0.165$ and $0.077$ while default-verifier accuracy stayed high. One filter run and one outcome-only run of the S11 Qwen3-4B replication also ended drifted.

Training on the strict verifier removes drift without lowering accuracy. In the studies that retrained drifting configurations on it (S17, S18), no run drifted ($0$ of $204$ evaluations on Qwen3-4B and $0$ of $1{,}584$ on Qwen3-0.6B), and neither did outcome-only GRPO at twice the learning rate (S29, $0$ of $408$). On the two Llama models, which trained on the strict reward throughout, $1$ of $1{,}632$ evaluations drifted. Default-verifier accuracy was non-inferior in both preregistered tests, and strict-verifier accuracy rose, mostly on the seeds whose default-reward runs had drifted. We therefore report strict-verifier accuracy alongside default-verifier accuracy and recommend the strict verifier as the training reward.

\subsubsection{Strict verifier as the training reward on Qwen3-4B with level 5 (S18)}
A first check, in S9, trained three Qwen3-4B outcome-only runs on levels 4--5 with the strict verifier as the reward: against the default-reward runs of the same seeds, accuracy changed by $+0.008$ $[-0.020,+0.036]$ (non-inferior at $-0.02$), and none of their $51$ evaluations drifted. S18 repeated the S14 outcome-only configuration with the strict verifier as the training reward, on the same twelve seeds, so that each run shares its prompt stream and evaluation draws with the S14 run of its seed. Held-out evaluation still used the default verifier, with strict-verifier accuracy logged. In every one of the $8{,}384$ training iterations, the logged training reward equals the strict reward. The preregistered primary contrast, default-verifier accuracy at $B$ under the strict reward minus under the default reward, was $+0.008$ $[-0.019,+0.036]$ (non-inferiority at $-0.02$, $p=.022$; crossed bootstrap $[-0.018,+0.035]$; higher in $6$ of $12$). No strict-reward run drifted ($0$ of $204$ evaluations, against $16$ of $204$ under the default reward). Strict-verifier accuracy at $B$ was $0.631$ against $0.520$ ($+0.111$ $[-0.032,+0.254]$, $p=.115$, secondary). The difference comes from the two seeds whose default-reward runs had drifted ($+0.627$ and $+0.546$); on the other ten seeds it lies between $-0.022$ and $+0.076$. Truncated samples at $B$ fell from $17\%$ to $13\%$, and training time per run was $2.17$ against $2.14$ hours.

\begin{table*}[t]\centering
\caption{\textbf{Drifted evaluations by configuration} (default minus strict-verifier accuracy $>0.1$), with the runs drifted at the last evaluation in parentheses.}
\label{tab:drift}
\setlength{\tabcolsep}{3pt}\small
\begin{tabular}{lllll}\toprule
Study & Model, pool, horizon & outcome-only & critic filter & random choice \\\midrule
S14 & Qwen3-4B, level 5, $B$ & $16/204$ ($2/12$) & $1/204$ ($0/12$) & -- \\
S18 & same, strict training reward & $0/204$ ($0/12$) & -- & -- \\
S16 & Qwen3-0.6B, L3--5, $2B$ & $33/396$ ($2/12$) & $4/396$ ($1/12$) & -- \\
S17 & same, strict training reward & $0/792$ ($0/24$) & $0/792$ ($0/24$) & -- \\
S23 & Qwen3-0.6B, L3--5, $B$, learning rate $2{\times}10^{-6}$ & $49/408$ ($5/24$) & $36/408$ ($3/24$) & -- \\
S34 & Qwen3-0.6B, L3--5, $B$, fresh seeds & $5/408$ ($1/24$) & $1/408$ ($1/24$) & -- \\
S13 & Qwen3-1.7B, L3--5, $B$ & $0/306$ & $0/306$ & $0/306$ \\
S19 & Qwen3-1.7B, level 5, $B$ & $0/408$ & $0/408$ & -- \\
S15 & Qwen2.5-1.5B, L3--5, $B$ & $0/204$ & $0/204$ & $0/204$ \\
S20 & same, fresh seeds & $16/408$ ($2/24$) & $1/408$ ($0/24$) & $0/408$ ($0/24$) \\
S33 & same, further fresh seeds & $0/408$ ($0/24$) & $8/408$ ($1/24$) & -- \\
S21 & Qwen3-0.6B, levels 1--2, $B$ & $1/408$ ($0/24$) & $2/408$ ($1/24$) & -- \\
S24 & Llama-3.2-3B, L3--5, $B$, strict training reward & $0/408$ & $0/408$ & -- \\
S27 & Llama-3.2-1B, L3--5, $B$, strict training reward & $0/408$ & $1/408$ ($0/24$) & -- \\
S28 & Qwen2.5-1.5B, L3--5, $B$, learning rate $2{\times}10^{-6}$ & $11/408$ ($1/24$) & $12/408$ ($1/24$) & -- \\
S29 & Qwen3-0.6B, L3--5, $B$, $2{\times}10^{-6}$, strict training reward & $0/408$ ($0/24$) & -- & -- \\
S30 & Qwen3-0.6B, L3--5, $B$, filter target $0.25$ & -- & $9/408$ ($1/24$) & -- \\
S31 & Qwen2.5-1.5B, L3--5, $B$, filter target $0.25$ & -- & $5/408$ ($1/24$) & -- \\
S32 & Qwen2.5-1.5B, L3--5, $B$, learning rate decaying from $2{\times}10^{-6}$ & $16/408$ ($1/24$) & $5/408$ ($1/24$) & -- \\
\bottomrule\end{tabular}
\end{table*}

\subsection{Summary}

Matching compute reverses the ranking that the per-update gradient test gives: the densest reward ranks first per update and last per token, and a learned critic, which needs no continuations, matches outcome-only GRPO within two points. The same protocol confirms a real gain where one exists. Choosing prompts whose predicted success is nearest $1/2$ improves accuracy on two models, replicates on fresh seeds and on test problems that no study had seen, comes from the choice of prompts, and follows a rule that predicted the outcome of five of six preregistered tests. The protocol also measures that gain precisely: it is a head start worth about a fifth of the budget, comparable to what a retuned learning rate buys. A claim that a dense reward or a prompt-selection rule helps should therefore state the budget it matches, the verifier that defines its reward, how its baseline was tuned, and the models on which it replicated.
\clearpage

\section{Training-Time Accounting}
\label{app:time}
Generated tokens are not the only cost of an intervention. An inference engine generates continuations more cheaply per token than a training step processes rollouts, and a critic adds forward passes and fits that generate no tokens at all. We therefore also measure training time, stage by stage, and compare configurations at equal training time.

Every run logs, for each iteration, the wall time of prompt selection (including the prompt critic's prefill pass), generation of rollouts, credit (including Monte-Carlo continuations and the learned critic's features and fits), the optimizer update, and the copy of the updated weights into the generation engine. Training time is the sum of these over the iterations before an evaluation; evaluation, logging, and checkpointing are timed separately and excluded. Each run uses one GPU (an H200, for every run in this section) on nodes shared with other jobs. We compare configurations only within a study or across studies run on the same GPU type, paired by seed.

\begin{figure}[t]
\centering
\includegraphics[width=\columnwidth]{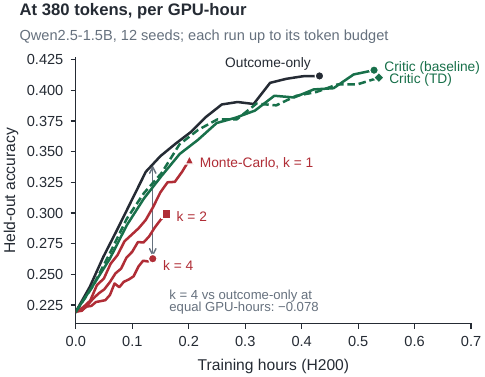}
\caption{\textbf{At $380$ tokens, the Monte-Carlo reward trails outcome-only GRPO at its own training time.} Exploratory reanalysis of preregistered runs on H200 GPUs: accuracy against training hours for Qwen2.5-1.5B on MATH L4--5 (seeds 5--16, the same for every configuration), each run up to its token budget. The Monte-Carlo rewards reach the budget sooner because their continuations take less time per token than training does; the arrow marks the $k=4$ reward's deficit at its own training time (per-seed mean). \Cref{fig:time}a and \Cref{app:p39} run every configuration past outcome-only GRPO's training time, at $768$ tokens.}
\label{fig:time-h200}
\end{figure}

\textbf{Equal-time contrasts} (exploratory reanalyses of preregistered runs). For each seed, $T^*$ is the shorter of the two runs' training times up to their evaluation at $B$. Each run's accuracy is interpolated linearly in training time at $T^*$, so the configuration that reaches $B$ first is read at $B$, and the other is read between two of its evaluations. These contrasts stop when the faster configuration reaches its budget. A Monte-Carlo reward's runs end before outcome-only GRPO has trained as long as it needs to reach $B$, so \Cref{app:p39} gives every configuration that much time in a preregistered study. \Cref{tab:time} lists every contrast, \Cref{tab:stages} gives the time per stage, and \Cref{fig:time-h200} shows the training curves of the Monte-Carlo rewards and critics.

\begin{table*}[t]
\centering
\caption{\textbf{Matched generated tokens versus matched training time, on the same runs.} Configuration minus base in held-out accuracy at the first evaluation at or after $B$ (equal tokens) and at equal training time $T^*$ (exploratory), with $95\%$ paired-$t$ CIs over seeds; training hours before the evaluation at $B$ on H200 GPUs. Qwen2.5-1.5B on the level 4--5 pool at $380$ tokens unless marked; prompt-selection rows on the level 3--5 pool.}
\label{tab:time}
\small
\setlength{\tabcolsep}{4pt}
\resizebox{\textwidth}{!}{%
\begin{tabular}{@{}llcccc@{}}
\toprule
Configuration & Base & $n$ & Equal tokens & Equal training time & Hours to $B$ (configuration / base) \\
\midrule
Monte-Carlo, $k=4$ & outcome-only & $12$ & $-0.149$ & $-0.078$ $[-0.088,-0.068]$ & $0.136$ / $0.431$ \\
Monte-Carlo, $k=2$ & outcome-only & $12$ & $-0.112$ & $-0.051$ $[-0.062,-0.040]$ & $0.160$ / $0.431$ \\
Monte-Carlo, $k=1$ & outcome-only & $12$ & $-0.068$ & $-0.022$ $[-0.028,-0.017]$ & $0.201$ / $0.431$ \\
Critic, TD form & outcome-only & $24$ & $-0.004$ & $-0.012$ $[-0.017,-0.007]$ & $0.536$ / $0.431$ \\
Critic, baseline form & outcome-only & $24$ & $+0.007$ & $-0.007$ $[-0.012,-0.001]$ & $0.531$ / $0.431$ \\
Monte-Carlo, $k=4$, $768$ tokens & outcome-only & $3$ & $-0.061$ & $-0.016$ $[-0.045,+0.013]$ & $0.202$ / $0.556$ \\
Critic, TD form, $768$ tokens & outcome-only & $3$ & $-0.011$ & $-0.018$ $[-0.050,+0.015]$ & $0.786$ / $0.556$ \\
Critic, baseline form, $768$ tokens & outcome-only & $3$ & $+0.008$ & $+0.006$ $[-0.021,+0.033]$ & $0.653$ / $0.556$ \\
Monte-Carlo, $k=4$, $4$ steps per batch & outcome-only & $3$ & $-0.071$ & $-0.019$ $[-0.079,+0.042]$ & $0.165$ / $0.424$ \\
Outcome-only, $4$ steps per batch & outcome-only & $3$ & $-0.059$ & $-0.032$ $[-0.082,+0.018]$ & $0.819$ / $0.424$ \\
VinePPO-style & outcome-only & $3$ & $-0.150$ & $-0.080$ $[-0.104,-0.057]$ & $0.136$ / $0.424$ \\
VinePPO-style, $4$ steps per batch & outcome-only & $3$ & $-0.070$ & $-0.023$ $[-0.048,+0.003]$ & $0.172$ / $0.424$ \\
Outcome-only, lr $2\times10^{-6}$ & outcome-only & $3$ & $-0.034$ & $-0.036$ $[-0.095,+0.022]$ & $0.486$ / $0.424$ \\
Outcome-only, lr $3\times10^{-6}$ & outcome-only & $3$ & $-0.111$ & $-0.096$ $[-0.184,-0.009]$ & $0.477$ / $0.424$ \\
\midrule
Critic filter & outcome-only & $60$ & $+0.008$ & $+0.005$ $[+0.002,+0.008]$ & $0.516$ / $0.451$ \\
\quad seeds of the third study & outcome-only & $24$ & $+0.007$ & $+0.006$ $[+0.001,+0.010]$ & $0.517$ / $0.450$ \\
History prompt filter & outcome-only & $36$ & $+0.006$ & $+0.007$ $[+0.002,+0.011]$ & $0.452$ / $0.452$ \\
Random choice of candidates & outcome-only & $36$ & $-0.001$ & $-0.002$ $[-0.006,+0.002]$ & $0.496$ / $0.452$ \\
Outcome-only, decaying lr & filter at $10^{-6}$ & $24$ & $-0.005$ & $-0.003$ $[-0.010,+0.004]$ & $0.474$ / $0.514$ \\
Filter, decaying lr & outcome-only, decaying lr & $24$ & $+0.002$ & $+0.002$ $[-0.005,+0.009]$ & $0.544$ / $0.474$ \\
Critic filter, Qwen3-0.6B & outcome-only & $54$ & $+0.017$ & $+0.010$ $[+0.006,+0.015]$ & $0.570$ / $0.522$ \\
\quad seeds of the fourth study & outcome-only & $24$ & $+0.012$ & $+0.004$ $[-0.001,+0.010]$ & $0.568$ / $0.525$ \\
\bottomrule
\end{tabular}}
\end{table*}

\begin{table}[t]
\centering
\caption{\textbf{Where the training time goes.} Mean seconds per run in each stage before the evaluation at $B$, on H200 GPUs; Qwen2.5-1.5B, seeds of \Cref{tab:time}.}
\label{tab:stages}
\footnotesize
\setlength{\tabcolsep}{3pt}
\begin{tabular}{@{}lrrrrr@{}}
\toprule
Configuration & Select & Generate & Credit & Update & Total \\
\midrule
Outcome-only (L4--5) & $0$ & $604$ & $0$ & $940$ & $1{,}552$ \\
Monte-Carlo, $k=4$ & $0$ & $53$ & $357$ & $80$ & $491$ \\
Monte-Carlo, $k=2$ & $0$ & $97$ & $331$ & $147$ & $576$ \\
Monte-Carlo, $k=1$ & $0$ & $168$ & $292$ & $261$ & $723$ \\
Critic, TD form & $0$ & $607$ & $366$ & $948$ & $1{,}929$ \\
Critic, baseline form & $0$ & $601$ & $362$ & $939$ & $1{,}910$ \\
Outcome-only (L3--5) & $0$ & $627$ & $0$ & $988$ & $1{,}624$ \\
Critic filter & $186$ & $642$ & $0$ & $1{,}019$ & $1{,}857$ \\
History prompt filter & $1$ & $627$ & $0$ & $990$ & $1{,}627$ \\
\bottomrule
\end{tabular}
\end{table}

Totals include the weight sync ($1$ to $9$ seconds per run). The Monte-Carlo rewards spend most of their time on continuations (the credit stage), which the generation engine samples in large batches, whereas outcome-only GRPO spends most of its time on its many updates. A continuation token therefore costs less time than a rollout token that is also trained on, which is why the token budget charges the Monte-Carlo rewards more heavily than the time budget does. The learned critic's features and fits add about a quarter to outcome-only GRPO's training time, and the prompt critic's prefill pass adds about a tenth.

\section{Pilot Studies of Sample Reuse and Other Estimators (S3, S4)}
\label{app:pilots}

Two preregistered pilot studies tested ways to make dense rewards cheaper, as well as alternative estimators. They used Qwen2.5-1.5B with the protocol of S1 (MATH L4--5 pool, $380$ tokens, $B=12.25$M, learning rate $10^{-6}$) and seeds 0--2 on H200 GPUs, and compared against the outcome-only and $k=4$ Monte-Carlo runs of S2 on the same seeds. With three seeds, these are pilots: their intervals are wide, and only large effects are resolved.

\textbf{Continuation reuse at the gradient level.} Training on the continuations themselves uses every generated token for training. Each continuation becomes a training sequence, with the prompt and the rollout prefix as context, and receives leave-one-out credit among the draws at its cut, in the style of SPO-tree and TreeRL. On $64$ fixed batches, measured with the gradient efficiency of \Cref{sec:q2}, this was $6.4\times$ $[3.3,18.8]$ as efficient as outcome-only GRPO per update with $k=4$ continuations per cut and $3.7\times$ $[2.1,10.1]$ with $k=2$, but $0.60\times$ $[0.31,1.75]$ and $0.64\times$ $[0.35,1.73]$ per generated token, against $0.22\times$ $[0.15,0.43]$ for the Monte-Carlo reward without reuse on the same batches. The preregistered condition for training it, a per-token efficiency of at least $0.8$, was not met.

\textbf{Rollout reuse and the VinePPO-style advantage.} Accuracy at $B$: outcome-only GRPO $0.408$, the $k=4$ Monte-Carlo reward $0.260$. Four clipped optimizer steps per batch (two epochs of two minibatches, clip $0.2$) gave $0.337$ for the reward ($+0.077$ $[0.049,0.105]$, Holm $p=.007$) and $0.349$ for outcome-only GRPO ($-0.059$ $[-0.102,-0.016]$, $p=.027$); the preregistered interaction, the change in outcome-only GRPO's lead, was $-0.136$ $[-0.175,-0.097]$ (Holm $p=.009$). The VinePPO-style value advantage (Monte-Carlo value differences alone, with $V(\text{prompt})$ the leave-one-out group mean) gave $0.258$, equivalent to the reward ($-0.002$ $[-0.024,+0.019]$, TOST $p=.036$), and $0.337$ with the same reuse ($+0.000$ $[-0.042,+0.043]$ against the reused reward). REINFORCE with a leave-one-out baseline instead of the group-normalized advantage changed neither configuration materially (outcome-only $0.418$, Monte-Carlo reward $0.259$).

\textbf{Process rewards from a frozen PRM.} Process-supervised GRPO (adapted from DeepSeekMath) with the frozen Qwen2.5-Math-PRM-7B and without the verifier reward collapsed: the policy stopped boxing its answers ($0.005$ boxes per completion), and accuracy fell to $0.043$ ($-0.365$ $[-0.389,-0.341]$). Adding the outcome reward to the process rewards did not prevent the collapse ($0.040$). Used instead as a baseline for the outcome reward that depends only on the state, the same PRM was equivalent to outcome-only GRPO ($-0.002$ $[-0.023,+0.018]$, TOST $p=.033$).

\textbf{Learning rate and group size.} Outcome-only GRPO at $2\times10^{-6}$ and $3\times10^{-6}$ ended $0.034$ $[0.004,0.064]$ and $0.111$ $[-0.003,+0.224]$ below its run at $10^{-6}$ on this pool. Neither critic form gained over outcome-only GRPO at the same higher rate, with four optimizer steps per batch, or with groups of $2$ or $4$ completions (every contrast within $[-0.019,+0.021]$, none significant). \Cref{tab:time} gives these pilots on the training-time axis.

\section{Equal Training Time Past the Token Budget (S35)}
\label{app:p39}

The equal-time contrasts of \Cref{app:time} stop at a Monte-Carlo reward's own training time at $B$, which is shorter than outcome-only GRPO's. This study asks whether a reward catches up when given as much training time as outcome-only GRPO. A preregistered study (fixed before any of its training runs) answers this question with $768$-token completions, where $20\%$ of the initial samples are truncated instead of $73\%$. It trains Qwen2.5-1.5B on the level 4--5 pool, with the strict verifier as every configuration's training reward, learning rate $10^{-6}$, $8$ prompts $\times$ $8$ completions per iteration, $B=17.0$M generated tokens, an evaluation every $B/16$, seeds $200$--$207$, and one A100 GPU per run (time axes are compared only within the study). Every configuration trains past $B$, with budgets set from the time ratios of \Cref{tab:time} so that its training time passes $T_c$, the training time that outcome-only GRPO needs to reach $B$ on the same seed. Outcome-only GRPO trains to $1.25B$, the $k=4$ Monte-Carlo reward to $4B$, the $k=1$ reward to $3B$, continuation reuse with $k=2$ (every continuation is also trained on, as in \Cref{app:pilots}) to $2B$, and the VinePPO-style value advantage with four clipped optimizer steps per batch to $3B$. The primary contrasts T1--T4 compare each configuration with outcome-only GRPO at $T_c$, with the configuration's accuracy interpolated linearly in training time (paired $t$ over seeds, two-sided, Holm over the four). They were predicted to be negative for both Monte-Carlo rewards, with no predicted direction for the two variants with reuse. The same contrasts at equal generated tokens, and at the shorter of the two runs' training times at $B$, are secondary.

\begin{figure*}[t]
\centering
\includegraphics[width=0.92\textwidth]{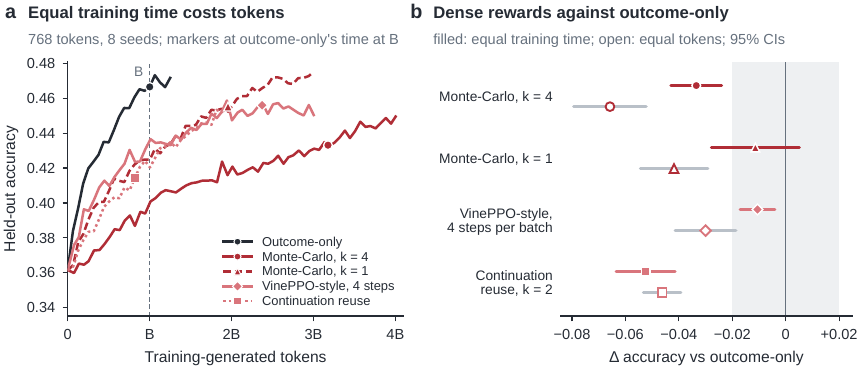}
\vspace{-6pt}
\caption{\textbf{Equal training time past the token budget (S35).} Qwen2.5-1.5B on MATH L4--5 at $768$ tokens, strict training reward, $8$ seeds on A100 GPUs. \textbf{(a)} Mean accuracy against tokens generated in training, over each configuration's whole run; markers show where each configuration's training time reaches $T_c$, the time outcome-only GRPO needs to reach $B$. Matching that time gives the $k=4$ reward $3.2B$ tokens and continuation reuse $0.82B$. \textbf{(b)} Each configuration minus outcome-only GRPO at $T_c$ (filled; preregistered primary) and at equal generated tokens (open; secondary), with $95\%$ paired-$t$ CIs over seeds; the gray band marks $\pm0.02$.}
\label{fig:p39}
\end{figure*}

\begin{table*}[t]
\centering
\caption{\textbf{Equal training time past the token budget (S35).} Qwen2.5-1.5B on MATH L4--5 at $768$ tokens, strict training reward, $8$ seeds, A100 GPUs. Configuration minus outcome-only GRPO in held-out accuracy: at $T_c$, outcome-only GRPO's training time at $B$ (preregistered primary, Holm over the four configurations; TOST at $\pm0.02$; seeds in which the configuration was lower), at equal generated tokens $B$, and at the earlier of the two runs' training times at $B$ (both secondary), with $95\%$ paired-$t$ CIs over seeds. Continuation reuse reaches $B$ after outcome-only GRPO, so its earlier time is $T_c$. Last column: mean training hours before the evaluation at $B$.}
\label{tab:p39}
\small
\setlength{\tabcolsep}{4pt}
\resizebox{\textwidth}{!}{%
\begin{tabular}{@{}lccccccc@{}}
\toprule
& \multicolumn{4}{c}{Equal training time $T_c$ (primary)} & Equal tokens & Earlier time at $B$ & Hours \\
\cmidrule(lr){2-5}
Configuration & $\Delta$ $[95\%$ CI$]$ & Holm $p$ & TOST $p$ & Lower & $\Delta$ $[95\%$ CI$]$ & $\Delta$ $[95\%$ CI$]$ & to $B$ \\
\midrule
Outcome-only GRPO & -- & & & & -- & -- & $1.29$ \\
Monte-Carlo, $k=4$ & $-0.033$ $[-0.043,-0.024]$ & $2\times10^{-4}$ & $.99$ & $8/8$ & $-0.066$ $[-0.079,-0.052]$ & $-0.024$ $[-0.034,-0.014]$ & $0.40$ \\
Monte-Carlo, $k=1$ & $-0.011$ $[-0.028,+0.005]$ & $.15$ & $.13$ & $7/8$ & $-0.042$ $[-0.054,-0.029]$ & $-0.011$ $[-0.021,-0.001]$ & $0.63$ \\
VinePPO-style, $4$ steps per batch & $-0.011$ $[-0.017,-0.004]$ & $.013$ & $.005$ & $7/8$ & $-0.030$ $[-0.041,-0.019]$ & $+0.003$ $[-0.008,+0.015]$ & $0.52$ \\
Continuation reuse, $k=2$ & $-0.052$ $[-0.063,-0.041]$ & $<10^{-4}$ & $1.00$ & $8/8$ & $-0.046$ $[-0.053,-0.039]$ & $-0.052$ $[-0.063,-0.041]$ & $1.58$ \\
\bottomrule
\end{tabular}}
\end{table*}

\textbf{Primary.} Every configuration's training time passed $T_c$ on every seed, by a factor of at least $1.23$, so no budget had to be extended. $T_c$ was $1.29$ hours ($1.27$--$1.31$ across seeds), at which outcome-only GRPO's accuracy was $0.467$, up from $0.361$ at the start. The $k=4$ reward trailed by $0.033$ $[0.024,0.043]$ (T1, Holm $p=2\times10^{-4}$, lower in all $8$ seeds; crossed seed-by-problem interval $[-0.048,-0.019]$), as predicted. The $k=1$ reward trailed by $0.011$ $[-0.005,+0.028]$ (T2, Holm $p=.15$, lower in $7$ seeds), a difference neither shown below zero nor within $\pm0.02$ (TOST $p=.13$). Its predicted deficit was therefore not confirmed, so our claim about the $k=1$ reward in training time is limited to its own training time. Continuation reuse trailed by $0.052$ $[0.041,0.063]$ (T3, Holm $p<10^{-4}$, lower in all $8$; crossed $[-0.069,-0.036]$), and the VinePPO-style advantage with four steps per batch by $0.011$ $[0.004,0.017]$ (T4, Holm $p=.013$, lower in $7$), a deficit within $\pm0.02$ (TOST $p=.005$) whose crossed interval reaches zero ($[-0.025,+0.003]$). This configuration therefore matches outcome-only GRPO per GPU-hour within $\pm0.02$, while trailing it.

\textbf{Secondary.} At equal generated tokens, all four configurations trailed, by $0.066$ ($k=4$), $0.042$ ($k=1$), $0.046$ (continuation reuse), and $0.030$ (VinePPO-style), each with $p<.001$ and lower in all $8$ seeds. At the shorter of the two runs' training times at $B$, which for every configuration except continuation reuse is the configuration's own, the $k=4$ and $k=1$ rewards trailed by $0.024$ $[0.014,0.034]$ ($p<.001$) and $0.011$ $[0.001,0.021]$ ($p=.038$), and the VinePPO-style variant was even with outcome-only GRPO, at $+0.003$ $[-0.008,+0.015]$ (TOST $p=.006$). Compared only when the faster configuration reaches its budget, as in \Cref{tab:time}, the VinePPO-style variant would have tied outcome-only GRPO; it rose fastest at first and then levelled off near $0.45$ (\Cref{fig:time}a, \Cref{fig:p39}a).

\textbf{Descriptive.} Training hours to $B$ were $0.40$ for the $k=4$ reward, $0.52$ for the VinePPO-style variant, $0.63$ for the $k=1$ reward, and $1.58$ for continuation reuse, against $1.29$ for outcome-only GRPO, so in outcome-only GRPO's training time they could generate $3.2B$, $2.4B$, $2.0B$, and $0.82B$ tokens. Before $B$, outcome-only GRPO spent $60\%$ of its training time on updates and $39\%$ on generation; the $k=4$ reward spent $66\%$ on its credit stage, which includes generating its continuations; and continuation reuse spent $78\%$ on updates, which include training on the continuations ($45$ seconds per iteration, against outcome-only GRPO's $6$). On the mean curves, the $k=1$ reward reached outcome-only GRPO's accuracy at $B$ after $1.25\,T_c$; the other three did not within their runs, which ended at $1.27\,T_c$ ($k=4$), $1.29\,T_c$ (VinePPO-style), and $2.5\,T_c$ (continuation reuse). The share of truncated samples fell from $20\%$ at the start to $9\%$ for outcome-only GRPO at $B$ and to between $9\%$ and $16\%$ for the other configurations at $T_c$. With the strict reward, no evaluation drifted (default minus strict-verifier accuracy above $0.1$), and strict-verifier accuracy at $B$ was $0.004$ to $0.012$ below default-verifier accuracy. Peak reserved GPU memory was $68$--$73$ GiB per run; the study took $105$ A100 GPU-hours, including a smoke test of every configuration.

\textbf{Adaptations, not reproductions.} The two configurations with reuse adapt published methods to our trainer and recipe. \Cref{tab:adapt} lists what each keeps and what it changes; the differences in branch points, samples per branch point, and update rule change both the variance of the credit and its cost. These results therefore describe the adaptations, not the published methods in their own settings.

\begin{table*}[t]
\centering
\caption{\textbf{What our dense-reward variants keep from the methods they adapt.} Published settings are the main ones of each paper \citep{kazemnejad2024vineppo,guo2025spo,hou2025treerl}. Our configurations share one trainer and recipe (MATH L4--5, $8$ prompts $\times$ $8$ completions per iteration, Adam at a constant $10^{-6}$, the strict reward in S35). The Monte-Carlo rewards ($k=1$, $2$, $4$) are our construction: segment credit from the same kind of continuations, with a fixed $V(\text{prompt})=0.5$, added to the group advantage, with one on-policy step. A shared prefix is the part of a completion before a branch point.}
\label{tab:adapt}
\footnotesize
\renewcommand{\arraystretch}{1.15}
\begin{tabularx}{\textwidth}{@{}>{\raggedright\arraybackslash}p{1.75cm}>{\raggedright\arraybackslash}X>{\raggedright\arraybackslash}X>{\raggedright\arraybackslash}X>{\raggedright\arraybackslash}X@{}}
\toprule
& VinePPO & Our VinePPO-style variant & SPO-tree; TreeRL & Our continuation reuse \\
\midrule
Credit & $r+V(s_{t+1})-V(s_t)$ for each reasoning step; advantages normalized over the batch & $V$ at a segment's end minus at its start, with $V(\text{prompt})$ the leave-one-out group mean and $V(\text{end})=R$; not normalized & SPO-tree: a node's value minus its siblings' mean, on tokens with probability below $0.9$. TreeRL: $[V(n)-V(\text{root})]+[V(n)-V(\text{parent})]$, over $\sqrt{\text{leaves below } n}$ & Rollouts: group advantage plus segment credit (the $k=2$ reward). Continuations: their outcome minus the other draw's at their cut \\
Branch points & Every reasoning step (newline and punctuation boundaries) & $5$ fixed offsets ($128$, $256$, \dots, $640$ tokens) & SPO-tree: fixed token intervals, $3$ levels. TreeRL: the $2$ most uncertain tokens in each of $6$ trees & The same $5$ offsets; one level \\
Samples per point & $9$ rollouts & $4$ continuations & SPO-tree: $6$ children per node ($216$ paths). TreeRL: $2$ branches per fork ($30$ responses) & $2$ continuations \\
Trains on continuations & No & No & SPO-tree: segments with nonzero credit. TreeRL: every root-to-leaf path & Every continuation, with the prompt and prefix as context \\
Shared prefixes & -- & -- & SPO-tree: one training segment per node. TreeRL: in every path through them, downweighted & Trained once, in the rollout \\
Update & Clipped ($0.2$), $2$ epochs, minibatch $64$, $512$ episodes per step; KL penalty $10^{-4}$ & Clipped ($0.2$), $2$ epochs $\times$ $2$ minibatches of $32$, $64$ completions per iteration; no KL penalty & SPO-tree: clipped, $1$ epoch, KL $10^{-4}$. TreeRL: policy gradient, one step per $16$ prompts, KL $10^{-4}$ & One on-policy step over rollouts and continuations, the loss divided by their number; no clipping or KL penalty \\
Models and length & RhoMath 1.1B and DeepSeekMath 7B fine-tuned on the task; $1{,}024$ tokens; temperature $0.6$ & Qwen2.5-1.5B-Instruct; $768$ tokens; temperature $1$ & SPO-tree: RhoMath 1.1B; R1-Distill-Qwen-1.5B at $2$K--$4$K tokens. TreeRL: $7$B--$14$B models; $8{,}192$ tokens & Qwen2.5-1.5B-Instruct; $768$ tokens; temperature $1$ \\
\bottomrule
\end{tabularx}
\end{table*}

\section{Equal-Iteration Study: Full Results}
\label{app:equalround}

This section gives the complete account of the equal-iteration study that \Cref{sec:equalround} summarizes.

\begin{figure*}[t]
\centering
\includegraphics[width=\textwidth]{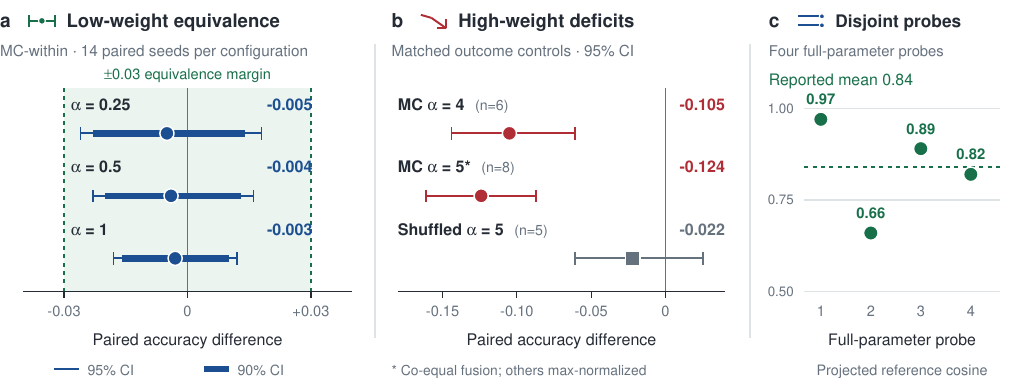}
\vspace{-14pt}
\caption{\textbf{Strong local agreement can coexist with weak training results.} Qwen2.5-1.5B, Monte-Carlo reward, equal iterations. \textbf{(a)} The three low-weight configurations satisfy the $\pm0.03$ equivalence criterion ($14$ paired seeds; thick and thin bars: $90\%$ and $95\%$ seed-bootstrap intervals). \textbf{(b)} High-weight configurations against their own matched outcome controls ($95\%$ intervals); the $\alpha=4$ deficit survives only the bootstrap Holm correction, and the starred configuration uses co-equal standardization, the others the max-normalized rule. \textbf{(c)} Four full-parameter probes with disjoint rollouts (mean projected reference cosine $0.84$), measured separately from the training runs.}
\label{fig:gap}
\end{figure*}

\Cref{fig:pipeline} traces the dense reward from the gradient probe to the training result. This experiment tests whether favorable local measurements pick out a useful dense reward in a short GRPO recipe. We fully fine-tune \mbox{Qwen2.5-1.5B-Instruct} \citep{qwen2.5,qwen2} on MATH problems of levels 4--5 \citep{hendrycks2021math} for $32$ iterations, with $8$ completions for each of $8$ prompts per iteration. Separate probes, with LoRA and with full-parameter gradients, measure the gradient geometry on fixed batches. Training samples at temperature $1.0$ with top-$p$ $0.95$ and uses Adam with gradient clipping, whereas the reported accuracy uses greedy decoding. Greedy accuracy is therefore a different target from the expected accuracy $J$ under sampling used in the local analysis. The probes also compare against a reference gradient of progress estimated from a finite number of Monte-Carlo continuations, denoted $\widehat h_{\mathrm{MC}}$, rather than against the exact population gradient (\Cref{app:lmrecipe,app:probeimpl}).

\textbf{Terms.} The Monte-Carlo reward assigns each of six segments of a completion the change in success rate across it, as estimated from $k$ continuations. In this study, the credit is centered within each completion (MC-within in the tables), normalized, and combined with the outcome advantage as $A_g+\alpha w_{g,h}$. There are two normalizations: dividing by the credit's largest absolute value in the batch (max-normalized, the default), or dividing by its standard deviation (co-equal standardization). The probes split the continuation samples into two halves: half A defines the progress reference $\widehat h_{\mathrm{MC}}$, and half B gives a candidate credit from disjoint samples. Gates act on the credit before centering. The sign-opposition gate keeps only the entries whose sign is opposite to $A_g$; the rollout-disagreement gate weights each segment by $p(1-p)$, where $p$ is the fraction of continuations that succeed from the segment's end. The implicit reward is $\log\pi_\theta-\log\pi_{\mathrm{ref}}$, the log ratio between the current policy and the reference policy.

\textbf{What the paired comparisons measure.} For each seed, we compare the change in accuracy from start to end of the dense-reward configuration with that of the outcome-only configuration, using the same validation prompts and scoring procedure. Seed-bootstrap intervals describe uncertainty across seeds on these fixed prompts. Probes and training configurations are separate measurements, so \Cref{fig:gap} sets them side by side rather than run by run. Matching seeds and training iterations also leaves a difference in compute, because the Monte-Carlo reward needs extra continuations from each prefix. \Cref{sec:q1} repeats the comparison at matched generated tokens; measuring the resources used by each stage is a separate validation step (\Cref{app:lmrecipe,app:stats}).

\subsection{Does agreement with the reference survive disjoint sampling?} 
\begin{figure*}[t]
\centering
\includegraphics[width=\textwidth]{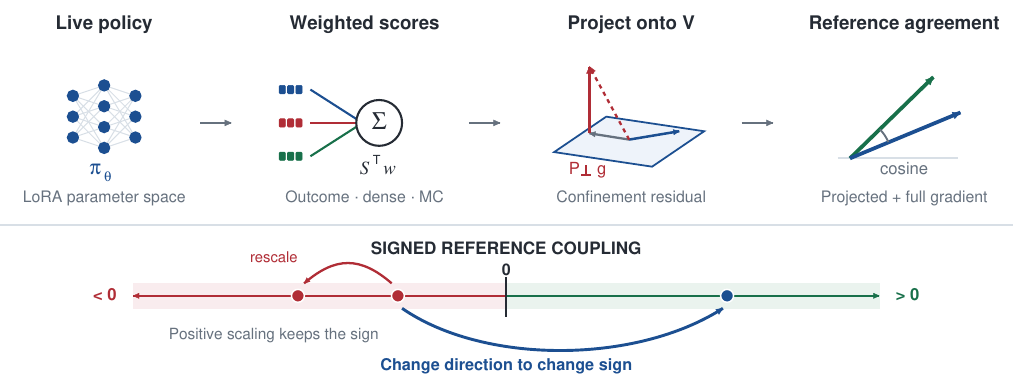}
\caption{\textbf{Gradient-reference measurements at a fixed policy and batch.} Weighted backward passes compute completion-score sums $\{q_g\}$, outcome gradient $\gout$, dense gradients $\grad(w)$, and Monte-Carlo reference $\widehat h_{\mathrm{MC}}$ in LoRA parameter space. Projecting onto $V=\Span\{q_g\}$ gives the confinement residual, projected alignment, and full-gradient reference cosine. The lower axis illustrates $\langle\Pperp\widehat h_{\mathrm{MC}},\Pperp\grad(w)\rangle$: positive rescaling preserves its sign, while a direction change can change it. This inner product measures agreement with an estimated reference. The population directional derivative and actual Adam update are separate quantities.}
\label{fig:probe}
\end{figure*}

The probe computes the summed scores of each completion, the outcome gradient, and the dense-reward gradient, and then projects the gradients onto $V$ (\Cref{fig:probe}). The outcome gradient's residual outside $V$ is $2{\times}10^{-13}$ in full-parameter space, a numerical check of the fixed-batch identity. For the MC-within reward, the strongest check uses four full-parameter probes with disjoint rollouts. Their alignments with an independently sampled half-A reference are $0.97,0.66,0.89,0.82$, with mean $0.84$. The larger values we also report, $0.95$ in pooled LoRA probes and $0.96$ in full-parameter probes, compare a credit computed from all rollouts with a reference that partly overlaps with it. Disjoint sampling removes the direct reuse of rollouts, but the reference still depends on the estimator's finite horizon and segmentation.

\begin{figure*}[t]\centering
\includegraphics[width=0.78\textwidth]{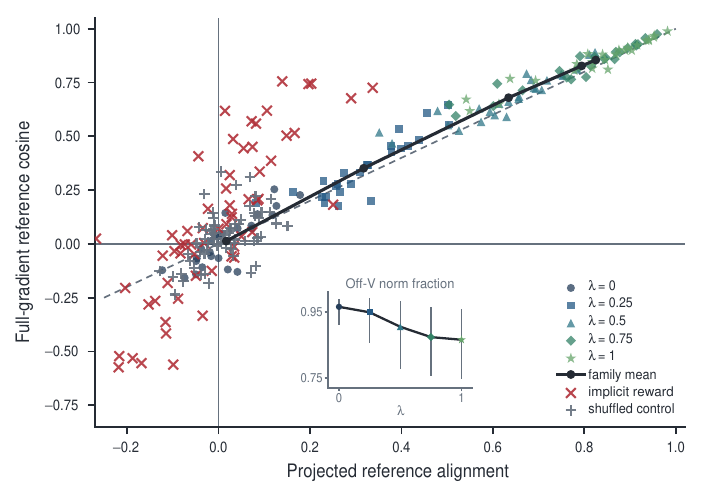}
\caption{\textbf{Projected and full-gradient reference agreement rise together.} Mixing weight $\lambda$ moves the credit from noise toward half-B Monte-Carlo reward. Both measurements use the disjoint half-A reference $\widehat h_{\mathrm{MC}}$. The horizontal axis is $\cos(\Pperp g,\Pperp\widehat h_{\mathrm{MC}})$; the vertical axis is $\cos(g,\widehat h_{\mathrm{MC}})$. Their means rise from $0.02$ to $0.83$ and from $0.01$ to $0.85$, respectively. The inset shows the mean and range of the fraction of the update's norm outside $V$, which falls from $0.96$ to $0.87$. Each of five weights has $25$ probe rows from five runs and five drifted iterations; rows from the same run are correlated. Self-log-ratio ($\times$) and shuffled ($+$) controls are also shown; two of $63$ implicit-reward rows lie outside the window at alignments $-0.33$ and $-0.30$. Both axes are descriptive gradient measurements against the same reference.}
\label{fig:law}\end{figure*}

\textbf{Does the credit add a useful direction?} The Monte-Carlo reward, the implicit reward, and a permutation of the Monte-Carlo reward within each completion all produce updates that lie largely outside $V$: the part outside $V$ carries $0.86$--$0.95$ of the update's norm (\Cref{tab:alignment}). Only the Monte-Carlo reward consistently aligns with the estimated reference. The pooled alignments of the implicit-reward and shuffled rewards are $-0.01\pm0.08$ and $-0.01\pm0.06$. Shuffling keeps the magnitudes of the token credits unchanged, but it can change the norm of the projected gradient and the Adam step. Interpolating between an independent half-B credit and noise also links the alignment of the projected components to the cosine between the full gradient and the reference (\Cref{fig:law}). Similar descriptive patterns appear across \nmodels{} checkpoints from five model families (\Cref{tab:multimodel}). These probes measure raw credits before the optimizer acts on them. The next question is whether this well-aligned reward improves training.

\subsection{Does the gradient test pick out a better training reward?} The outcome-only baseline improves greedy validation accuracy by $+0.056$ $[+0.041,+0.071]$ (\Cref{tab:fullft}). At $\alpha=0.25,0.5,1$, a prespecified expansion to $n=14$ paired seeds supports equivalence with the baseline within $\pm0.03$: the $90\%$ intervals lie inside this margin. The result holds with Holm correction over the three configurations, under percentile-bootstrap, paired-$t$ two one-sided (TOST), and studentized-bootstrap tests. Smaller benefits and harms remain unresolved. At $\alpha=5$, the paired deficit is $-0.124$ $[-0.161,-0.087]$ ($n=8$, $d_z=-2.15$), which survives Holm correction over the nine Monte-Carlo weight configurations of the primary harm analysis (\Cref{app:stats}) under both bootstrap and paired-$t$ tests. The deficit at $\alpha=4$ survives only the bootstrap test. In this short training recipe, therefore, strong agreement with the reference coexists with equivalence at low weights and harm at high weights.

\begin{table*}[t]\centering
\caption{\textbf{Low-weight equivalence and high-weight harm in the primary full-fine-tuning recipe.} Qwen2.5-1.5B trained on MATH L4--5. Gain is final$-$start greedy validation accuracy; ``vs.\ outcome'' is the paired same-seed difference, with $95\%$ seed-bootstrap CIs and effect size $d_z$. Only the three $n{=}14$ low-weight configurations establish equivalence within $\pm0.03$, under bootstrap TOST ($p{=}.0095,.0033,.0004$), paired-$t$ TOST, and studentized-bootstrap checks. Effects smaller than this margin remain unresolved. Harm at $\alpha{=}5$ survives bootstrap and paired-$t$ Holm correction over nine configurations; $\alpha{=}4$ survives only the bootstrap construction. The shuffled reward keeps token-credit magnitudes unchanged, but can change projected-gradient norms and Adam steps. Its $\alpha{=}5$ comparison also changes the rule for combining credit with the outcome. The $\alpha{=}4$ comparison in \Cref{tab:transition} uses the same rule. These controls show differences between the rewards (\Cref{app:stats,app:mechanism}).}
\label{tab:fullft}
\setlength{\tabcolsep}{0pt}\small
\begin{tabular*}{\textwidth}{@{\extracolsep{\fill}}lccrrl@{\hspace{2pt}}}\toprule
Training reward & $n$ & gain (final$-$start) & \multicolumn{1}{c}{vs.\ outcome} & \multicolumn{1}{c}{$d_z$} & regime \\\midrule
GRPO (outcome) & $14$ & $+0.056\,[+0.041,+0.071]$ & \multicolumn{1}{c}{--} & \multicolumn{1}{c}{--} & baseline \\
Monte-Carlo, $\alpha{=}0.25$ & $14$ & $+0.051\,[+0.037,+0.067]$ & $-0.005\,[-0.026,+0.018]$ & $-0.12$ & equivalent \\
Monte-Carlo, $\alpha{=}0.5$ & $14$ & $+0.052\,[+0.038,+0.066]$ & $-0.004\,[-0.023,+0.016]$ & $-0.09$ & equivalent \\
Monte-Carlo, $\alpha{=}1$ & $14$ & $+0.053\,[+0.041,+0.065]$ & $-0.003\,[-0.018,+0.012]$ & $-0.09$ & equivalent \\
Monte-Carlo, $\alpha{=}4$ & $6$ & $-0.049\,[-0.067,-0.029]$ & $-0.105\,[-0.144,-0.061]$ & $-1.83$ & harm (bootstrap only) \\
Monte-Carlo, $\alpha{=}5$ & $8$ & $-0.071\,[-0.096,-0.048]$ & $-0.124\,[-0.161,-0.087]$ & $-2.15$ & harm (all tests) \\
Shuffled Monte-Carlo, $\alpha{=}5$ & $5$ & $+0.038\,[+0.022,+0.054]$ & $-0.022\,[-0.061,+0.025]$ & $-0.41$ & inconclusive \\
\bottomrule\end{tabular*}
\end{table*}

\subsection{How do the choices for combining credit with the outcome affect the comparison?} 
\begin{figure*}[t]
\centering
\includegraphics[width=\linewidth]{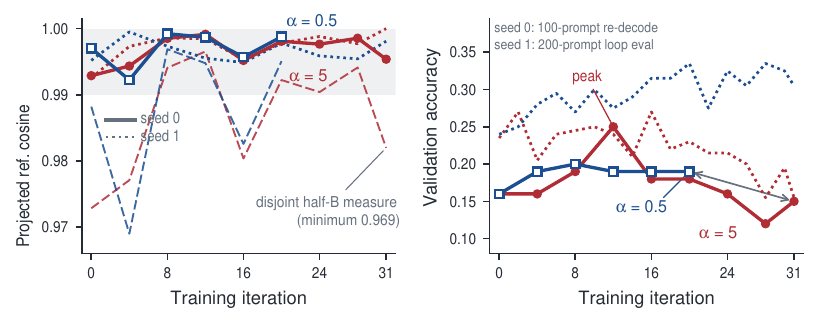}
\caption{\textbf{Reference agreement stays high as greedy accuracy falls.} \textbf{(left)} Cosine between the components outside $V$ of the full-parameter update and of the Monte-Carlo progress reference, during Qwen2.5-1.5B training. The minimum is $0.992$ for the full credit against the half-A reference, and $0.969$ for the disjoint half-B credit (long-dashed). Solid and dotted curves show seeds $0$ and $1$. \textbf{(right)} At $\alpha=5$, greedy accuracy peaks and then falls, while at $\alpha=0.5$ it remains stable or improves. Peak-to-final drops of $0.100$ and $0.115$ stay below the descriptive $0.15$ collapse threshold. Seed $0$ uses a $100$-prompt checkpoint re-evaluation and seed $1$ the training loop's $200$-prompt evaluation, so accuracy levels across seeds use different evaluation sets. These two runs give a descriptive comparison of reference cosine and greedy accuracy (\Cref{app:mechanism}).}
\label{fig:drift}
\end{figure*}

We next test choices made between estimating the reward and applying the optimizer. The coherent and shuffled rewards behave differently at high weight. Their paired difference-in-differences is $-0.128$ $[-0.155,-0.098]$ at $\alpha=5$, but those configurations use different rules for combining credit with the outcome. At $\alpha=4$, where the rules match, it is $-0.074$ $[-0.111,-0.040]$ over five common seeds (\Cref{app:mechanism}). These controls match the magnitudes of the token credits, but the magnitudes of the projected gradients and the actual Adam steps can still differ. Along two paired training runs, agreement with the reference remains high both in the low-weight configuration and in the degrading high-weight configuration (minimum $0.969$ against the strict half-B reference; \Cref{fig:drift}). Agreement thus stays high while accuracy falls, which points to how the credit is combined and how the optimizer acts, rather than to a decay in alignment, as questions for further study.

\textbf{Can gating help?} A prespecified expansion of the sign-opposition gate at $\alpha=1$ gives $+0.008$ $[-0.004,+0.021]$ against its matched outcome-only control ($n=32$, one-sided paired-$t$ $p=0.100$, $d_z=0.23$; \Cref{tab:gates}). The expansion includes eight exploratory seeds, and its interval includes a gain of $2.1$ percentage points. The smaller gate comparisons are descriptive. The sign-opposition configuration at $\alpha=3$ gives $+0.036$ $[-0.013,+0.080]$, on the same validation prompts and scoring procedure as its outcome-only control; no matched ungated comparison is available. Rollout-disagreement gating improves on the ungated reward with the same rule by $+0.056$ $[+0.007,+0.094]$ at $\alpha=3$ ($n=5$), but the paired-$t$ test does not exclude zero ($p=0.089$). Masking changes the total amount of credit as well as where it is placed, and centering afterward can put credit back on the masked positions (\Cref{app:gates}).

\begin{table*}[t]\centering
\caption{\textbf{Gating results depend on the matched control.} Qwen2.5-1.5B full fine-tuning; paired greedy-accuracy differences with $95\%$ seed-bootstrap intervals. The $n=32$ expansion includes eight pilot seeds. The $\alpha=3$ opposition row uses its matched $150$-problem outcome control; no ungated configuration was run on that evaluation set. The disagreement and ungated $\alpha=3$ rows use the same $200$-problem evaluation set and combination rule. Masking changes both placement and total credit. Rows with small $n$ are descriptive; the disagreement-versus-ungated paired-$t$ test gives $p=0.089$.}
\label{tab:gates}
\setlength{\tabcolsep}{0pt}\small
\begin{tabular*}{\textwidth}{@{\extracolsep{\fill}}lccrl@{\hspace{2pt}}}\toprule
Dense reward & $\alpha$ & $n$ & \multicolumn{1}{c}{vs.\ outcome ($d_z$)} & vs.\ ungated, same $\alpha$ \\\midrule
\multicolumn{5}{l}{\emph{Sign-opposition gate, $150$-problem evaluation:}}\\
sign-opposition, expanded & $1$ & $32$ & $+0.008\,[-0.004,+0.021]$\;($+0.23$) & -- \\
sign-opposition, exploratory & $3$ & $3$ & $+0.036\,[-0.013,+0.080]$\;($+0.76$) & -- \\
\midrule
\multicolumn{5}{l}{\emph{Disagreement gate and ungated reward, $200$-problem evaluation:}}\\
rollout-disagreement & $3$ & $5$ & $-0.007\,[-0.044,+0.031]$\;($-0.14$) & $+0.056\,[+0.007,+0.094]$ \\
ungated (ungated) & $3$ & $6$ & $-0.064\,[-0.107,+0.002]$\;($-0.82$) & -- \\
\bottomrule\end{tabular*}
\end{table*}

\subsection{How does the result vary across recipes?} Finally, we check how far the pattern extends. The Qwen2.5-7B sweep has $43$ dense runs, with $3$--$5$ seeds per configuration. No dense-reward configuration's mean gain exceeds that of the outcome-only baseline, which itself gains $0.21$, and the instability at high weights varies by seed (\Cref{tab:sevenb}). SmolLM3-3B ($n=10$ for the outcome-only configuration and the configurations at $\alpha=0.5$ and $5$) and a small NuminaMath study show similar point estimates at lower statistical resolution (\Cref{tab:crossfamily,tab:numina}). The single-seed Phi-4 pilots are positive, with paired gains of $+0.093$ and $+0.080$ at $\alpha=0.5$ and $5$ (\Cref{app:endtask}). These positive pilots show why each recipe needs its own replication.

\textbf{Do the reward source and the rollout budget matter?} With co-equal standardization, the Monte-Carlo reward also gives a small difference at low weight and a deficit at high weight ($n=5$; \Cref{tab:transition}). An ablation of the implicit reward with a frozen reference shows a large loss in accuracy ($d_z=-7.81$, $n=3$, descriptive). The implicit reward uses the current policy in its numerator, whereas PRIME uses a separately trained implicit PRM \citep{cui2025prime}. A frozen PRM trained on step labels also fails to improve accuracy when converted to credit centered within each completion, but its segmentation differs from the PRM's native setup and its scoring head was rebuilt, which limits this comparison (\Cref{tab:prm}). Finally, a $20$-iteration sweep over $k\in\{2,6,16\}$ continuations at $\alpha=1$ ($n=4$) detects no improvement at those budgets. Noise in the estimator remains a possible contributor, although in the matched-compute recipe continuation sampling contributes only $4\%$ of the second moment of the $k=4$ update (\Cref{sec:q1}). In the primary short-horizon recipe, then, high agreement between the gradient and the reference did not by itself pick out a reward that beat the declared margin, which is why the matched-compute studies of \Cref{app:matched} measure training outcomes directly.

\IfFileExists{figures/fig_pipeline.tex}{\begin{figure*}[t]
\centering
\includegraphics[width=\textwidth]{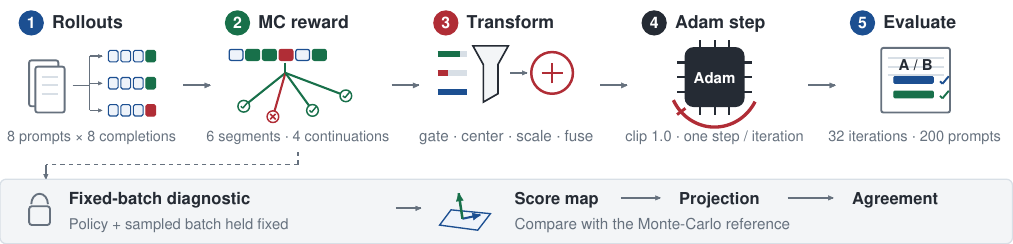}
\caption{\textbf{How a dense reward goes from its estimate to the training result.} In the primary full-fine-tuning recipe, the Monte-Carlo reward is estimated, optionally gated, centered, normalized, and combined with outcome-only GRPO before global gradient-norm clipping and one Adam step per iteration. Gating comes before centering, which can put credit back on cleared tokens. The $\alpha=5$ MC-within configuration uses co-equal standardization; the other primary weights and shuffled controls use the max-normalized rule. Training comparisons use the same seed, validation prompts, and scoring procedure. The diagnostic branch measures the fixed-batch score map and agreement with a finite-budget Monte-Carlo reference. The training branch measures same-seed greedy-accuracy differences after Adam updates (\Cref{tab:hparams,app:lmrecipe}).}
\label{fig:pipeline}
\end{figure*}
}{}

\textbf{Compute.} The equal-iteration experiments of \Cref{sec:equalround} used a node with eight H100-80GB GPUs. A $12$-iteration probe on one GPU takes about $20$ minutes, a $32$-iteration training run for the $1.5$B model takes $1$--$2$ GPU-hours, and a $7$B run on two GPUs takes $4$--$6$ GPU-hours. The $27$B/$35$B Qwen3.6 and $20$B probes use two-GPU sharding for inference, and the equal-iteration study as a whole, with its ablations of weight, continuation budget, combination rule, and model family, used a few hundred GPU-hours. The matched-compute studies of \Cref{sec:protocol} used one GPU per run (an H200 for all but $22$ of about $1{,}320$ GPU-hours) and report training time per run (\Cref{app:matched}). A systems-performance evaluation would also need the runtime and memory of each stage.

\subsection{Gate protocols and comparisons}
\label{app:gates}
Gates test whether concentrating the credit on the segments where it should matter most helps: the opposition gate keeps the credit that opposes the outcome advantage, and the disagreement gate favors segments whose continuations disagree.

The Monte-Carlo gates use the Qwen2.5-1.5B full-fine-tuning recipe: $32$ iterations, $8$ prompts, group size $8$, learning rate $10^{-5}$, no KL penalty, six segments, and $k=4$ continuations. The opposition gate keeps the entries whose sign is opposite to $A_g$, and the disagreement gate weights each segment by $p(1-p)$, where $p$ is the fraction of continuations that succeed. Both are applied before centering and normalization. We normalize the peak credit within each configuration, which does not match the total credit mass, the norm of the projected gradient, or the size of the actual optimizer step across configurations. Centering also changes which positions have nonzero credit. We trained no control that keeps the entries agreeing with the outcome's sign while preserving the mask.

\textbf{Expanded opposition gate.} The pilot at $\alpha=1$ gave a paired mean of $+0.013$ ($d_z=0.35$), with four of eight differences negative. We fixed the analysis plan before adding seeds and then analyzed the combined result, with $n=32$, once. All $64$ dense and control runs reached iteration $31$. The paired mean is $+0.008$, with $95\%$ bootstrap interval $[-0.004,+0.021]$, one-sided $t_{31}=1.31$, $p=0.100$, and $d_z=0.23$; $13/32$ differences are negative. No gated run meets the collapse criterion, a drop of $0.15$ from peak to final accuracy; the largest drop is $0.067$. The expansion includes the eight pilot seeds. Its post hoc power at the observed effect is $0.37$, and the interval remains compatible with an improvement of $2.1$ points.

\textbf{Other opposition weights.} On the same $150$-problem evaluation set and against matched outcome controls, $\alpha=0.5,2,3$ give respectively $+0.015$ $[-0.043,+0.073]$ ($n=4$), $+0.013$ $[-0.048,+0.075]$ ($n=4$), and $+0.036$ $[-0.013,+0.080]$ ($n=3$, $d_z=0.76$). A fourth $\alpha=3$ seed stopped at iteration $8$ and is excluded. All eleven completed runs remain below the collapse-event threshold (largest drop $0.080$). These exploratory comparisons are not corrected for multiple comparisons.

\textbf{Disagreement versus ungated reward.} At $\alpha=3$, both rewards use the same max-normalized combination rule, six segments, $k=4$, $200$-problem evaluation, and seeds $0$--$4$. Their paired difference is $+0.056$ $[+0.007,+0.094]$, $d_z=1.00$. The reported paired differences are $-0.035,+0.050,+0.070,+0.080,+0.115$; a paired-$t$ test gives $t_4=2.23$, $p=0.089$. The difference grows from $+0.006$ at iteration $15$ to $+0.029$ at $23$ and $+0.056$ at $31$. At $\alpha=1$, the matched difference is $+0.021$ $[+0.005,+0.033]$ (all $5^5$ bootstrap resamples), $d_z=1.13$, four of five differences positive, with $t_4=2.54$, $p=0.064$. That configuration's difference against outcome is $+0.009$ $[-0.022,+0.047]$, $d_z=0.21$. Both comparisons are exploratory, use five seeds, and have no correction for multiple comparisons. They change the mask while keeping the recipe fixed, but they do not hold the size of the resulting gradient or optimizer step fixed.

\section{Language-model setup and implementation}
\label{app:lmsetup}
\subsection{Policy, data, and training recipe}
\label{app:lmrecipe}
\begin{table*}[t]\centering
\caption{\textbf{End-task training recipe and evaluation target.} Settings for the primary full-fine-tuning runs in \Cref{tab:fullft,tab:gates,tab:transition}, with exceptions below and details for each comparison in \Cref{app:gates}. Each iteration takes one on-policy gradient step, with gradient-norm clipping but no importance ratio or ratio clipping. Training and Monte-Carlo continuations use temperature $1.0$ and top-$p$ $0.95$; reported validation accuracy uses greedy decoding, which measures a different target from the accuracy of the theory. The LoRA probe uses rank-$16$ query/value adapters, $12$ iterations, $\beta_{\mathrm{KL}}{=}0.04$, $k{=}6$, and MATH L3--5 training, with other shared settings listed in \Cref{app:probeimpl}.}
\label{tab:hparams}
\setlength{\tabcolsep}{5pt}\small
\begin{tabular}{@{}ll@{\hspace{14pt}}ll@{}}\toprule
Setting & Value & Setting & Value \\\midrule
policy & Qwen2.5-1.5B-Instruct & optimizer & Adam, $\mathrm{lr}=10^{-5}$ \\
trained parameters & all ($1.5$B, full FT) & grad-norm clip & $1.0$ \\
iterations & $32$ & inner epochs per iteration & $1$ (single step) \\
prompts per iteration & $8$ & microbatch (completions) & $4$ \\
group size $G$ & $8$ & ratio clipping & none (on-policy) \\
max new tokens & $380$ & $\beta_{\mathrm{KL}}$ & $0$ (full FT) \\
MC continuations $k$ & $4$ & MC continuation length & $200$ tokens \\
segments per completion & $6$ & $\alpha$ schedule & constant \\
train data & MATH L4--5 & val data & MATH L3--5, $N{=}200$ \\
eval cadence & every $2$ iterations & gain & final $-$ start val.\ acc. \\
\bottomrule\end{tabular}

\vspace{2pt}
{\small\textbf{Seeds per configuration} (contiguous from $0$; the $n$ column of each table gives the count). Outcome baseline and $\alpha{=}0.25,0.5,1$: $n{=}14$ after the pre-registered expansion. $\alpha{=}2,3,4$: $n{=}6$. $\alpha{=}5$: $n{=}8$. Shuffled control: $n{=}5$. Prover and PRM configurations use their own pre-registered seed lists (\Cref{app:generality}).

\textbf{Deviations.} $7$B (\Cref{tab:sevenb}): Qwen2.5-7B, $\alpha$ swept to $24$, $3$--$5$ seeds per configuration. Cross-family (\Cref{tab:crossfamily}): SmolLM3-3B. NuminaMath (\Cref{tab:numina}): train split swapped, $n{=}3$. Validation set: the gate, prover and cross-family configurations use a $150$-problem L4--5 split rather than the primary $200$-problem L3--5 split; valid contrasts require the same evaluation instrument. Coequal-standardization ablations (\Cref{tab:transition}) standardize the outcome advantage and the dense credit separately. The implicit-reward configurations use a frozen policy reference; they are not PRIME replications.}
\end{table*}

\textbf{Training and evaluation.} The main policy is \texttt{Qwen2.5-1.5B-Instruct}. LoRA probes train rank-$16$ adapters (adapter scale $32$) on query and value projections; the main training study updates all parameters. Each iteration samples eight completions for each of eight prompts, using temperature $1.0$, top-$p$ $0.95$, and at most $380$ new tokens. Adam uses $\beta_1=0.9$, $\beta_2=0.999$, no weight decay, constant learning rate, global gradient-norm clipping at $1.0$, and \texttt{bfloat16} weights. There is one on-policy gradient step per iteration, no inner-epoch loop, and no importance-ratio clipping. Dense weights stay constant throughout each run. The six-segment end-task credits use up to $200$ continuation tokens per prefix.

The verifier extracts the last number inside the final \verb|\boxed{}|, then falls back to the last number after \verb|####| or the last number in the completion. It strips a \verb|<think>| prefix and uses numerical tolerance $10^{-4}$, or exact string matching otherwise. The same verifier scores continuations. Validation uses greedy decoding. The reported greedy accuracy and the expected accuracy under sampling therefore measure different quantities. Because of top-$p$ sampling, finite continuation length, fixed segmentation, and the initial success prior, the probe reference also differs from an exact population gradient.

\textbf{Probe and full-fine-tuning configurations.} Five geometry runs train LoRA for $12$ iterations at learning rate $10^{-5}$ on MATH L3--5 with a KL-to-initial-policy penalty $0.04$ ($k_3$ estimator) and $k=6$ continuations, split three per half. We measure every reward gradient at the same checkpoints from outcome-only training. Three-seed LoRA accuracy runs use the same horizon, learning rate, and KL penalty, with $120$ validation prompts. Full-parameter probes and full fine-tuning omit the KL penalty. Main full fine-tuning uses $32$ iterations, learning rate $10^{-5}$, a $568$-problem MATH L4--5 training pool, and a $200$-problem MATH L3--5 evaluation set. Gains are the recorded final-minus-start accuracy (iteration $31$ minus iteration $0$). An initial recipe check gave outcome gain $+0.052$ $[+0.031,+0.074]$; the larger $n=14$ baseline gives $+0.056$ $[+0.041,+0.071]$. At learning rate $2\times10^{-6}$ and in the reported GSM8K control, the outcome baseline does not measurably improve.

\textbf{Rules for combining credit with the outcome.} Co-equal standardization scales the outcome and dense terms separately. The main Monte-Carlo configuration at $\alpha=5$ uses co-equal standardization, whereas the other primary weights and the shuffled controls use the max-normalized rule. Comparisons with outcome-only GRPO remain well defined, because it has no dense term to combine, but comparisons between dense-reward configurations must account for this difference in rules.

\textbf{Model identifiers.} The eight-checkpoint probe panel uses \texttt{Qwen/Qwen2.5-1.5B-Instruct}, \texttt{Qwen/Qwen2.5-Math-1.5B-Instruct}, \texttt{Qwen/Qwen2.5-Math-7B-Instruct}, \texttt{HuggingFaceTB/SmolLM2-1.7B-Instruct}, \texttt{microsoft/Phi-3.5-mini-instruct}, \texttt{openai/gpt-oss-20b}, \texttt{Qwen/Qwen3.6-27B}, and \texttt{Qwen/Qwen3.6-35B-A3B}. The final two rows use one seed and are descriptive. Larger models use two-GPU sharding. The panel reports measurements at these checkpoints.
\begin{table*}[!tbp]\centering
\caption{\textbf{Monte-Carlo reward aligns with the measured progress reference.} Mean$\pm$SD over probe rows pooled across drifted iterations of five runs; rows within a run are correlated. The MC-within row compares the full-rollout credit with its overlapping half-A reference $\widehat h_{\mathrm{MC}}$, so its $0.95$ projected cosine uses overlapping samples. A separate full-parameter half-B/half-A check averages $0.84$ over four probes (\Cref{app:probeimpl}). The final column measures cosine with the reference gradient. The shuffled control keeps token-credit magnitudes unchanged, but can change projected-gradient norms and Adam steps.}
\label{tab:alignment}
\setlength{\tabcolsep}{4pt}\small
\begin{tabular}{lccc}\toprule
Dense reward & $\|\Pperp g\|/\|g\|$ & $\cos(\Pperp g,\Pperp\widehat h_{\mathrm{MC}})$ & $\cos(g,\widehat h_{\mathrm{MC}})$ \\\midrule
implicit reward & $0.89\pm0.07$ & $-0.01\pm0.08$ & $+0.03\pm0.16$ \\
MC-within & $0.86\pm0.07$ & $+0.95\pm0.04$ & $+0.96\pm0.03$ \\
MC-shuffled control & $0.95\pm0.04$ & $-0.01\pm0.06$ & $-0.01\pm0.11$ \\
\bottomrule\end{tabular}\end{table*}

\begin{table*}[t]\centering
\caption{\textbf{Descriptive gradient-reference measurements across eight checkpoints.} Means and $95\%$ intervals are computed over each model's pooled probe rows. Rows from the same run are correlated, so these intervals are descriptive. The cosine columns compare raw gradients with the measured reference $\widehat h_{\mathrm{MC}}$. $\Delta$ is the paired MC-minus-shuffled difference. ``Family $r$'' correlates projected alignment and full-gradient reference cosine within a model's mixing family. Confinement residuals check the projector numerically. Single-seed rows are descriptive. Shuffling keeps token magnitudes unchanged, but can change projected-gradient norms (\Cref{app:lmsetup}).}
\label{tab:multimodel}
\setlength{\tabcolsep}{1.75pt}\small
\begin{tabular}{lcccccc}\toprule
 & $\|\Pperp\gout\|/\|\gout\|$ & \multicolumn{2}{c}{$\cos(g,\widehat h_{\mathrm{MC}})$} & $\Delta$\,(MC$-$shuf) & family $r$ & seeds/ \\
\cmidrule(lr){3-4}
Model & (residual) & MC & shuf & [$95\%$ CI] & [$95\%$ CI] & rows \\\midrule
Qwen2.5-1.5B & $0.010$ & $+0.97$ & $-0.05$ & $+1.01\,[+0.92,+1.12]$ & $0.98\,[0.97,0.99]$ & $2/5$ \\
Qwen2.5-Math-1.5B & $0.009$ & $+0.97$ & $-0.07$ & $+1.04\,[+0.93,+1.18]$ & $1.00\,[1.00,1.00]$ & $2/6$ \\
SmolLM2-1.7B & $0.010$ & $+0.95$ & $-0.04$ & $+0.99\,[+0.94,+1.03]$ & $1.00\,[1.00,1.00]$ & $2/5$ \\
Phi-3.5-mini-3.8B & $0.011$ & $+0.93$ & $-0.02$ & $+0.95\,[+0.89,+1.01]$ & $0.98\,[0.95,0.99]$ & $2/6$ \\
Qwen2.5-Math-7B & $<\!10^{-3}$ & $+0.97$ & $-0.08$ & $+1.04\,[+1.01,+1.09]$ & $0.99\,[0.99,1.00]$ & $1/4$ \\
gpt-oss-20B & $0.009$ & $+0.86$ & $-0.03$ & $+0.90\,[+0.67,+1.08]$ & $1.00\,[0.99,1.00]$ & $1/3$ \\
Qwen3.6-27B (dense) & $0.010$ & $+0.93$ & $+0.06$ & $+0.87\,[+0.71,+0.98]$ & $0.89\,[0.76,0.96]$ & $1/3$ \\
Qwen3.6-35B-A3B (MoE) & $0.009$ & $+0.98$ & $+0.02$ & $+0.96\,[+0.93,+1.02]$ & $0.98\,[0.96,0.99]$ & $1/3$ \\
\bottomrule\end{tabular}\end{table*}

\subsection{Statistical reporting}
\label{app:stats}
Unless stated otherwise, end-task intervals are $95\%$ percentile-bootstrap intervals with $10^4$ seed resamples. A gain row resamples per-seed final-minus-start gains; a paired row resamples same-seed dense-minus-outcome differences. These intervals describe uncertainty across seeds on the fixed validation examples. The standardized paired effect is $d_z=\overline\Delta/s_\Delta$, without a small-sample correction. Reported bootstrap $p$ values are twice the proportion of resampled means on the opposite side of zero. Because such $p$ values can overstate the evidence when $n$ is small, we give paired-$t$ sensitivity checks for the principal claims. Brackets denote intervals; $\pm$ denotes the sample SD unless explicitly labeled SE. SD uses the $n-1$ variance denominator; $\mathrm{SE}=\mathrm{SD}/\sqrt n$.

\textbf{Small samples.} At $n=3$, a $95\%$ percentile-bootstrap interval equals the observed minimum and maximum (an all-one-seed resample has probability $3^{-3}=0.037>0.025$), so these rows describe the observed range. For the $n=6$, $\alpha=3$ transition, exhaustive enumeration of all $6^6=46{,}656$ resamples gives $[-0.108,+0.001]$, and the paired-$t$ test gives $t_5=-2.01$, $p=0.101$, with five of six differences negative. For the other paired rows with $n\le7$, exhaustive enumeration does not change whether the interval crosses zero. Probe rows pooled across checkpoints are correlated, so their spread is descriptive.

\textbf{Groups of comparisons and registration.} The primary harm analysis includes nine Monte-Carlo weight configurations on the Qwen2.5-1.5B, $200$-problem, iteration-$31$ recipe: five in Table~\ref{tab:fullft}, two transition weights, and two co-equal-standardization configurations in Table~\ref{tab:transition}. The three low-weight equivalence tests form a separate Holm family. This group of comparisons excludes other models, datasets, evaluation sets, and the $20$-iteration estimator sweep; the source study has $31$ Monte-Carlo configurations at iteration $31$ ($30$ after merging two with the same configuration), $22$ of them outside the nine-configuration family. Analysis plans were locked before adding data for the low-weight seed expansion, complementary-prover confirmation, SmolLM3 expansion, and $n=32$ gate expansion. The primary low-weight expansion adds seeds $8$--$13$ or $5$--$13$ to existing configurations; the gate expansion retains its eight pilot seeds. Because the expansions add seeds to existing configurations, they extend those configurations rather than replicate them independently.

\textbf{Additional drift seeds.} Two completed $\alpha=5$ drift-probe seeds extend the paired estimate from $-0.124$ $[-0.161,-0.087]$, $d_z=-2.15$, $n=8$, to $-0.123$ $[-0.153,-0.093]$, $d_z=-2.40$, $n=10$. One eligible $\alpha=0.5$ co-equal-standardization seed changes $-0.004$ $[-0.032,+0.034]$ to $+0.001$ $[-0.026,+0.033]$; its second drift run stops at iteration $20$ and is excluded under the stated iteration-$28$ admission rule. Drift seeds use their own outcome controls run at the same time. These runs serve as a sensitivity analysis; the nine-configuration family and the seed sets of the main tables are unchanged.

\textbf{Validation sets.} The main $200$-problem MATH L3--5 set is used for the weight sweep, low-weight equivalence, PRM, implicit-reward, and disagreement-versus-ungated comparisons. A $150$-problem L4--5 set is used for opposition gates, prover, and cross-family configurations. A separate $115$-problem set is used only for the degenerate-slice ablation. Starting accuracies differ: Qwen2.5-1.5B/OLMo-2 means are $0.24$--$0.26$ on the $200$-problem set and $0.15$--$0.20$ on the $150$-problem set; SmolLM3 starts at $0.32$--$0.34$, Phi-4-mini at $0.25$, NuminaMath at $0.28$--$0.29$, and the degenerate slice at $0.03$--$0.10$. We therefore compare configurations only within an evaluation set. Table~\ref{tab:gates} uses the available opposition estimate from the same evaluation set and omits the unmatched ungated comparison.

\textbf{Deficits and collapse events.} A negative paired configuration mean is different from a collapse within one run. We define a collapse event descriptively as final accuracy at least $0.15$ below that run's peak. We chose this threshold after inspecting variation in baseline drops. At $1.5$B, the Monte-Carlo $\alpha=5$ and $4$ configurations have $0/8$ and $0/6$ such events; their largest drops are $0.14$ and $0.11$, versus $0.055$ for outcome-only. Still, all eight and six seeds have negative absolute gains and paired differences. One of five $\alpha=3$ co-equal-standardization seeds has a drop exactly $0.150$. One $\alpha=5$ and two $\alpha=4$ seeds peak at iteration $0$, so a rise-then-decay trajectory does not describe every deficit. Across $85$ eligible gate/control runs ($86$ launched, one stopping at iteration $8$), no drop reaches $0.15$; the largest is $0.080$.

\textbf{Per-configuration bootstrap values.} For Table~\ref{tab:fullft}, the two-sided bootstrap $p$ values are $0.638,0.740,0.714$ at $\alpha=0.25,0.5,1$, below $10^{-3}$ at $4,5$, and $0.304$ for the shuffled control. For Table~\ref{tab:transition}, they are $0.014,0.060$ for weights $2,3$; $0.747,0.0004$ for co-equal-standardization weights $0.5,3$; $0.134,0.390,0.248$ for shuffled weights $1,3,4$; and below $10^{-3}$ for the three-seed implicit-reward row.

\subsection{Gradient probes}
\label{app:probeimpl}
This subsection describes how the gradient tests of \Cref{sec:tests} are computed at language-model scale.

\textbf{Score-gradient map.} At a fixed checkpoint, we compute a credit gradient with one weighted backward pass:
\[
 \grad(w)=\nabla_\theta\sum_{n,t}w_{n,t}\,\mathbf1[\mathrm{generated}]\log\pi_\theta(a_{n,t}).
\]
We accumulate gradients over completion microbatches using logsumexp token probabilities. The probe builds $V$ from at most $24$ completions and forms $\gout$ from the same subsample, so the least-squares projection uses this fixed, subsampled $V$. It costs $O(N)$ forward passes, with the backward graphs kept in memory. The reported outcome residual $\goutVresid$ checks the numerical projection rather than the theorem itself; with full-parameter gradients, the residuals are as small as $2\times10^{-13}$.

\textbf{Progress reference and sample overlap.} Following Monte-Carlo prefix-value methods \citep{kazemnejad2024vineppo,wang2023mathshepherd}, the probe uses three interior cuts and the endpoint, with up to $200$ continuation tokens, temperature $1.0$, top-$p$ $0.95$, and initial success prior $0.5$. Each segment's credit is the change in success rate across it. Half A of the continuations defines $\widehat h_{\mathrm{MC}}$, and half B, where labeled, provides a candidate credit from disjoint samples. The credit computed from all rollouts, which the primary pooled measurement uses, overlaps with half A. After projection, it differs from the centered full credit only by the projection residual (agreement within $1.4\times10^{-3}$ over $135$ rows), so the reported agreement with the reference, $\alignAB$, uses overlapping samples. The full-parameter alignments using disjoint samples are $0.97,0.66,0.89,0.82$, mean $0.84$. The overlapping full-parameter mean is $0.96$, with mean$\pm1.96$SE interval $[0.92,0.99]$ (percentile bootstrap $[0.92,0.98]$), shuffled mean $-0.05$, and off-$V$ fraction $0.93$.

\textbf{Controls and interpolation.} Permuting values within each completion preserves the distribution of token-credit magnitudes. It does not, in general, preserve $\|\Pperp S^\top w\|$, because the Gram matrix of the scores is not isotropic, and it need not preserve the actual Adam step. The interpolation family is $w_\lambda=\lambda\widehat{\mathrm{mc}}_B+(1-\lambda)\xi$, $\lambda\in\{0,1/4,1/2,3/4,1\}$, with unit-normalized, zero-sum half-B and Gaussian rewards. The fractions of their updates that lie outside $V$ are similar, but their budgets are not exactly matched. The reported within-bin and magnitude-adjusted correlations are $\famWithin$ and $\famPartialTwo$; the aligned-minus-shuffled reference-cosine gap is $\alignedVsShuf$. These measurements describe agreement with the measured reference at fixed probes.

\textbf{Implicit reward.} This credit is $w_{n,t}=\log\pi_\theta(a_{n,t})-\log\pi_{\mathrm{ref}}(a_{n,t})$, with the actor itself in the numerator and the initial policy in the denominator. We do not scale it in the probes, because the cosine and the fraction outside $V$ do not depend on scale. Averaged over each run, the cosines of the five runs stabilized by a KL penalty are $-0.05,+0.06,+0.01,+0.01,+0.10$; without the KL penalty, values at single checkpoints reach $\pm\lrDriftMax$, so alignment with the estimated reference can change sign. At initialization, the credit is zero because the actor and the reference are the same, so these summaries use only iterations after the first. This implicit reward differs from PRIME's separate implicit PRM trained with outcome labels \citep{cui2025prime}.

\subsection{End-task results and controls}
\label{app:endtask}
This subsection gives the end-task results behind \Cref{sec:equalround}, with the checks behind each claim.

\textbf{Choice of recipe.} A short LoRA recipe and a GSM8K recipe barely improved the outcome-only baseline, so they could not separate the rewards; the main study therefore uses full fine-tuning on MATH.

\textbf{Low-weight equivalence.} At $n=14$ paired seeds, the $90\%$ percentile-bootstrap intervals for $\alpha=0.25,0.5,1$ are respectively $[-0.023,+0.014]$, $[-0.020,+0.013]$, and $[-0.016,+0.010]$, with TOST $p=0.0095,0.0033,0.0004$. All pass Holm correction over the three configurations at the $\pm0.03$ margin. Paired-$t$ TOST gives $p=0.025,0.012,0.003$, also passing Holm, and studentized-bootstrap $90\%$ intervals also lie inside the margin. The reported minimum effects detectable by superiority tests are $0.025$--$0.035$ at $80\%$ power; estimated power at an effect equal to the margin ranges from $0.67$ to $0.92$. These results support equivalence at the stated resolution.

\textbf{High-weight deficits.} The $\alpha=5$ deficit remains significant after bootstrap and paired-$t$ Holm correction over all nine primary weight configurations. The $\alpha=4$ deficit passes bootstrap Holm correction but misses the paired-$t$ step ($p=0.0065$ versus threshold $0.0063$). Unadjusted one-sided paired-$t$ values are $0.0033$ and $0.00025$ for $\alpha=4,5$; studentized-bootstrap $95\%$ intervals are $[-0.155,-0.026]$ and $[-0.172,-0.074]$. Using only seven configurations for correction would change the $\alpha=4$ conclusion, but the reported claim uses all nine. The co-equal-standardization $\alpha=3$ configuration passes the bootstrap correction, while the $\alpha=2$ transition passes only the uncorrected test. The implicit-reward row has $n=3$ and $d_z=-7.81$, so it remains descriptive despite its large deficit.
\begin{table*}[t]\centering
\caption{\textbf{Changing reward weight and the credit combination rule.} Qwen2.5-1.5B full fine-tuning; gains and paired differences use greedy accuracy and the $200$-problem evaluation set. Intervals and $d_z$ follow \Cref{tab:fullft}. The $\alpha=2,3$ transition rows do not survive Holm correction. The middle block applies co-equal standardization to the same Monte-Carlo reward; the bottom block uses the policy's implicit reward against a frozen reference. These ablations use rewards that differ from PRIME's learned implicit PRM. Shuffled-control intervals do not establish equivalence at $\pm0.03$. Estimates with small $n$ are descriptive (\Cref{app:stats}).}
\label{tab:transition}
\setlength{\tabcolsep}{3.5pt}\small
\begin{tabular}{lccrl}\toprule
Training reward & $n$ & gain (final$-$start) & \multicolumn{1}{c}{vs.\ outcome ($d_z$)} & regime \\\midrule
MC-within, $\alpha{=}2$ & $6$ & $-0.009\,[-0.048,+0.026]$ & $-0.065\,[-0.101,-0.016]$\;($-1.08$) & transition \\
MC-within, $\alpha{=}3$ & $6$ & $-0.008\,[-0.040,+0.038]$ & $-0.064\,[-0.107,+0.002]$\;($-0.82$) & transition \\
MC-shuffled control, $\alpha{=}1$ & $5$ & $+0.035\,[+0.018,+0.063]$ & $-0.025\,[-0.056,+0.006]$\;($-0.62$) & uncertain \\
MC-shuffled control, $\alpha{=}3$ & $5$ & $+0.040\,[+0.023,+0.065]$ & $-0.020\,[-0.058,+0.027]$\;($-0.37$) & uncertain \\
MC-shuffled control, $\alpha{=}4$ & $5$ & $+0.022\,[-0.026,+0.070]$ & $-0.038\,[-0.094,+0.023]$\;($-0.49$) & uncertain \\
\midrule
\multicolumn{5}{l}{\emph{Co-equal standardization (same MC reward, $k$, segments, seeds $0$--$4$):}}\\
MC-within, $\alpha{=}0.5$ & $5$ & $+0.056\,[+0.031,+0.077]$ & $-0.004\,[-0.032,+0.034]$\;($-0.09$) & tie (n.c.) \\
MC-within, $\alpha{=}3$ & $5$ & $-0.080\,[-0.122,-0.029]$ & $-0.140\,[-0.186,-0.066]$\;($-1.74$) & degrades \\
\midrule
\multicolumn{5}{l}{\emph{Self-log-ratio (co-equal standardization, frozen reference; $n{=}3$, descriptive):}}\\
Self-log-ratio ($\alpha{=}0.5$) & $3$ & $-0.260\,[-0.285,-0.220]$ & $-0.313\,[-0.355,-0.275]$\;($-7.81$) & \textbf{collapse} \\
\bottomrule\end{tabular}
\end{table*}

\textbf{Curve summaries.} We reanalyze the $n=14$ low-weight runs with three measures: iterations to first reach the baseline's mean final accuracy $0.308$ (baseline median $20/32$ iterations), area under the accuracy curve relative to start (baseline $+0.038$), and per-run peak (baseline mean $0.324$). The paired dense-minus-outcome estimates, in that order, are:
\begin{itemize}[leftmargin=*,itemsep=1pt]
\item $\alpha=0.25$: $-0.1$ $[-4.0,+3.3]$ iterations, $-0.001$ $[-0.014,+0.014]$ area, and $-0.008$ $[-0.018,+0.003]$ peak.
\item $\alpha=0.5$: $-0.2$ $[-6.3,+6.1]$, $+0.000$ $[-0.013,+0.013]$, and $+0.002$ $[-0.008,+0.013]$.
\item $\alpha=1$: $+4.4$ $[-0.7,+9.4]$ ($d_z=0.44$), $-0.007$ $[-0.019,+0.006]$, and $-0.004$ $[-0.014,+0.005]$.
\end{itemize}
These post hoc summaries detect no speed advantage. Runs that never reach the target are coded as $32$ iterations, so the time-to-target statistic is censored.

\textbf{Continuation-budget sweep.} At $\alpha=1$, the $k\in\{2,6,16\}$ study has four seeds per setting and ends at iteration $19$ for all twelve runs. No $32$-iteration comparison is available. It detects no difference at that horizon. A separate co-equal-standardization $\alpha=1$ set also ends before the primary iteration-$31$ horizon; neither short study is included in the nine-configuration group of comparisons.

\textbf{Positive pilots.} The Phi-4 outcome-only configuration has three seeds and mean gain $+0.142$. Its dense $\alpha=0.5$ and $5$ configurations each have one seed, with same-seed paired gains $+0.093$ and $+0.080$. These positive pilots exceed the primary equivalence margin but provide no seed-level uncertainty estimate. They remain unconfirmed, but they limit any broad negative claim.

\textbf{Qwen2.5-7B sweep.} Full fine-tuning uses two GPUs, six completions for each of six prompts, $32$ iterations, learning rate $10^{-5}$, and $150$ validation prompts. Seeds $0$--$4$ are used except at $\alpha=4$, which has seeds $0$--$2$. The outcome baseline gains $+0.209$, while no dense-reward configuration mean exceeds it across $43$ dense runs and weights up to $24$. The descriptive collapse-event count is $3/5$ at each of $\alpha=5,8$, versus $0/5$ for outcome-only (largest baseline drop $0.11$). One seed at each weight ends at zero accuracy after peaks $0.46$ and $0.43$. High-weight seeds that do not collapse stay close enough to the baseline that configuration means at $\alpha=12,16,24$ remain positive. Expanding the high-weight configurations from three to five seeds increased event counts and changed the $\alpha=5$ mean gain from positive to negative.
\begin{table*}[t]\centering
\caption{\textbf{Accuracy losses in the $7$B sweep vary across seeds.} Full fine-tuning of Qwen2.5-7B on MATH L4--5. Gain is final$-$start greedy validation accuracy (mean$\pm$SD); ``vs.\ outcome'' gives the paired difference, $95\%$ seed-bootstrap CI, and paired effect size $d_z$. All rows are descriptive ($n{\le}5$); tests are not adjusted for multiple comparisons. A collapse event is a final accuracy at least $0.15$ below that run's peak; this threshold was selected after inspecting baseline drops. The outcome-only configuration has $0/5$ events, compared with $3/5$ at each of $\alpha{=}5$ and $8$. These counts describe instability in this recipe. The $\alpha{=}4$ comparison uses only three shared seeds, so subtracting the displayed configuration means does not give its paired difference.}
\label{tab:sevenb}
\setlength{\tabcolsep}{3.5pt}\small
\begin{tabular}{lccrc}\toprule
Training reward & $n$ & gain (final$-$start) & \multicolumn{1}{c}{vs.\ outcome ($d_z$)} & collapse events \\\midrule
GRPO (outcome) & $5$ & $+0.209\pm0.048$ & \multicolumn{1}{c}{--} & $0/5$ \\
MC-within, $\alpha{=}0.5$ & $5$ & $+0.133\pm0.090$ & $-0.076\,[-0.167,-0.013]$\;($-0.75$) & $1/5$ \\
MC-within, $\alpha{=}1$ & $5$ & $+0.204\pm0.071$ & $-0.005\,[-0.091,+0.079]$\;($-0.05$) & $0/5$ \\
MC-within, $\alpha{=}2$ & $5$ & $+0.147\pm0.091$ & $-0.063\,[-0.159,+0.036]$\;($-0.49$) & $1/5$ \\
MC-within, $\alpha{=}4$ & $3$ & $+0.124\pm0.047$ & $-0.100\,[-0.180,\;0.000]$\;($-1.09$) & $0/3$ \\
MC-within, $\alpha{=}5$ & $5$ & $-0.008\pm0.220$ & $-0.217\,[-0.411,-0.051]$\;($-0.96$) & $3/5$ \\
MC-within, $\alpha{=}8$ & $5$ & $-0.063\pm0.178$ & $-0.272\,[-0.385,-0.151]$\;($-1.80$) & $3/5$ \\
MC-within, $\alpha{=}12$ & $5$ & $+0.157\pm0.053$ & $-0.052\,[-0.107,-0.003]$\;($-0.79$) & $0/5$ \\
MC-within, $\alpha{=}16$ & $5$ & $+0.144\pm0.069$ & $-0.065\,[-0.143,+0.009]$\;($-0.66$) & $0/5$ \\
MC-within, $\alpha{=}24$ & $5$ & $+0.077\pm0.173$ & $-0.132\,[-0.288,+0.021]$\;($-0.64$) & $2/5$ \\
\bottomrule\end{tabular}
\end{table*}

\subsection{Controls for interpreting performance loss}
\label{app:mechanism}
The high-weight deficits raise a question: why does a reward that aligns well with the reference hurt training? These controls check the most direct explanations: a decay of alignment during training, where the credit falls, the coherence of the credit, and changes in length and entropy.

\textbf{Alignment along training.} Paired full-fine-tuning runs at $\alpha=0.5,5$, two seeds each, save checkpoints every four iterations. Full-parameter probes use $G=16$, $k=10$. The alignment between the full credit and the overlapping half-A reference stays at or above $0.992$ (from $0.993$ to $1.000$ for seed $1$). With disjoint half-B samples, it ranges from $0.969$ to $1.000$; the shuffled reward's alignment remains near zero; and the fraction of the update outside $V$ ranges from $0.62$ to $0.98$. In seed $1$, the $\alpha=5$ configuration declines from peak $0.27$ to $0.155$, while the low-weight configuration gains $+0.07$. The two high-weight probe runs' drops, $0.100$ and $0.115$, are below the descriptive $0.15$ event threshold. These observations show that agreement with the reference persists even as greedy accuracy falls.

\textbf{Credit on segments where continuations disagree.} At initialization, $12$ prompts with four completions each are cut at eight interior points plus the endpoint, yielding $432$ segments and $2{,}592$ continuations at $K=6$. Credit magnitude is $|\widehat p_s-\widehat p_{s-1}|$, with $\widehat p_0=0.5$. A continuation-success variance above $0.1$ means at least one success and one failure: the smallest nonzero variance at $K=6$ is $(1/6)(5/6)=0.139$. Such segments account for $19.7\%$ of segments and carry $38.4\%$ of credit; the remaining $61.6\%$ lies on unanimous segments. For disagreement segments, the ratio of credit share to segment share is $1.95\times$, with correlation $0.30$ and mean magnitude $0.22$ versus $0.085$.

\textbf{Shuffled-credit comparison.} At $\alpha=5$, the coherent and shuffled configurations use different rules for combining credit with the outcome; their paired difference-in-differences is $-0.128$ $[-0.155,-0.098]$, $d_z=-3.46$, over five common seeds. At $\alpha=4$, both use the max-normalized rule, and the difference is $-0.074$ $[-0.111,-0.040]$, $d_z=-1.62$, $n=5$, with gains $-0.052$ and $+0.022$ on that common-seed subset. Matching the rule removes one confound; the norms of the projected gradients, the clipping, the policy KL, and the Adam steps can still differ between the two configurations.

\textbf{Length, entropy, and drift.} On a fixed $100$-prompt subsample, response length stays within $1\%$ of the low-weight configuration through $28$ iterations while high-weight accuracy moves from $0.25$ to $0.12$; the fraction of degenerate completions does not rise. High-weight entropy rises from $0.21$ to $0.58$, versus $0.18$--$0.20$ for low weight, and KL from the initial policy is about $30$ times larger. These changes occur together and are consistent with several optimization explanations, such as rising entropy or over-optimization of the proxy reward.

\subsection{Other families, provers, reward sources, and datasets}
\label{app:generality}
To see whether the equal-iteration pattern depends on the model family, on where the progress estimates come from, or on the dataset, we repeat the comparison with each of them changed.

\textbf{SmolLM3-3B.} The same trainer uses \texttt{HuggingFaceTB/SmolLM3-3B}, with extended thinking disabled in the chat template. Outcome, $\alpha=0.5$, and $\alpha=5$ configurations were expanded to seeds $0$--$9$ under a fixed paired-$t$ TOST analysis plan at $\pm0.03$ and one-sided harm test at $0.005$. The low-weight $90\%$ interval $[-0.034,+0.005]$ excludes a gain of $0.03$ (reported $p=0.001$) and extends $0.004$ beyond the lower equivalence boundary. The high-weight paired effect is $-0.092$, $d_z=-0.77$, one-sided $p=0.02$, short of the specified $0.005$ criterion. Nine of ten differences are negative; two seeds drop $0.34$ and $0.17$ from peak, one ends at gain $-0.267$, and a third drop is $0.14$. At $\alpha=2$, one of three runs has a negative final gain. These results reproduce the qualitative pattern of the primary study at lower statistical resolution.
\begin{table*}[t]\centering
\caption{\textbf{SmolLM3 reproduces the primary pattern at lower resolution.} Greedy validation-accuracy gain after full fine-tuning on MATH L4--5, with seed-bootstrap $95\%$ CIs and paired same-seed differences from outcome-only. The outcome, $\alpha{=}0.5$, and $5$ configurations use a preregistered $n{=}10$ expansion. At $\alpha{=}0.5$, the paired $90\%$ $t$ interval $[-0.034,+0.005]$ falls just outside equivalence at $\pm0.03$. At $\alpha{=}5$, the mean paired difference is $-0.092$ (one-sided $p{=}0.02$, against a prespecified harm threshold of $0.005$); two of ten runs meet the descriptive collapse-event criterion. The $\alpha{=}2$ row is descriptive ($n{=}3$).}
\label{tab:crossfamily}
\setlength{\tabcolsep}{4pt}\small
\begin{tabular}{lccccc}\toprule
Family / reward & $n$ & gain & $95\%$ CI & vs.\ outcome & regime \\\midrule
SmolLM3-3B, outcome & $10$ & $+0.153$ & $[+0.143,+0.167]$ & -- & baseline \\
\quad $\alpha{=}0.5$ & $10$ & $+0.139$ & $[+0.114,+0.162]$ & $-0.015$ & inconclusive \\
\quad $\alpha{=}2$ & $3$ & $+0.087$ & $[-0.093,+0.193]$ & $-0.062$ & deficit (desc.) \\
\quad $\alpha{=}5$ & $10$ & $+0.061$ & $[-0.024,+0.123]$ & $-0.092$ & deficit (desc.) \\
\bottomrule\end{tabular}
\end{table*}

\textbf{Complementary prover.} A frozen \texttt{Qwen2.5-Math-1.5B-Instruct} supplies continuations from the actor's prefixes, following the complementary-prover motivation of \citet{setlur2024pav}; the actor receives the resulting dense credit. At $\alpha=1$, three paired differences are $+0.007,-0.040,+0.040$, giving $+0.002$ $[-0.040,+0.040]$. A three-seed $\alpha=0.5$ pilot gives $+0.051$ $[+0.007,+0.100]$. This was one of six exploratory comparisons. It motivated an $n=8$ confirmation, which required a paired interval excluding zero and a result that passed Holm correction. The expanded estimate is $+0.023$ $[-0.007,+0.052]$, $d_z=0.49$, from differences $+0.100$, $+0.007$, $+0.047$, $-0.060$, $+0.020$, $+0.040$, $+0.027$, $+0.000$. It falls short of the first criterion, and its interval still allows meaningful gains.

\textbf{Step-labeled PRM.} The frozen \texttt{Qwen2.5-Math-PRM-7B} \citep{zhang2025qwenprm} scores actor-generated completions at the Monte-Carlo configuration's interior cuts, using the step separator, canonical system prompt, and boxed-answer format. Differences in the PRM's positive-class probabilities form a credit, with initial prior $0.5$, combined with the outcome by the same rule. These differ from the model's native double-newline step boundaries, which may weaken the reward. We rebuilt the scoring head from the released weights on a standard backbone; a convention check gives approximately $1.0$ for correct chains and $0.0$ for a corrupted step.

All $16$ planned PRM runs completed. At $\alpha=0.5$, the paired effect is $-0.009$ $[-0.049,+0.024]$, $n=8$, superiority one-sided $p=0.66$; TOST fails ($p_{\mathrm{lower}}=0.16$). At $\alpha=2$, it is $-0.090$ $[-0.125,-0.040]$, $n=3$. At $\alpha=5$, it is $-0.133$ $[-0.188,-0.077]$, $n=5$, one-sided $p=0.007$, above the specified $0.005$ harm criterion. The latter two comparisons remain descriptive. These results apply to this conversion and scoring pipeline.
\begin{table*}[!htb]\centering
\caption{\textbf{A frozen step-labeled PRM gives inconclusive low-weight and negative high-weight estimates.} Qwen2.5-1.5B full fine-tuning on MATH L4--5 using differences between Qwen2.5-Math-PRM-7B scores as within-completion credit. Gain is final$-$start greedy accuracy; intervals are $95\%$ seed-bootstrap CIs and contrasts are paired against outcome-only. All $16$ planned runs completed. At $\alpha{=}0.5$, equivalence at $\pm0.03$ is not established; the interval still permits harm beyond the margin. Higher-weight deficits are descriptive, and the $\alpha{=}5$ one-sided $p{=}0.007$ misses the prespecified $0.005$ harm threshold. The step boundaries differ from the native PRM setup, and the scoring head was rebuilt. These changes may weaken the reward and limit conclusions about the released PRM (\Cref{app:generality}).}
\label{tab:prm}
\setlength{\tabcolsep}{4pt}\small
\begin{tabular}{lccrl}\toprule
Training reward & $n$ & gain (final$-$start) & \multicolumn{1}{c}{vs.\ outcome ($d_z$)} & regime \\\midrule
GRPO (outcome) & $14$ & $+0.056\,[+0.041,+0.071]$ & \multicolumn{1}{c}{--} & baseline \\
PRM, $\alpha{=}0.5$ & $8$ & $+0.044\,[+0.007,+0.076]$ & $-0.009\,[-0.049,+0.024]$\;($-0.15$) & inconclusive \\
PRM, $\alpha{=}2$ & $3$ & $-0.028\,[-0.035,-0.020]$ & $-0.090\,[-0.125,-0.040]$\;($-2.03$) & deficit (desc.) \\
PRM, $\alpha{=}5$ & $5$ & $-0.073\,[-0.123,-0.023]$ & $-0.133\,[-0.188,-0.077]$\;($-1.89$) & deficit (desc.) \\
\bottomrule\end{tabular}
\end{table*}

\subsection{Degenerate-group channel ablation}
\label{app:degencontrol}
Groups whose completions all fail give outcome-only GRPO no signal. Dense rewards are often motivated as a way to learn from such groups, whereas prompt selection (\Cref{sec:selection}) avoids them, so we ask whether dense rewards help where they dominate. On a curated MATH slice, the fraction of groups with identical outcomes is $0.894$ (per-iteration range $0.50$--$1.00$). Qwen2.5-1.5B is full-fine-tuned for $32$ iterations with group size eight, co-equal standardization, $\alpha=1$, and seeds $0$--$2$. The dense credit is zero-sum within each completion, which removes its completion-mean term but generally leaves a component in $V$; at language-model scale, the measured component in $V$ carries about $0.26$ of the squared norm. Outcome gains are $+0.017,+0.009,+0.052$ (mean $+0.026$), versus $-0.070,+0.009,+0.009$ (mean $-0.017$) for the dense-reward configuration. The paired effect is $-0.043$ $[-0.087,0.000]$ over three seeds (descriptive).

All completions in the degenerate groups are wrong to the reported precision, and training accuracy is about $0.02$. Co-equal standardization amplifies a low-signal progress reward, and a frozen math-specialized prover succeeds on $13\%$ of $150$ tested prompts with $k=4$, so progress is hard to estimate in this setting: where the outcome gradient is zero, a dense signal can exist without helping. Attributing the gains of other degenerate-group methods \citep{palandye2026granularity,agrawal2026verigate} to a particular channel would need a control that trains only the completion-mean term.

\textbf{NuminaMath.} A $1{,}500$/$200$ train/validation slice of NuminaMath-CoT uses the cn\_k12 and synthetic-AMC sources, filtered to numeric answers supported by the verifier. It uses the same trainer for $32$ iterations and seeds $0$--$2$; the outcome baseline gains $+0.098$ (\Cref{tab:numina}). These rows use small samples and are descriptive.
\begin{table*}[!htb]\centering
\caption{\textbf{A small NuminaMath study points the same way.} Qwen2.5-1.5B full fine-tuning on a numeric-answer NuminaMath slice, using the primary recipe and verifier. Gain is final$-$start greedy validation accuracy (mean and $95\%$ seed-bootstrap CI); ``vs.\ outcome'' is the paired same-seed difference. At $n{=}3$, the $\alpha{=}0.5$ estimate is $+0.013$ and the $\alpha{=}5$ estimate is $-0.030$, with all three high-weight runs below their paired baselines. These observations motivate larger comparisons.}
\label{tab:numina}
\setlength{\tabcolsep}{4pt}\small
\begin{tabular}{lccccc}\toprule
Dataset / reward & $n$ & gain & $95\%$ CI & vs.\ outcome & regime \\\midrule
NuminaMath, outcome & $3$ & $+0.098$ & $[+0.075,+0.115]$ & -- & baseline \\
\quad $\alpha{=}0.5$ & $3$ & $+0.112$ & $[+0.105,+0.115]$ & $+0.013$ & gain (desc.) \\
\quad $\alpha{=}5$ & $3$ & $+0.068$ & $[+0.050,+0.100]$ & $-0.030$ & deficit (desc.) \\
\bottomrule\end{tabular}
\end{table*}

\begin{figure*}[t]\centering
\includegraphics[width=\textwidth]{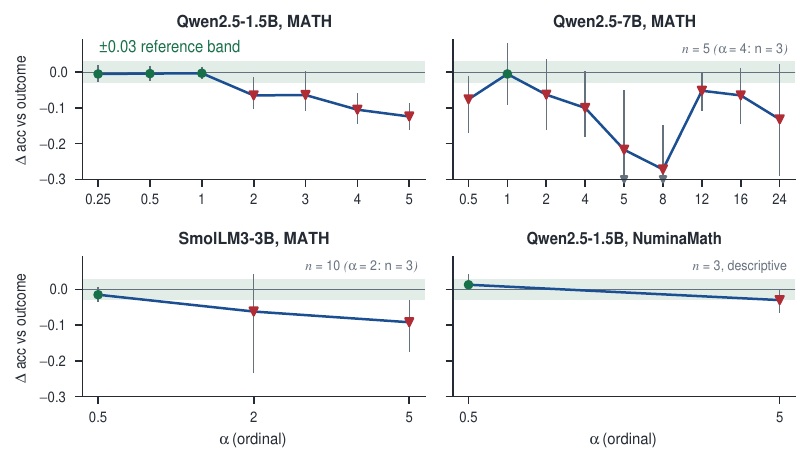}
\caption{\textbf{Reported weight sweeps in four training settings.} Paired greedy-accuracy differences versus outcome-only, with asymmetric $95\%$ seed-bootstrap intervals and a $\pm0.03$ reference band. Color labels summarize the plotted estimates; only the primary low-weight $1.5$B configurations have certified equivalence. The $7$B and NuminaMath configurations have $3$--$5$ and $3$ seeds; SmolLM3 endpoints have $10$. Carets show where the $7$B lower tails extend to $-0.411$ and $-0.385$. The positive single-seed Phi-4 pilots appear in \Cref{app:endtask}.}
\label{fig:doseresponse}
\end{figure*}

\section{What a Dense Reward Adds to One Update}
\label{sec:theory}

This section states the fixed-batch results behind the alignment test of \Cref{sec:tests}. They need only linear algebra and apply to any differentiable policy on a fixed batch. Proofs are in \Cref{app:proofs}, and \Cref{tab:notation} collects the notation.

\textbf{Notation.} We use the notation of \Cref{sec:grpo}, with completions of length $H$ and parameters $\theta\in\R^P$. Stacking the token scores $s_{g,h}$ into $S\in\R^{GH\times P}$, a token-level credit $w$ (one coefficient per sampled token) contributes
\begin{equation}
\label{eq:gradmap}
\grad(w)=S^\top\vecop(w),
\end{equation}
before token-specific clipping, importance weighting, KL terms, and optimizer state, which determine the effective update. Let $B\in\R^{G\times GH}$ sum tokens within each completion (in this section and in \Cref{app:proofs}, $B$ denotes this matrix, not the token budget), $K=SS^\top$, $V=\colspace(S^\top B^\top)=\Span\{q_1,\dots,q_G\}$, $\PV=\operatorname{Proj}_V$, and $\Pperp=I-\PV$. Any reweighting of whole completions updates within $V$; this is broader than the outcome baseline, whose centered coefficients span only $S^\top B^\top\mathbf1^\perp$. We write $w=\wtraj+\within$, where $(\wtraj)_{g,h}=H^{-1}\sum_jw_{g,j}$ and $B\vecop(\within)=0$, and $J(\theta)=\E_{\tau\sim\pi_\theta}R(\tau)$ for the expected accuracy, with gradient $h=\nabla J(\theta)$.

\subsection{What can a dense reward add to one update?}

Outcome coefficients and completion means produce updates in $V$ (\Cref{prop:decomp,prop:confine}). The next result gives the exact condition under which a token-level credit stays in this space, accounting for the score map's kernel: different token credits can produce the same gradient.
\begin{theorem}[Sample-conditional equivalence]
\label{thm:collapse}
For every credit $w$,
\[
\begin{gathered}
\grad(w)\in V\quad\Longleftrightarrow\\
\vecop(w)\in\colspace(B^\top)+\ker(S^\top).
\end{gathered}
\]
Moreover, $\dim\bigl(\Pperp S^\top\ker B\bigr)=\rank(S)-\dim V$. If $\rank(S)=GH$, equivalence holds exactly for completion-constant credits, and the residual dimension is $G(H-1)$.
\end{theorem}
Centering credit within each completion does not generally make its gradient orthogonal to $V$ (\Cref{thm:overlap}). Once a credit adds a direction outside $V$, we can ask whether that direction helps a local target.

\begin{proposition}[First-order projection identity]
\label{thm:nns}
For arbitrary $h=\nabla J(\theta)$ and dense direction $d=\grad(w)$, the first-order advantage of $d$ over its matched projection $\PV d$ is
\[
\langle h,d-\PV d\rangle=\langle\Pperp h,\Pperp d\rangle.
\]
\end{proposition}
Only the components of $d$ and $h$ outside $V$ decide whether $d$ improves on its projection to first order. This is the alignment test of \Cref{sec:tests}, which \Cref{sec:equalround} measures against a sampled reference. It compares a direction with its own projection at one policy, whereas an outcome-only training run follows its own sequence of policies and sampled batches.

\section{Notation}
\label{app:notation}
\Cref{tab:notation} collects the notation for the sampled score map, the population gradient, and the experiments. These objects describe different parts of the analysis and need not coincide.
\begin{table*}[htbp]\centering\small
\setlength{\tabcolsep}{4pt}
\caption{\textbf{Notation and scope.} All projectors use the Euclidean parameter-space metric unless stated otherwise.}
\label{tab:notation}
\begin{tabular}{@{}>{\raggedright\arraybackslash}p{0.25\textwidth}@{\hspace{6pt}}>{\raggedright\arraybackslash}p{0.70\textwidth}@{}}\toprule
Symbol & Meaning \\\midrule
$G,H,P$ & Group size, completion length and parameter dimension \\
$A_g$ & Group-relative completion coefficient on the sampled batch \eqref{eq:adv} \\
$s_{g,h},q_g$ & Token score and completion score sum \\
$S,B,K=SS^\top$ & Sampled score matrix, completion-sum operator and score Gram matrix \\
$\grad(w)=S^\top\vecop(w)$ & Conditional score contribution of token coefficients \\
$\wtraj,\within$ & Completion-constant and within-completion zero-sum parts of $w$ \\
$V=\colspace(S^\top B^\top)$ & Comparison span allowing arbitrary completion coefficients \\
$\PV,\Pperp$ & Projectors onto $V$ and its orthogonal complement \\
$Y=S^\top\ker B$ & Within-completion credits mapped into parameter space \\
$\gout=S^\top B^\top A$ & Unclipped sampled outcome score contribution \\
$J,h=\nabla J$ & Population expected accuracy and its full gradient \\
$\widehat h_{\mathrm{MC}}$ & Estimated Monte-Carlo progress-gradient reference \\
$\alpha,\Rbud$ & Experimental dense weight and total continuation budget \\
\bottomrule
\end{tabular}
\end{table*}

\begin{figure*}[t]
\centering
\includegraphics[width=\textwidth]{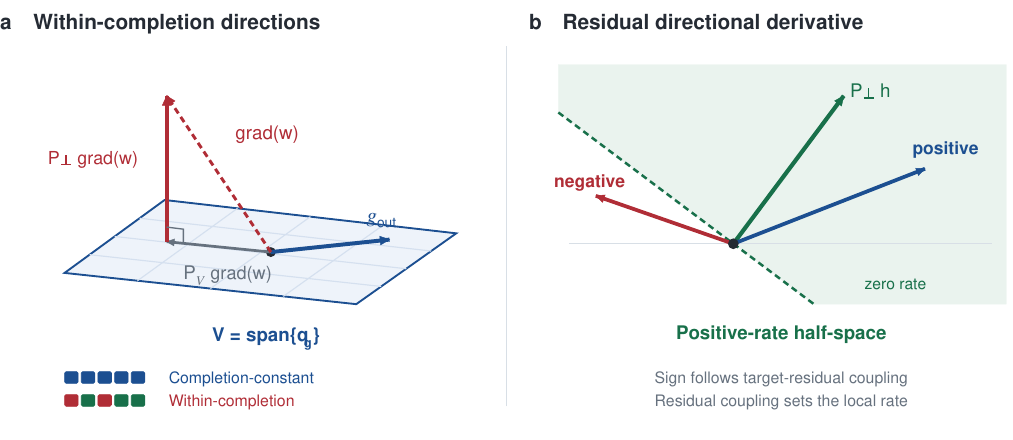}
\caption{\textbf{Which part of a dense update is new?} \textbf{(a)} Completion-constant coefficients produce updates in the span $V$ of completion score sums. A within-completion credit can add a residual direction $\Pperp\grad(w)$, while also having a component $\PV\grad(w)$ in $V$. \textbf{(b)} For a target gradient $h$, the residual directional derivative $\langle\Pperp h,\Pperp\grad(w)\rangle$ is positive in the shaded half-space. The dashed line is the zero-rate boundary perpendicular to the target residual. This local comparison uses the projected direction as its baseline. It does not guarantee a gain over a separate outcome-only training run.}
\label{fig:schematic}
\end{figure*}

\section{Proofs and Supporting Geometry}
\label{app:proofs}

All vector spaces below are finite-dimensional and use the Euclidean inner product. We flatten credits before multiplying them by $S^\top$ or $B$. The results allow rank-deficient score matrices, including $S=0$.

\subsection{The sampled score map}

\begin{proposition}[Decomposition]
\label{prop:decomp}
For every $w$, $\grad(w)=\grad(\wtraj)+\grad(\within)$ and $\grad(\wtraj)=\sum_g\bar w_gq_g\in V$.
\end{proposition}
\begin{proof}
Apply the linear map $S^\top$ to the decomposition of token coefficients. A completion-constant credit is $B^\top\bar w$, so its image is $S^\top B^\top\bar w$.
\end{proof}

\begin{proposition}[Outcome confinement on the realized batch]
\label{prop:confine}
The set of score contributions with arbitrary real completion coefficients equals $V$, and $\gout\in V$. Restricting coefficients to $\mathbf1^\perp$ gives $S^\top B^\top\mathbf1^\perp$, of dimension at most $G-1$.
\end{proposition}
\begin{proof}
The unrestricted set is the image of $Q=S^\top B^\top$. Restricting its domain gives the second image and rank bound. The sampled rewards further constrain the normalized, zero-sum advantages. They select one update from this comparison space.
\end{proof}

\begin{proof}[Proof of \Cref{thm:collapse}]
The condition $S^\top w\in\colspace(S^\top B^\top)$ holds if and only if there is a $c$ with $S^\top(w-B^\top c)=0$. This is exactly $w\in\colspace(B^\top)+\ker(S^\top)$. If $S$ has full row rank, its transpose is injective, so the kernel is zero. The dimension claim follows from the next result.
\end{proof}

\begin{theorem}[Reachable residual dimension]
\label{thm:reach}
Let $Y=S^\top\ker B$ and $W=\colspace(S^\top)$. Then $W=V+Y$ and
\[
\dim(\Pperp Y)=\rank(S)-\dim V\le G(H-1).
\]
Equality with the right side holds when $\rank(S)=GH$.
\end{theorem}
\begin{proof}
The coefficient space decomposes as $\R^{GH}=\colspace(B^\top)\oplus\ker B$. Applying $S^\top$ gives $W=V+Y$. On $Y$, projection onto $V^\perp$ has kernel $V\cap Y$. Therefore,
$\dim(\Pperp Y)=\dim Y-\dim(V\cap Y)=\dim W-\dim V$.
It is at most $\dim\ker B=G(H-1)$. If $S^\top$ is injective, $\dim W=GH$ and $\dim V=G$. At $S=0$ both sides of the exact identity are zero; at $H=1$ the within space is zero.
\end{proof}

\begin{theorem}[Projection through the score Gram matrix]
\label{thm:overlap}
For $B\omega=0$, the projection of $S^\top\omega$ onto $V$ is $S^\top B^\top c$, where
\begin{equation}
\label{eq:normaleq}
(BKB^\top)c=BK\omega.
\end{equation}
The Moore--Penrose solution gives the minimum-norm coefficient vector if the matrix is singular. If $S^\top\omega\ne0$, the fraction of the squared norm that lies in $V$ is
\[
\frac{c^\top(BKB^\top)c}{\omega^\top K\omega}.
\]
Orthogonality holds exactly when $BK\omega=0$.
\end{theorem}
\begin{proof}
Least squares for $Q=S^\top B^\top$ and $d=S^\top\omega$ gives $Q^\top Qc=Q^\top d$. These normal equations are consistent and give a unique projected vector even when $c$ is not unique. Substituting $K=SS^\top$ gives \eqref{eq:normaleq}. The numerator is the squared norm of that vector, and the denominator is $\|d\|^2$. If $d=0$, the fraction is undefined. If $K=\lambda I$ for $\lambda>0$, every within-completion credit is orthogonal to $V$; positive overlap therefore does not hold in every case.
\end{proof}

\begin{proof}[Proof of \Cref{thm:nns}]
The complementary orthogonal projectors are self-adjoint and satisfy $\PV\Pperp=0$. Thus $\langle h,d\rangle=\langle\PV h,\PV d\rangle+\langle\Pperp h,\Pperp d\rangle$. Subtracting the rate of $\PV d$ proves the identity. We do not need to represent $h$ through the finite sampled matrix. If we use an estimator $\widehat h$, the rate error satisfies
\[
\big|\langle\Pperp(\widehat h-h),\Pperp d\rangle\big|
\le\|\widehat h-h\|\,\|\Pperp d\|.
\]
An empirical alignment measurement alone does not bound this estimation error.
\end{proof}

\subsubsection{Telescoping is not confinement}
\label{app:telescoping}
Take one two-token completion with linearly independent scores $s_1,s_2$. The credit $w=(1,-1)$ has zero total, yet $s_1-s_2\notin\Span\{s_1+s_2\}=V$. It can represent the potential increments along a path with values $0,1,0$. Broadcasting the total instead produces the zero credit in $V$. Potential-based shaping results concern particular reward, return and baseline constructions \citep{ng1999shaping}; a telescoping total alone does not constrain the sampled score sum weighted by token coefficients. Likewise, when all group rewards agree, $A=0$ and the outcome update is zero, but this does not guarantee that a nonzero dense update improves $J$.

\section{The Expected GRPO Update}
\label{sec:lifted}

The results of \Cref{sec:theory} describe one sampled batch. This section gives the expected outcome-only GRPO update over fresh samples, which the matched-compute methods use to explain how adding a dense reward can change the relative weight of prompts (\Cref{app:matched}).

\subsection{The population gradient is not confined to a sampled span}

\begin{proposition}[Expected binary group-relative score]
\label{prop:population}
Fix a prompt and a differentiable distribution over completions. Assume its support is independent of the parameters and that differentiation can pass under the expectation. Let $R(\tau)\in\{0,1\}$ be a fixed verifier, $p(\theta)=\E R\in(0,1)$, $G\ge2$, and sample completions iid. Define $q_g=\nabla\log\pi_\theta(\tau_g)$, $K_R=\sum_gR_g$, and let $s_k$ be the group reward standard deviation when $K_R=k$. With advantages \eqref{eq:adv},
\begin{equation}
\label{eq:population-grpo}
\begin{gathered}
\E\left[\frac1G\sum_g A_gq_g\right]=\lambda_G(p)\nabla p,\\
\lambda_G(p)=\frac{\E[K_R(G-K_R)/(G^2(s_{K_R}+\epsilon))]}{p(1-p)}>0.
\end{gathered}
\end{equation}
Groups with $K_R\in\{0,G\}$ contribute zero.
\end{proposition}
\begin{proof}
The score identity gives $\E q=0$ and $\E[Rq]=\nabla p$, hence
$\E[q\mid R=1]=\nabla p/p$ and $\E[q\mid R=0]=-\nabla p/(1-p)$.
Condition on $K_R=k$. By exchangeability, the expected sums of successful and failed completion scores are $k\nabla p/p$ and $-(G-k)\nabla p/(1-p)$. Their advantage coefficients are $(G-k)/(G(s_k+\epsilon))$ and $-k/(G(s_k+\epsilon))$. Substituting these values and applying the outer $1/G$ gives
\[
\E\left[\frac1G\sum_g A_gq_g\,\middle|\,K_R=k\right]
=\frac{k(G-k)}{G^2(s_k+\epsilon)p(1-p)}\nabla p.
\]
Averaging proves the formula. Since $p\in(0,1)$ and $G\ge2$, a mixed group has positive probability. The multiplier is therefore positive. At $p=0$ or $p=1$, all sampled advantages are zero and the displayed ratio is undefined.
\end{proof}

This score-function identity does not identify a new obstacle to population learning. Population GRPO analyses also study the normalization and success dynamics \citep{mroueh2026grpo}. Across prompts, $\lambda_G(p)$ can vary and reweight their gradients. Bias from clipping, off-policy sampling or other transformations needs a separate analysis. An update can belong to its sampled $V$ without its expectation belonging to any one realized $V$.

\end{document}